\documentclass{article}

\usepackage[nonatbib, final]{nips_2017}

\usepackage[utf8]{inputenc} 
\usepackage[T1]{fontenc}    
\usepackage{hyperref}       
\usepackage{url}            
\usepackage{booktabs}       
\usepackage{amsfonts}       
\usepackage{nicefrac}       
\usepackage{microtype}      
\usepackage{graphicx}
\usepackage{array}
\usepackage{subcaption}
\usepackage{tikz-cd}
\usepackage[ruled,vlined]{algorithm2e}
\usepackage{booktabs}
\usepackage{amssymb}
\usepackage{amsmath}

\newcolumntype{M}[1]{>{\centering\arraybackslash}m{#1}}
\numberwithin{equation}{section}

\graphicspath{
	{./figures/},
	{./figures/dg/},
	{./figures/elm_0123_des/table/exp2_mins/},
	{./figures/elm_0123_des/table/exp5_mins/},
	{./figures/elm_0123_des/table/exp3_mins/},
	{./figures/elm_0123_des/table/exp4_mins/},
	{./figures/elm_0123_des/table/exp1_mins/},
	{./figures/mag/},
	{./figures/ising/},
	{./figures/meta_prac/},
	{./figures/meta_prac2/},
	{./figures/SK_GWL/},
	{./figures/SK_AD/},
	{./figures/ising_overlay/},
	{./figures/ising_interpolation/},
    {./figures/elm_0123_des/},
	{./figures/03gen/},
	{./figures/elm_tree_spots1/},
	{./figures/elm_tree_spots2/},
	{./figures/elm_tree_digit/},
	{./figures/neb_comp/},
	{./figures/ivy_ims/},
	{./figures/dg_ivy/},
	{./figures/dg_ivy_multi/},
	{./figures/dg_cat/},
	{./figures/meta_theory/},
	{./figures/GMM_2D/},
	{./figures/spotstripes/},
	{./figures/galaxy/}
}

\title{Building a Telescope to Look Into High-Dimensional Image Spaces}
\author{
  Mitch Hill \\
  University of California, Los Angeles\\
  \texttt{mkhill@ucla.edu} \\
  \And
  Erik Nijkamp \\
  University of California, Los Angeles \\
  \texttt{enijkamp@ucla.edu} \\
  \And
  Song-Chun Zhu \\
  University of California, Los Angeles\\
  \texttt{sczhu@stat.ucla.edu} \\
}

\begin{document}

\maketitle

\begin{abstract}
In Grenander's work, an image pattern is represented by a probability distribution whose density is concentrated on different low-dimensional subspaces in the high-dimensional image space. Such probability densities have an astronomical number of local modes corresponding to typical pattern appearances. Related groups of modes can join to form macroscopic image basins (known as Hopfield memories in the neural network community) that represent pattern concepts. Grenander pioneered the practice of  approximating an unknown image density with a Gibbs density. Recent works continue this paradigm and use neural networks that capture high-order image statistics to learn Gibbs models capable of synthesizing realistic images of many patterns. However, characterizing a learned probability density to uncover the Hopfield memories of the model, encoded by the structure of the local modes, remains an open challenge. In this work, we present novel computational experiments that map and visualize the local mode structure of Gibbs densities. Efficient mapping requires identifying the global basins without enumerating the countless modes. Inspired by Grenander's jump-diffusion method, we propose a new MCMC tool called Attraction-Diffusion (AD) that can capture the macroscopic structure of highly non-convex densities by measuring \textit{metastability} of local modes. AD involves altering the target density with a \textit{magnetization} potential penalizing distance from a known mode and running an MCMC sample of the altered density to measure the stability of the initial chain state. Using a low-dimensional generator network to facilitate exploration, we map image spaces with up to 12,288 dimensions ($64\times 64$ pixels in RGB). Our work shows: (1) AD can efficiently map highly non-convex probability densities, (2) metastable regions of pattern probability densities contain coherent groups of images, and (3) the perceptibility of differences between training images influences the metastability of image basins.
\end{abstract}

\section{Introduction}\label{sec:intro}
\subsection{Motivation}\label{subsec:motivation}
Representing image patterns requires reconciling the common structure present among images with the variability that exists between images, and addressing this tension is the central theme of Ulf Grenander's pioneering body of work on Pattern Theory \cite{grenander_book2, grenander_jump, grenander_prob, grenander_book}. As a concrete example, a digit can be written in many different ways, but humans can still recognize a common concept across the change in appearance. Grenander studied classical mathematics early in his career, but came to believe that the models of the time were too rigid to capture the rich variation found in real-world phenomena. Shifting his focus, he initiated the study of Pattern Theory in the 1960's, at a time when almost no literature or known uses existed. His countless contributions have led the field to the prominent role it plays today in many academic disciplines and practical applications.

Stochastic models, where an image $I$ is treated as a sample from a probability density $f$ over the image space, are well-suited for accommodating the tension between structure and variation that exists in real-world patterns. The statistical concept of a probability density $f$ and the physical concept of a diffusion process on the potential energy manifold $U = - \log f$ are equivalent and throughout the paper we use both perspectives interchangeably, although we focus more on the second view. Since $I$ is a random sample from $f$, image appearance can vary stochastically, but the probability of observing an image is virtually zero except for a small region around the modes of $f$, enforcing structure in the sampled images. Stochastic image models in high-dimensional spaces are the principle objects of study in Grenander's work.

When modeling image patterns, the true density $f$ is unknown. Grenander realized early in his career that designing an analytical formulation of $f$ from first principles was a hopeless task for real-world patterns. Instead, Grenander sought a family of probability models $\mathcal{P}$ flexible enough to approximate many different pattern densities. Real images, treated as independent samples from $f$, are used to find a model $p\in \mathcal{P}$ that is a good approximation for $f$, usually by MLE. Grenander was particularly interested in the family of Gibbs distributions defined on a graph over the pixel lattice, and he validates the capabilities of this family in many experiments. Recent advances have further increased the representational capacity of Gibbs image models (see Section \ref{subsec:pattern_model}).

\begin{figure}[h]
\centering
\begin{tabular}{cc}
\includegraphics[height=3cm]{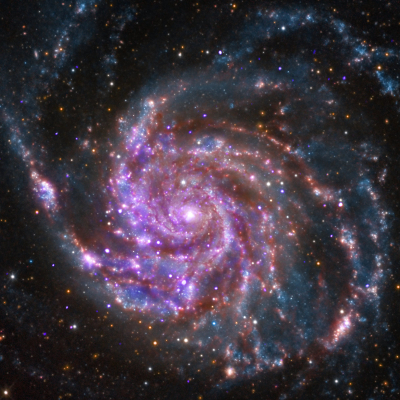}$\quad$& $\quad$
\includegraphics[height=3cm]{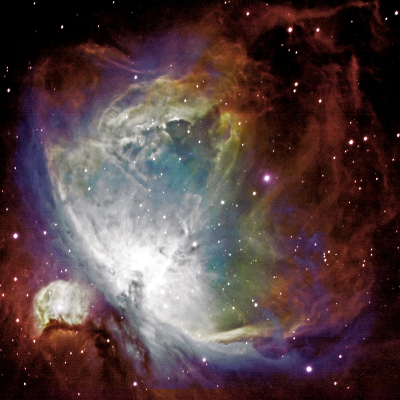}
\end{tabular}
\caption{Analogies for energy basins of images in different entropy regimes. Low-entropy images have distinct appearances and create galaxies with macroscopic substructure, like the arms of the spiral galaxy on the left. High-entropy images such as textures cannot be easily distinguished and form wide energy basins with little substructure, like the nebula on right. See Section \ref{subsec:infoscaling}.}
\label{fig:galaxy}
\end{figure}

In this paper, we investigate the structure of a learned Gibbs density $p$ (or equivalently, energy $E = - \log p$) trained to model an unknown image density $f$. During training, the density learns to form modes around the samples of $f$, and local minima of $E$ can be interpreted as ``memories" of the training data, as in Hopfield's model \cite{hopfield}. Regions of the image space separated only by low barriers in $E$ represent groups of images/memories that are conceptually similar. One can imagine the image space as a vast and mostly empty universe, $E$ as gravitational potential energy, and the local minima of $E$ as dense stars that lie on the pattern manifold. Groups of related local minima separated by low energy barriers (such as different images of the same digit) form connected clusters of pattern images, which are ``galaxies" in the image universe (see Figure \ref{fig:galaxy}).

Following the approach of Bovier \cite{bovier}, one can formally characterize image galaxies by dividing the image space into \emph{metastable} regions, such that a diffusion process on $E$ mixes over short time-scales within a region, while mixing occurs over long time-scales between regions. In other words, a local MCMC sample of $p$ initiated from an image galaxy will travel in the same galaxy for a very long time, because random fluctuations are enough to overcome small energy barriers within the galaxy, while much larger energy barriers restrict movement between galaxies. This view is closely related to Grenander's jump-diffusion method \cite{grenander_jump}, which uses a combination of local diffusion in a limited region of the state space and global proposals that jump between separate regions of the state space to facilitate sampling. Our primary goal in this paper is to computationally identify metastable regions in an image density while only visiting a few of the local modes within each region, because exhaustive enumeration of modes is computationally infeasible.

The galaxies represent different concepts in the image pattern, and by finding the galaxies of an image density we can reduce the vast high-dimensional image space to a few groups that summarize the major pattern appearances. We are also interested in measuring the energy barriers between galaxies, because they encode similarity between groups. The structure of a learned image density encodes memories of the pattern manifold, but this information is hidden in $p$ and must be recovered through mapping. Landscape structure varies according to the pattern being modeled, as in Figure \ref{fig:galaxy} and Figure \ref{fig:dg_ivy_multi}. In particular, we conjecture that the depth/stability of image basins is related to the human ability to distinguish between pattern images. This idea can be understood by examining the energy landscape of the same pattern at different scales (see Section \ref{subsec:infoscaling}).

The formulation, tools, and goals of this paper can all be traced back to Grenander's legacy of research on image models. Grenander was among the first to understand the importance of Gibbs distributions as a flexible and powerful family of models for representing complex data, and he used Gibbs models extensively throughout his work. He is one of the pioneers of MCMC computing, and his celebrated jump-diffusion method is closely linked with metastable descriptions of energy landscapes. Nonetheless, Grenander faced several major obstacles which prevented him from realizing the full potential of  probabilistic image representations. The challenges are listed below.

\medskip

\begin{enumerate}
\item Difficulty of defining meaningful potential functions for Gibbs models
\item Patterns of different scales require separate representations
\item Sampling from high-dimensional image distributions is expensive
\item Energy functions of images have highly non-convex structure
\end{enumerate}

\medskip

Recent advances in image modeling have made great progress towards resolving the first two issues (see Section \ref{subsec:pattern_model}), and this paper tackles the last two difficulties. By overcoming central challenges of Grenander's time, our work is the first to computationally map the structure of Hopfield memories of a Gibbs image distribution. We make several major contributions to the study of probabilistic image models and non-convex energy functions, including:

\medskip

\begin{enumerate}
\item An MCMC tool for detecting metastable regions of highly non-convex energy landscapes
\item A new procedure for mapping the macroscopic structure of non-convex energy landscapes at different resolutions
\item A new method for finding low-energy interpolations between local minima in both discrete and continuous energy landscapes
\item Use of a low-dimensional generator network to facilitate sampling and mapping in the high-dimensional image space
\item Novel energy-based mappings of pattern concepts in both the image space and the latent space of a generator network
\item Experimental evidence linking the perceptibility of difference among pattern images and the stability of image basins in a learned landscape
\end{enumerate}

\medskip

The paper is organized as follows. In Section 1, we give an overview of our motivation, method, and results. Section 2 summarizes previous work that is relevant to our research. Section 3 introduces Attraction-Diffusion, our proposed MCMC technique, and Section 4 describes a framework for mapping the energy landscape using Attraction-Diffusion. In Section 5, we apply our new method to map the local minima structure of the SK spin-glass Hamiltonian and energy-based image models learned by neural networks.

\subsection{Information Scaling and the Energy Landscape}\label{subsec:infoscaling}
 Image scale should have a strong influence on the structure of image memories. In one of the central paradigms of pattern representation, Julesz identifies two major regimes of image scale: texture and texton. \textit{Textures} are high-entropy patterns defined as groups of images sharing the same statistics among nearby pixels \cite{Julesz62}. \textit{Textons}, on the other hand, are low-entropy patterns, and can be understood as the atomic building elements or local, conspicuous features such as bars, blobs or corners \cite{Julesz1981}.

\begin{figure}[h]
\centering
\begin{tabular}{ccccc}
	\includegraphics[height=1.5cm,trim={.1cm 0 0 0},clip]{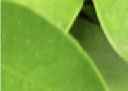}&
	\includegraphics[height=1.5cm]{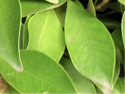}&
	\includegraphics[height=1.5cm]{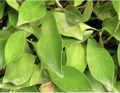}&
	\includegraphics[height=1.5cm]{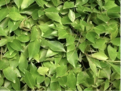}&
	\includegraphics[height=1.5cm,trim={.3cm 0 0 0},clip]{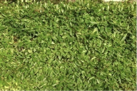}
\end{tabular}
\begin{tabular}{cc}
	\textbf{texton} 
	\hspace{0.9cm}
	$\underset{scale}{\xrightarrow{\hspace*{6cm}}}$
	\hspace{0.7cm}
	\textbf{texture}
\end{tabular}
\caption{Ivy leaves at different scales. As image scale increases from left to right, an increasing variety of image groups can be identified, until one reaches the threshold of perceptibility, after which it becomes difficult to distinguish between images. The fourth scale is close to the perceptibility threshold of humans, while the fifth scale is beyond human perceptibility. A regime transition from explicit, sparse structure to implicit, dense structure occurs as the threshold is crossed. A similar transition occurs in the energy landscape (see Figure \ref{fig:dg_ivy_multi}).}
\label{fig:scalespace}
\end{figure}

As illustrated in Figure~\ref{fig:scalespace}, texton-scale images have explicit structure that is easily recognizable, and this structure allows humans to reliably sort texton images into coherent groups. Texture-scale images have implicit structure, and it is usually difficult or impossible to find groups among images of the same texture, because no distinguishing features can be identified within a texture ensemble. As image scale increases, the number of recognizable image groups tends to increase until one reaches the threshold of perceptibility, where texton-scale images transition into texture-scale images and humans begin to lose the ability to identify distinguishing features \cite{scalespace}. Beyond the threshold of perceptibility, texture images cannot be told apart or reliably sorted into groups. Change of image scale causes a change in the statistical properties of an image, and we call this phenomenon \textit{Information Scaling}. 

\begin{figure}[h]
	\centering
	\includegraphics[width=\textwidth]{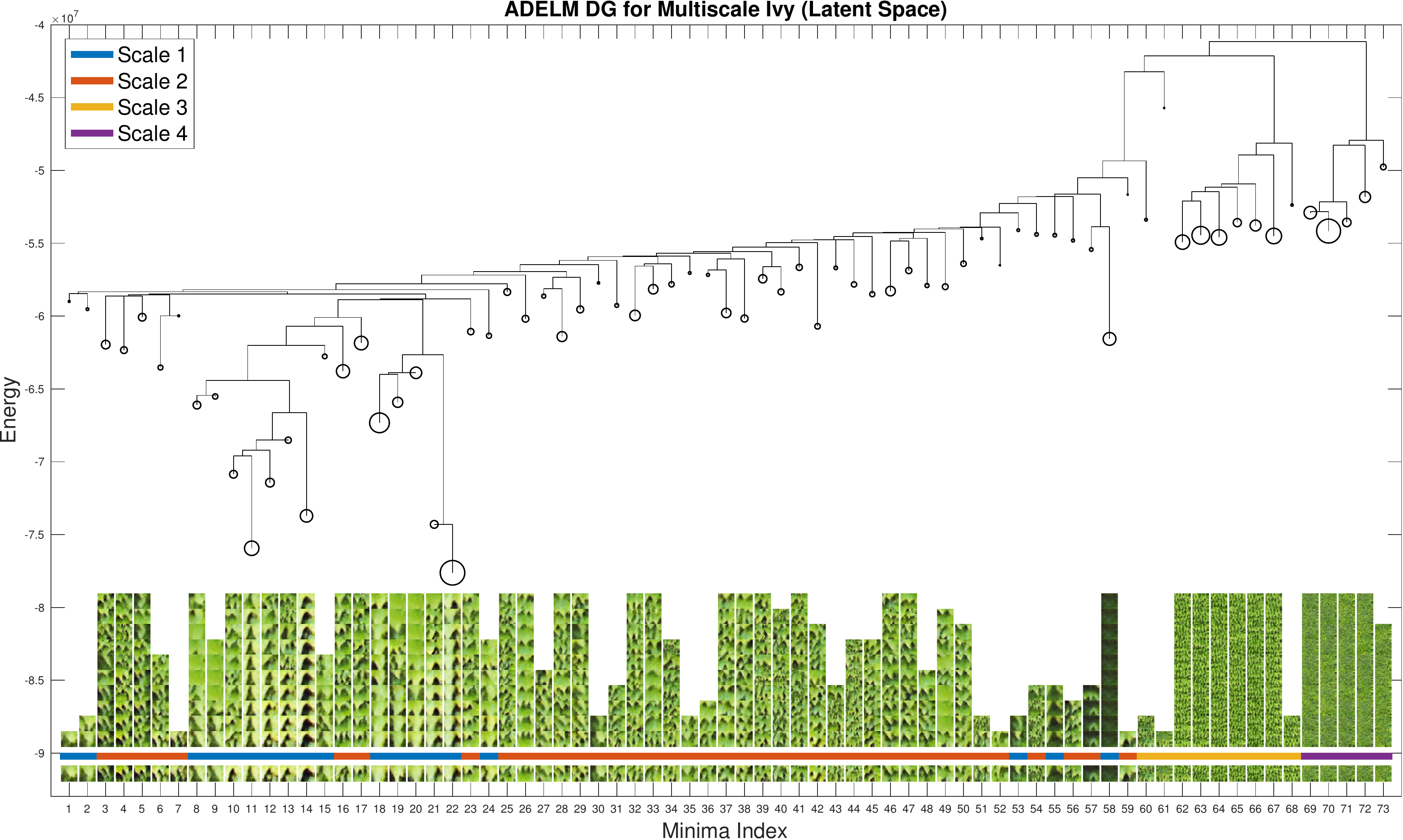}
	\caption{Landscape of ivy image patches at four different scales. Images from Scale 1 and Scale 2 are textons, while images from Scale 3 and Scale 4 are textures. The texton-scale images account for the majority of the basins in the landscape. More basins are identified for Scale 2 than Scale 1 because Scale 2 has a richer variety of distinct appearances, while the Scale 1 minima have lower energy, since appearances from this scale are more reliable. The texture-scale images form separate basins with little substructure. Basin members from each scale are shown in Figure \ref{fig:dg_ivy_multi_table}. See Section \ref{subsec:exp_coop} for a full explanation.} 	\label{fig:dg_ivy_multi}
\end{figure}

We conjecture that Information Scaling is reflected in the structure of the image landscape, and that there is a connection between the perceptibility of differences between pattern images and the stability/depth of local minima images. When the landscape models texton-scale images, where groups among the images can easily be distinguished, we expect to find many separate, stable basins in the landscape encoding the separate appearances of the groups. Landscapes that model texture-scale images, on the other hand, should exhibit behavior similar to human perception and form a single macroscopic basin of attraction with many shallow local minima to encode the texture. By mapping images from the same pattern at multiple scales, we show that the transition in perceptibility that occurs between scales results in a transition in the landscape structure of image memories (see Figure \ref{fig:dg_ivy_multi}).

\begin{figure}[h]
\begin{center} 
	\textbf{Multiscale Ivy (Latent Space)} \\ 
	
	\hspace*{-.3cm}\begin{tabular}{M{.8cm}M{.8cm}M{10cm}M{1cm}} \toprule 
		Min. & Basin & Randomly Selected Members & Member \\ 
		Index & Rep. & (arranged from low to high energy) & Count \\ \midrule 
		11 & \includegraphics[scale = 0.25]{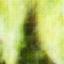} & \includegraphics[scale = 0.25]{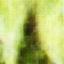} \includegraphics[scale = 0.25]{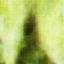} \includegraphics[scale = 0.25]{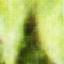} \includegraphics[scale = 0.25]{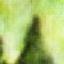} \includegraphics[scale = 0.25]{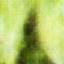} \includegraphics[scale = 0.25]{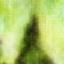} \includegraphics[scale = 0.25]{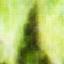} \includegraphics[scale = 0.25]{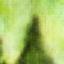} \includegraphics[scale = 0.25]{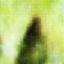} \includegraphics[scale = 0.25]{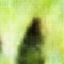} \includegraphics[scale = 0.25]{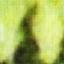} \includegraphics[scale = 0.25]{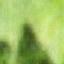} \includegraphics[scale = 0.25]{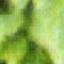} \includegraphics[scale = 0.25]{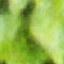} \includegraphics[scale = 0.25]{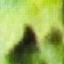}& 102\\ 
		22 & \includegraphics[scale = 0.25]{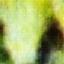} & \includegraphics[scale = 0.25]{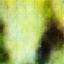} \includegraphics[scale = 0.25]{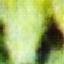} \includegraphics[scale = 0.25]{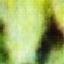} \includegraphics[scale = 0.25]{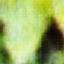} \includegraphics[scale = 0.25]{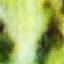} \includegraphics[scale = 0.25]{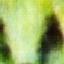} \includegraphics[scale = 0.25]{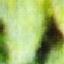} \includegraphics[scale = 0.25]{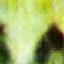} \includegraphics[scale = 0.25]{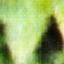} \includegraphics[scale = 0.25]{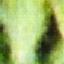} \includegraphics[scale = 0.25]{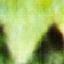} \includegraphics[scale = 0.25]{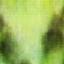} \includegraphics[scale = 0.25]{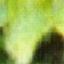} \includegraphics[scale = 0.25]{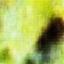} \includegraphics[scale = 0.25]{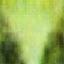}& 295\\ 
		3 & \includegraphics[scale = 0.25]{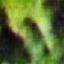} & \includegraphics[scale = 0.25]{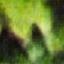} \includegraphics[scale = 0.25]{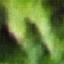} \includegraphics[scale = 0.25]{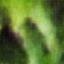} \includegraphics[scale = 0.25]{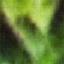} \includegraphics[scale = 0.25]{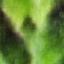} \includegraphics[scale = 0.25]{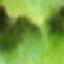} \includegraphics[scale = 0.25]{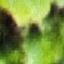} \includegraphics[scale = 0.25]{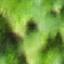} \includegraphics[scale = 0.25]{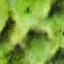} \includegraphics[scale = 0.25]{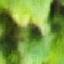} \includegraphics[scale = 0.25]{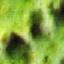} \includegraphics[scale = 0.25]{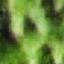} \includegraphics[scale = 0.25]{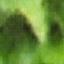} \includegraphics[scale = 0.25]{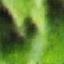} \includegraphics[scale = 0.25]{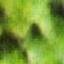}& 37\\ 
		38 & \includegraphics[scale = 0.25]{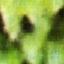} & \includegraphics[scale = 0.25]{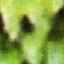} \includegraphics[scale = 0.25]{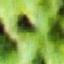} \includegraphics[scale = 0.25]{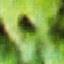} \includegraphics[scale = 0.25]{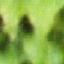} \includegraphics[scale = 0.25]{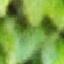} \includegraphics[scale = 0.25]{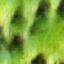} \includegraphics[scale = 0.25]{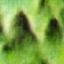} \includegraphics[scale = 0.25]{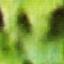} \includegraphics[scale = 0.25]{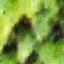} \includegraphics[scale = 0.25]{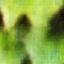} \includegraphics[scale = 0.25]{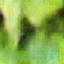} \includegraphics[scale = 0.25]{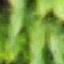} \includegraphics[scale = 0.25]{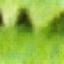} \includegraphics[scale = 0.25]{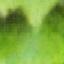} \includegraphics[scale = 0.25]{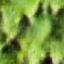}& 25\\ 
		63 & \includegraphics[scale = 0.25]{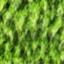} & \includegraphics[scale = 0.25]{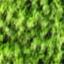} \includegraphics[scale = 0.25]{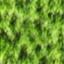} \includegraphics[scale = 0.25]{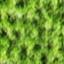} \includegraphics[scale = 0.25]{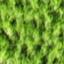} \includegraphics[scale = 0.25]{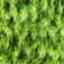} \includegraphics[scale = 0.25]{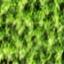} \includegraphics[scale = 0.25]{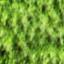} \includegraphics[scale = 0.25]{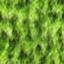} \includegraphics[scale = 0.25]{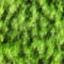} \includegraphics[scale = 0.25]{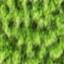} \includegraphics[scale = 0.25]{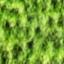} \includegraphics[scale = 0.25]{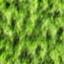} \includegraphics[scale = 0.25]{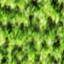} \includegraphics[scale = 0.25]{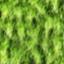} \includegraphics[scale = 0.25]{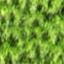}& 154\\ 
		64 & \includegraphics[scale = 0.25]{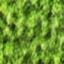} & \includegraphics[scale = 0.25]{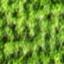} \includegraphics[scale = 0.25]{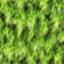} \includegraphics[scale = 0.25]{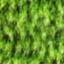} \includegraphics[scale = 0.25]{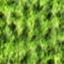} \includegraphics[scale = 0.25]{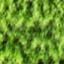} \includegraphics[scale = 0.25]{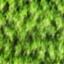} \includegraphics[scale = 0.25]{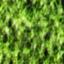} \includegraphics[scale = 0.25]{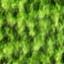} \includegraphics[scale = 0.25]{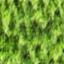} \includegraphics[scale = 0.25]{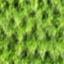} \includegraphics[scale = 0.25]{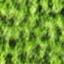} \includegraphics[scale = 0.25]{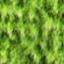} \includegraphics[scale = 0.25]{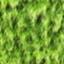} \includegraphics[scale = 0.25]{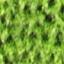} \includegraphics[scale = 0.25]{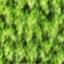}& 117\\ 
		70 & \includegraphics[scale = 0.25]{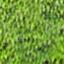} & \includegraphics[scale = 0.25]{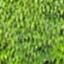} \includegraphics[scale = 0.25]{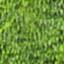} \includegraphics[scale = 0.25]{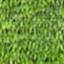} \includegraphics[scale = 0.25]{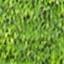} \includegraphics[scale = 0.25]{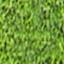} \includegraphics[scale = 0.25]{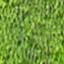} \includegraphics[scale = 0.25]{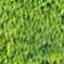} \includegraphics[scale = 0.25]{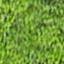} \includegraphics[scale = 0.25]{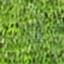} \includegraphics[scale = 0.25]{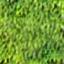} \includegraphics[scale = 0.25]{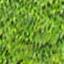} \includegraphics[scale = 0.25]{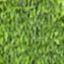} \includegraphics[scale = 0.25]{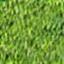} \includegraphics[scale = 0.25]{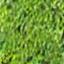} \includegraphics[scale = 0.25]{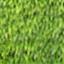}& 280\\ 
		71 & \includegraphics[scale = 0.25]{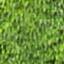} & \includegraphics[scale = 0.25]{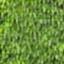} \includegraphics[scale = 0.25]{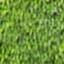} \includegraphics[scale = 0.25]{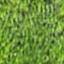} \includegraphics[scale = 0.25]{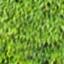} \includegraphics[scale = 0.25]{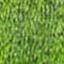} \includegraphics[scale = 0.25]{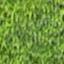} \includegraphics[scale = 0.25]{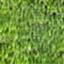} \includegraphics[scale = 0.25]{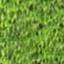} \includegraphics[scale = 0.25]{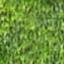} \includegraphics[scale = 0.25]{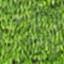} \includegraphics[scale = 0.25]{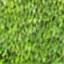} \includegraphics[scale = 0.25]{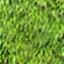} \includegraphics[scale = 0.25]{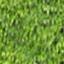} \includegraphics[scale = 0.25]{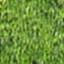} \includegraphics[scale = 0.25]{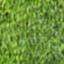}& 36\\ 
		\bottomrule
	\end{tabular} 
\end{center}
	\caption{Minima of multiscale ivy in latent space for the DG depicted in Figure~\ref{fig:dg_ivy_multi}. The appearance of randomly selected members is consistent with the appearance of the basin representative.} 	\label{fig:dg_ivy_multi_table}
\end{figure}

\subsection{Overview of Method and Experiments} \label{subsec:overview}
Characterizing the structure of energy functions of complex systems in terms of their local minima and the barriers between minima is an important but difficult task that can shed light on the behavior and properties of the system in question. In virtually all cases of interest, the size of the system is so vast that it is impossible to map the landscape by simply evaluating the energy of all possible states. Computational methods are needed to identify local minima and barriers while visiting only a tiny fraction of the system states. We refer to the task of computationally identifying the local minima structure of non-convex energy functions as \emph{Energy Landscape Mapping} (ELM).

Often, the number of local minima is also too vast for full enumeration. On the other hand, macroscopic structures (such as image ``galaxies") exist in many non-convex landscapes, even if the local structure is very noisy. Our work proposes a new MCMC ``telescope" that can efficiently discover macroscopic structures of complex landscapes in both continuous and discrete spaces. This is accomplished by updating MCMC samples using the energy function
\begin{equation}
E_{T, \, \alpha, \,X^\ast} (X) = E(X)/T + \alpha || X - X^\ast ||_2 \label{eqn:AD1}
\end{equation}
where $E$ is the target energy function, $T>0$ is temperature, $X^\ast$ is a known local minimum, and $\alpha>0$ is the strength of the penalty term. Our method can be viewed as a way of measuring the \emph{metastability} of a local minima in the target landscape $E$. Metastable basins can be found by carefully tuning $\alpha$ to accelerate the mixing time within basins while still respecting the long time-scales between basins. $E_{T,\, \alpha, \, X^\ast}$ can also be used to find low-energy interpolations between local minima. See Section \ref{sec:AD}.

In our experiments in Section \ref{sec:exp}, we map the structure of the DeepFRAME energy function (\ref{eqn:deepframe_energy1}) and the Co-Op Net energy function (\ref{eqn:coop_en1}) after the neural network weights have been learned (see Section \ref{subsec:pattern_model} for model descriptions). The DeepFRAME model and Co-Op Net model are good test settings for our mapping algorithm for several reasons.

\medskip

\begin{enumerate}
\item The energy functions should have multi-modal macroscopic structure if the training data can be grouped into different types of images (for example, handwritten digits). These modes can interpreted as Hopfield associative memories \cite{hopfield, xie, xie_coop}. The global energy basins will be noisy because of variation possible within the image groups. Our method is designed for this situation.
\item Since we are mapping energy functions defined over images, the ELM results should roughly correspond to human visual intuition if the mapping is successful. In this case, we can subjectively evaluate our ELM results.
\item Mapping the local minima structure of DeepFRAME and Co-Op energy functions is a novel application. Much work has been devoted to modeling real data using ConvNet functions, but less work has been done to investigate the structure of these functions after training.
\item Application to neural network models shows that our method can be successfully used on complex and modern energy functions.
\end{enumerate}

\medskip

We map image models trained to capture different patterns, and discover a variety of landscape structures. Despite the astronomical number of different image minima, we show that image memories form large structured basins, and that image appearance within global basins is very consistent. Opening up the black box of generative neural networks reveals that the models learn a handful of major image concepts, and our results support the conjecture that perceptibility in the training data influences the stability of image memories in the learned landscape.

\newpage

\section{Related Work}\label{sec:related}
\subsection{Probability Models of Image Patterns}\label{subsec:pattern_model} In practice, the true image density $f$ is unknown, and only training images, which are treated as independent samples from $f$, are available. To model the image space, one must approximate $f$ by selecting the density $p^\ast$ that is ``closest" to $f$ from a family of known densities $\{p\}$. When closeness is measured by KL-divergence, this procedure boils down to finding the MLE $p^\ast$ in the group $\{p\}$. To obtain an accurate approximation of $f$, the family of densities $\{ p\}$ must be flexible enough to accommodate the variation in the training data. 

Gibbs distributions defined on a pixel graph have been widely used as an effective family for modeling patterns of real images \cite{grenander_book, grenander_prob, geman, frame, zhu_texture}. This family of densities has the form
\begin{equation}
p(I) = \frac{1}{Z} \, \exp \left\{ -\sum_{C\in \mathcal{C}} \varphi_C (I_C) \right\} \label{eqn:gibbs}
\end{equation}
where $\mathcal{C}$ is the set of cliques of a graph $G$ over the pixel lattice, $\varphi_C$ are clique potentials over the pixels in clique $C$, and $Z$ is the normalizing constant. A clique is a group of pixels in which all pairs of pixels are adjacent on $G$. In early Gibbs image models, the cliques are groups of neighboring pixels and the potentials capture simple clique features, such as consistency of pixel intensity. However, these simple, hand-designed potentials are not capable of synthesizing realistic image patterns. The density (\ref{eqn:gibbs}) is very flexible, but the model is useless without a principled way to define clique potentials $\varphi_C$ that capture relevant features of the target density $f$.

Zhu et al. address this problem in the FRAME model \cite{frame} by using convolutional filters to define clique potentials. The FRAME density has the form 
\begin{equation}
p(I) = \frac{1}{Z} \, \exp  \left\{ -\sum_{k=1}^K \langle \lambda^{(k)}, \, H^{(k)}(I) \rangle \right\} \label{eqn:frame}
\end{equation}
where $H^{(k)}(I)$ is a histogram of image responses to convolutional filter $k$, and $\lambda^{(k)}$ is the potential for filter $k$. The potential $\lambda^{(k)}$ ensures that the histogram of filter responses $H^{(k)}(I)$ for the sampled image $I$ matches the histogram of filter responses $H^{(k)}(I_\textrm{obs})$ for the training image $I_\textrm{obs}$. The potentials $\lambda^{(k)}$ must be learned. Since filter convolution is a linear projection from the image space to a 1D subspace, matching the sample histogram $H^{(k)}(I)$ to the observed histogram $H^{(k)}(I_\textrm{obs})$ is equivalent to matching the marginal distribution of $p$ to the marginal distribution of $f$ in the 1D subspace of filter $k$. One can show that $p=f$ if and only if the marginal distribution of $p$ is the same as the marginal distribution of $f$ in all 1D linear subspaces. If the majority of the variation of $f$ is captured by a few marginal directions, matching only these marginal distributions should still give a close approximation for $f$. The FRAME model learns an image density $p$ by matching the marginal distribution of samples from $f$ in the most relevant filter subspaces.

In the original FRAME model, filters are selected from a pre-defined filter bank, which limits the kinds of patterns that can be represented. There is no guarantee that the filter bank can project onto the most relevant 1D subspaces of $f$, and synthesis results are poor when filters cannot capture important features of $f$. Hand-designing filter banks for each new pattern is not a viable solution, because this is just as difficult as hand-designing clique potentials. 

Recent trends in the neural network community have shown that learning the filters themselves during training can result in flexible and realistic image models. Including multiple layers of filter convolution can also lead to significantly better representations of complex data \cite{lecun_cnn, krizhevsky}. The DeepFRAME model \cite{lu, xie} extends the FRAME model to incorporate these new features. A DeepFRAME density for an image $I$ with $N$ pixels has the form
\begin{equation}
p(I | W) = \frac{1}{Z} \, \exp\{ F(I |  W)\} q(I)\label{eqn:deepframe_density1}
\end{equation}
where $q$ is the prior distribution  $\textrm{N}(0, \sigma^2 I_N)$ of Gaussian white noise, and the scoring function $F(\cdot | W)$ is defined by a ConvNet with weights $W$, which must be learned. The normalization constant $Z=\int \exp\left\{F(I |  W)\right\}q(I)dI$ is intractable. The associated energy function has the form
\begin{equation}
E(I | W) = - F(I|W) + \frac{1}{2\sigma^2} || I ||_2^2. \label{eqn:deepframe_energy1}
\end{equation}
We may interpret $p(I | W)$ as an exponential tilting of $q$ which has the effect of mean shifting. The non-linearity induced by the activation functions between network layers is essential for successful representation of real images. 

When the the activation functions are rectified linear units (ReLU), $F(\cdot | W)$ is piecewise linear in $I$, and the borders between linear regions are governed by the activations in the network \cite{montufar2014number}. Let $\Omega_{\delta,W} = \{I : \sigma_{k} (I | W) = \delta_{k}, \, 1\le k \le K \},$ where $W$ gives the network weights, $K$ is the number of activation functions in the entire network, $\sigma_{k} (I | W) \in \{0,1\}$ indicates whether activation function $k$ turns on for image $I$, and $\delta = (\delta_{1}, \, \dots , \, \delta_{K}) \in \{0, \, 1\}^K$. Since $F(I | W)$ is linear on $\Omega_{\delta,W}$ for all $\delta$, the energy can be written as
\begin{equation}
E(I | W) =   - (\langle I, \, B_{\delta,W} \rangle + a_{\delta,W}) + \frac{1}{2\sigma^2} || I ||_2^2 \label{eqn:piecewisegauss}
\end{equation}
for some constants $a_{\delta,W}$ and $B_{\delta,W}$, which shows that $I \sim \textrm{N}( \sigma^2 B_{\delta,W}, \sigma^2 I_N)$ on $\Omega_{\delta,W}$ and that $p(I | W)$ is piecewise Gaussian over the image space. This analysis also characterizes the local minima of $E(I | W)$, which are simply $\{ \sigma^2 B_{\delta, W} :  \sigma^2 B_{\delta, W} \in \Omega_{\delta,W} \}$. However, there is no guarantee that the Gaussian piece $\Omega_{\delta,W}$ contains its mode $\sigma^2 B_{\delta,W}$, and the number of Gaussian pieces is extremely large, so mapping a DeepFRAME model by identifying all Gaussian pieces is not viable.

Early image models often employ different representations to cover the scale spectrum. Sparse basis functions can effectively capture the features of texton images, while MRF distributions are more suitable for representing texture patterns \cite{frame, activebasis, scalespace}. The DeepFRAME density incorporates aspects of both families, because the filters serve as both implicit features and sparse basis functions for image synthesis \cite{lu,xie}. The DeepFRAME model provides a unified way to represent image patterns at many different scales, but we still expect to identify different structures in the image landscape across the scale spectrum.

Training an image model often requires repeatedly sampling from $p$. This can be computationally expensive in image spaces, because dimension scales with the square of the image width, so even small images are high-dimensional. Langevin Dynamics are often the sampling method of choice when conducting MCMC in high dimensions. Updating MCMC samples according to the Langevin Equation
\begin{equation}
dI(t) =  \nabla \log p(I(t))  dt + \sqrt{2} \, dW(t) ,
\end{equation}
where $W(t)$ is a Wiener process, preserves the distribution of $p$ \cite{geman_langevin, neal}. The gradient term in the Langevin Equation leads to faster convergence than methods such as Random-Walk Metropolis-Hastings and Gibbs sampling, which make no use of the local landscape geometry. Even with Langevin Dynamics, it is infeasible to obtain true independent samples of $p$ each time the model is updated. Contrastive Divergence (CD) \cite{hinton} and Persistent Contrasitve Divergence (PCD) \cite{pcd} are two common methods of obtaining approximate samples of $p$. In CD, the training images are used as the initial states of MCMC samples from $p$, while in PCD the images from the previous training iteration are used as the initial states. Unfortunately, CD and PCD appear to have an adverse effect on the structure of the learned landscape.

In this paper, we are interested in mapping the local minima structure of a learned energy $E$ of the form (\ref{eqn:deepframe_energy1}) for a DeepFRAME density (\ref{eqn:deepframe_density1}) which is trained to model the true, but unknown, image density $f$. However, mapping a DeepFRAME energy directly is problematic because the energy function learns many accidental low-energy regions that can obscure the relations between image basins. Since training relies on CD, the DeepFRAME energy only observes warm-start images that are already quite close to the pattern manifold. The landscape structure of a DeepFRAME energy is only meaningful in a small region around the pattern manifold, and vast low-energy basins can form in remote regions of the image space. Using a Gibbs sampler instead of Langevin Dynamics during training can alleviate the problem somewhat, at the cost of efficiency.

This difficulty can be overcome by introducing a generator network \cite{goodfellow, han} which learns a set of weights to transform a trivial latent distribution into a distribution over the image space that approximates the pattern manifold. The $N$-dimensional image data $I$ follows the distribution
\begin{equation}
I \sim \textrm{N}(g(Z|W_2 ), \tau^2 I_N) \label{eqn:gen_dist}
\end{equation}
with $Z\sim\textrm{N}(0,I_n)$ for $n \ll N$, variance parameter $\tau^2$, and weights $W_2$ of a ConvNet function $g$. The Co-Op Net Algorithm \cite{xie_coop} provides a way to simultaneously learn the weights $W_1$ of a DeepFRAME energy function $E(\cdot | W_1)$ and the weights $W_2$ of a generator network $g(\cdot | W_2)$ for any dataset. During training, the generator $g$ learns to mimic the manifold of $E$, while the energy function $E$ learns to model the real data. Following \cite{xie_coop}, the DeepFRAME network in the Co-Op Net model is referred to as the \emph{descriptor} network, because the energy function encodes a description of image features, which are expressed through filter activations.

By composing a generator network $g(\cdot | W_2)$ and descriptor energy $E(\cdot | W_1)$, we can define a new energy function
\begin{equation}
E_{W_1,W_2} (Z) = E(g(Z|W_2) |W_1)\label{eqn:coop_en1}
\end{equation}
over the latent space. This formulation is very similar to the DGN-AM model \cite{DGNAM} (see Section \ref{subsec:AM}). Sampling in the low-dimensional latent space vastly reduces computational cost, providing a way to efficiently explore the pattern manifold of realistically-sized images. Interestingly, it appears that the structure of image memories in the energy (\ref{eqn:coop_en1}) is more meaningful than the structure of memories in (\ref{eqn:deepframe_energy1}), because concatenating the generator and descriptor networks reduces the number of accidental low-energy regions found between pattern minima in the raw DeepFRAME energy (compare Figures \ref{fig:elm_tree_spots1} and \ref{fig:elm_tree_spots2}). The energy (\ref{eqn:coop_en1}) provides a way to characterize both the image space and the latent space of a generator network. Previous works have identified a handful of minima in the latent space using a similar energy function \cite{DGNAM}, but our work is the first to systematically explore and map the structure of a latent generator space.

\subsection{Macroscopic Structure of Non-Convex Landscapes} \label{subsec:complex_systems}
Energy functions associated with complex systems are often non-convex, and the degree of non-convexity in the energy landscape varies depending upon the system in question. In some settings, the landscape has only slight non-convexity, and optimizing the non-convex energy function leads to a solution close to the global minimum, as in \cite{loh}. In contrast, the loss surfaces of ConvNet classification and regression functions are highly non-convex, because symmetry-breaking occurs early in training as the filters compete to represent different features of the data. Eventually, the filters settle into one of an astronomical number of distinct parameterizations with nearly equivalent loss \cite{choromanska}.

Often, a highly non-convex landscape can have simple and recognizable global structure. A well-known example is the ``funnel" structure of potential energy surfaces associated with protein folding \cite{onuchic}. A funnel shape is  well-suited for guiding an unfolded or partially-folded protein to its native state. Weakly stable intermediate states might occur along the folding path, but random perturbations from the environment are enough to upset these shallow minima and allow the folding process to continue. Once the protein has reached its native state, its configuration is stable and resistant to small perturbations. The macroscopic landscape has a single global basin, despite the astronomical number of weakly stable intermediate states along the ``sides" of the funnel.

If the large-scale structure of an energy landscape is dominated by a manageable number of global basins, it should be possible to identify these energy basins and to estimate the energy barriers between them. In image landscapes, the global funnels represent the different concepts in the image patterns, since related image minima are separated by small energy barriers. Mapping only the large-scale features while ignoring local irregularities in a landscape is a key innovation of our paper. This approach distinguishes our work from previous efforts to characterize non-convex landscapes such as \cite{wales_ELM1, zhou, wales_ELM2, wales_ELM3}, which attempt to identify all local minima (or the $N$ lowest-energy minima) in the landscape, no matter how weak the basin of attraction. By focusing on macroscopic features, we define a new ELM framework that scales well with landscape dimension and/or complexity (see Section \ref{sec:AD} and Section \ref{sec:ELM}).

\subsection{Minimum Energy Path Estimation}\label{subsec:mep}
Energy barriers between local minima can be used to quantify ``closeness" of minima in the landscape, because the barriers provide a measure of the geodesic distance along the energy manifold between minima. Euclidean distance in the state space is a very poor approximation of geodesic distance. Wide, noisy basins can contain points that are far apart in Euclidean space, while two points which are nearby in Euclidean space might be separated by a large energy barrier, as in Figure \ref{fig:mag}. 

Finding energy barriers involves approximating the Minimum Energy Pathway (MEP) between the minima. The simplest approach is to find the maximum energy along the linear 1D subspace between two minima \cite{halgren}, but this often significantly overestimates the true energy barrier between points on the manifold, even over short distances (see Figure \ref{fig:meta_prac2} and Figure \ref{fig:neb_comp}). The chemical physics community has developed two major families of methods for MEP estimation. One branch of MEP methods, known as \emph{single-ended} methods, involves starting at a known local minimum and finding a transition state between minima by following the path of slowest ascent along the minimum-eigenvalue direction of the local Hessian \cite{cerjan, simons, zeng}. This method fails when Hessian information is not available or cannot be accurately approximated.

Another branch of MEP methods, called \emph{double-ended} methods, involves refining a \emph{chain-of-states} $(F_0 = X_a, F_1, \dots, F_N, F_{N+1} = X_b)$ between two minima $X_a$ and $X_b$ by minimizing the objective function
\begin{equation}
L(\{F_j \}_{j=1}^N ) = \sum_{j=1}^N E(F_j) + \sum_{j=0}^N \frac{N k}{2} || F_{j+1} - F_j ||_2^2 
\label{eqn:chain_of_states}
\end{equation}
where $E$ is the target energy and $k>0$ is a ``spring force" between successive chain states \cite{feynman, kuki}. Double-ended MEP methods require an initialization path, which by default is the 1D linear subspace between minima, since no other choices are available. Optimizing the loss (\ref{eqn:chain_of_states}) leads to misleading paths where the 1D energy barrier between successive images in the chain is significantly higher than the energy of the images in the chain. Modifications such as the Nudged Elastic Band (NEB) and Doubly-Nudged Elastic Band (DNEB) methods \cite{wales_neb} have been introduced to improve optimization by projecting energy and spring gradients onto the perpendicular and parallel components of the current path direction respectively. NEB and DNEB require numeric gradients and cannot be used in discrete spaces.

MEP methods have been successfully used to map the energy landscape of stable configurations of molecular systems \cite{wales_ELM1}. Similar methods have been applied to machine learning problems \cite{wales_ELM2, wales_ELM3}, but the results yield an overabundance of local minima and trivial, single-basin macroscopic structure. Our approach is related to the double-ended MEP methods, although we do not try to find the MEP explicitly. On the other hand, the barriers estimated by our method are often significantly lower than the barriers estimated by MEP methods (see Figure \ref{fig:neb_comp}), and our method can be used for MEP estimation in both discrete and continuous spaces. More importantly, we aim to formulate a more natural criterion the for evaluating the ``closeness" of two minima, based not on raw barrier height but on the stability of local minima under the time-evolution implied by the energy function.

\subsection{Generalized Wang-Landau Algorithm}\label{subsec:GWL}
Another approach to mapping non-convex energy functions is a version of the Generalized Wang-Landau (GWL) Algorithm \cite{wang, liang, atchade} which penalizes repeated visits to pre-defined energy bins within the basin of attraction of a local minimum. An MCMC sample is updated using the time-inhomogeneous, modified Metropolis-Hastings acceptance probability
\begin{equation}
\alpha(S \rightarrow S^\ast) = \textrm{min}\left(1, \frac{Q(S^\ast \rightarrow S)P(S^\ast)}{Q(S\rightarrow S^\ast) P(S)} \exp\left\{\gamma(N_{\varphi(S)}- N_{\varphi(S^\ast)}) \right\} \right)
\end{equation}
where $P$ is the target density, $Q$ is the transition probability, $\varphi(S)$ gives the indices $(i,j)$ of the basin $i$ and energy spectrum $j$ to which $S$ belongs, $N_{(i,j)}$ is the number of previous visits to bin $(i,j)$, and $\gamma > 0$ is a penalty for repeated visits to the same bin. 

In theory, this algorithm should result in a stationary distribution that visits each energy bin within each basin of attraction in the landscape with equal probability. Barriers between minima can be estimated by locating  and refining transition states along the MCMC path. Zhou \cite{zhou} demonstrated that the GWL algorithm can be effective in moderately-sized discrete landscapes by mapping the local minima structure of 100-dimensional SK spin-glasses. The GWL method has also been applied successfully to small-scale machine learning problems \cite{pavlovskaia}. However, the GWL Algorithm is ineffective in complex landscapes where the number of distinct local minima is too large for a full enumeration. Our new MCMC method addresses this problem by grouping minima that are separated only by small barriers, which greatly reduces the complexity of the landscape. The GWL Algorithm and our ELM method can be used together, although in the experiments presented in this paper, the GWL penalty was not necessary.

\subsection{Disconnectivity Graphs} \label{subsec:dg}
After a mapping is completed, it is useful to visually summarize the local minima and barriers that have been discovered. Visualizing all barriers between minima in a meaningful way is often an impossible task, because it is difficult to concisely represent the complex pairwise relations between the minima. \emph{Disconnectivity Graphs} \cite{becker}, or DG's, are a widely-used tool for displaying the most important features of an energy landscape. DG's reduce the complexity of the visualization task by displaying only the \emph{lowest} barrier at which two groups of minima merge in the landscape.

\begin{figure}[h]
	\centering
	\hspace*{-.5cm}
	\includegraphics[scale = .45]{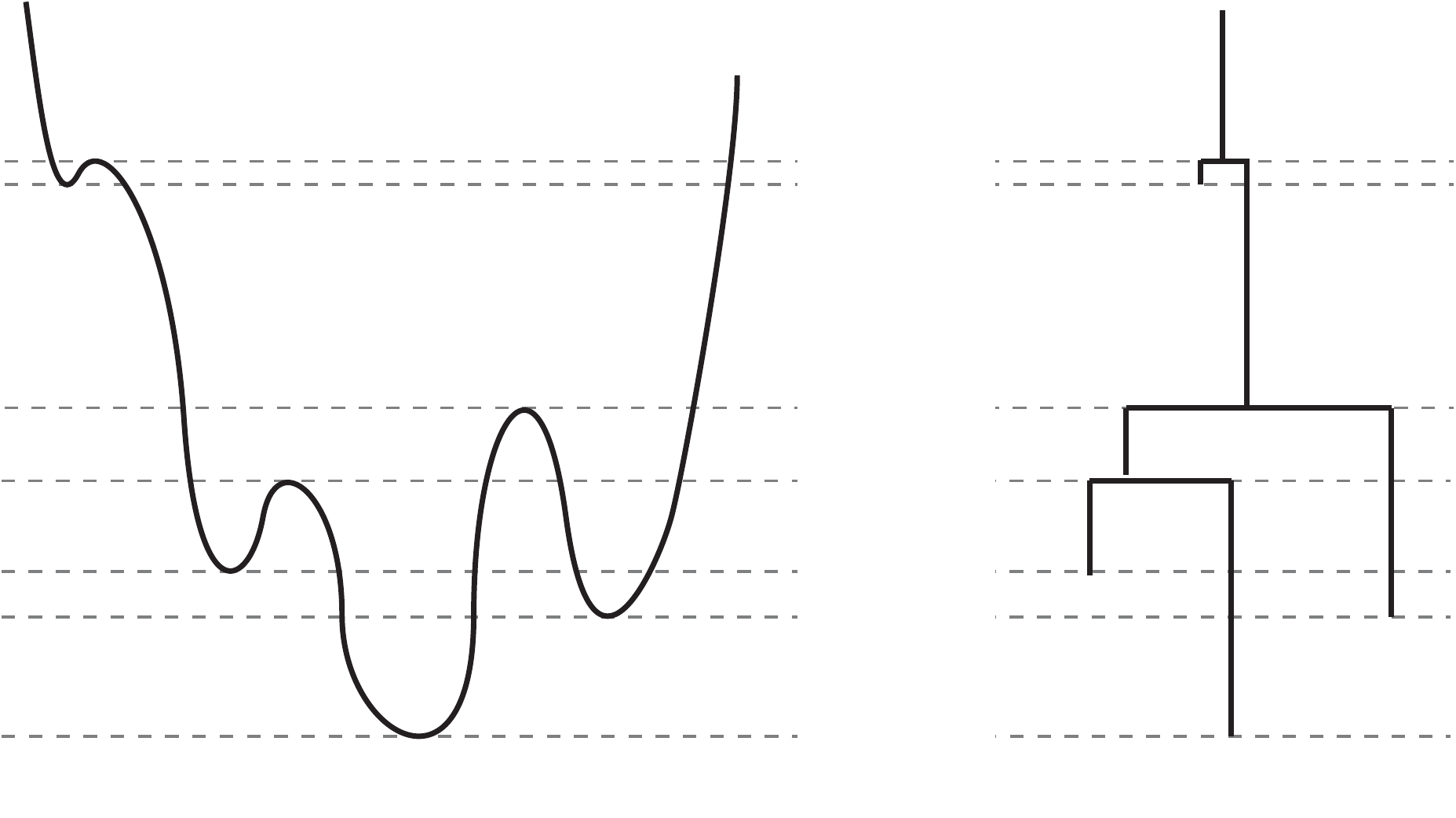}
	\caption{Illustration of Disconnectivity Graph construction. A 1D energy landscape (left) and its associated DG (right), which encodes minima depth and the lowest known barrier between basins.}
	\label{fig:dg}
\end{figure}

The leaf nodes in the DG represent local minima in the landscape, and the non-leaf nodes are placed at the lowest-energy barrier at which the basins of the child nodes merge (see Figure \ref{fig:dg}). Each child node has a single parent node, and the entire DG has a tree structure. The non-leaf nodes are often interpreted as ``superbasins" \cite{becker} of attraction which are composed of basins of attraction with similar properties. The main focus of our work is to identify superbasins of attraction without identifying all of the local minima within the superbasin.

\begin{figure}[h]
	\centering
	\hspace*{-.5cm}
	\includegraphics[width=.95\textwidth]{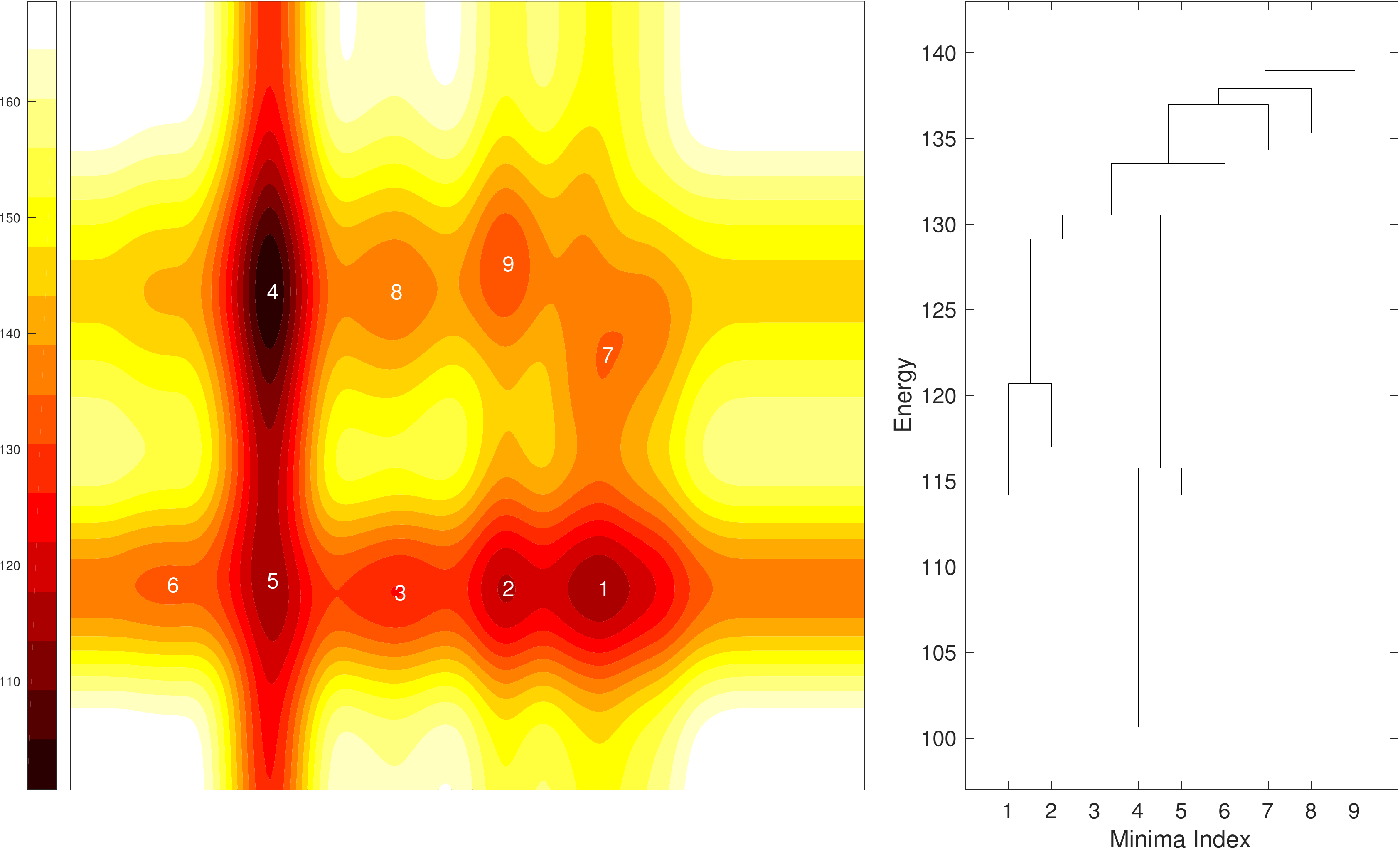}
	\caption{Landscape Visualization (left) and DG (right) of 2D landscape for GMM mean parameters.}
	\label{fig:GMM_2D}
\end{figure}

Figure \ref{fig:GMM_2D} shows a 2D landscape visualizing the loss of a Gaussian Mixture Model (GMM) as all but two mean parameters are held fixed. In this case, it is easy to see how the structure of the DG reflects the structure of the landscape, since we can visualize the loss function directly. In virtually all real cases, the landscape cannot be directly visualized or exhaustively explored via grid search, but high-dimensional landscape features can still be displayed effectively with a DG. 

A major issue with the DG visualization is the greedy nature of the branch-merging step. Merging basins at the lowest possible energy can prevent the appearance of true landscape features in the DG, because lower-energy groups of minima tend to disrupt the structure among higher-energy groups of minima. Nonetheless, DG's are a simple and often effective way of displaying the shape and connectivity of a landscape.

\subsection{Landscape Magnetization}\label{subsec:landmag}
Chaudhari and Soatto \cite{chaudhari_mag} use $t$-SNE \cite{vandermaaten} to visualize the behavior of the energy function
\begin{equation}
E^\ast (X) = E(X) + h^\intercal X \label{eqn:rand_mag}
\end{equation}
of a spin glass Hamiltonian $E$ subject to a random magnetization force $h^\intercal X$. As $|| h ||_2$ increases, the local minima structure of the magnetized landscape goes from a phase where the number of distinct minima is too large for enumeration, through a phase where a manageable number of macroscopic features emerge, to a final phase of trivialization where all minima merge into a single basin. The same behavior occurs during our mapping procedure as $\alpha$ is increased in the altered energy (\ref{eqn:AD1}). The random magnetization $h$ can be interpreted as a version of our penalty which uses a random distribution over the magnetization force $\alpha$ and target state $X^\ast$, because the Langevin Equation
\begin{equation}
dX(t) = -\left(  \nabla E(X(t)) +h  \right) dt + \sqrt{2} \, dW(t)
\end{equation}
associated with (\ref{eqn:rand_mag}) has the same dynamics as our energy (\ref{eqn:AD1}) when $\alpha = || h ||_2$ and $X^\ast = X + c h$ for any scalar $c\neq 0$. The authors only characterize the energy landscape using $t$-SNE plots, and do not attempt to systematically find basins of attraction and barriers in the landscape.

Chaudhari et al. \cite{chaudhari_grad} use a similar modification of the energy function to improve training for neural networks. The authors use an altered distribution
\begin{equation}
P_{\gamma,X^\ast}(X) = \frac{1}{Z_{\gamma,X^\ast}} \exp\left\{ - \left(E(X) + \gamma || X- X^\ast ||_2^2 \right) \right\}, \label{eqn:entropy_bias}
\end{equation}
where $X^\ast$ is the \emph{current} location and $\gamma > 0 $ is a regularization penalty, to find an entropy-biased gradient which favors movement toward wide, flat valleys in the landscape of $E$. As in our energy (\ref{eqn:AD1}), the penalty term is used to overcome local irregularities, but the interpretations and applications are very different. In \cite{chaudhari_grad}, the altered density (\ref{eqn:entropy_bias}) is used for the conventional purpose of training of network parameters, while we use the altered energy (\ref{eqn:AD1}) as a metric for metastability and as a tool for mapping landscape structure.

\subsection{Activation Maximization}\label{subsec:AM}
Our experiments on image models are closely related to the Activation Maximization (AM) field of neural network research \cite{AM1, AM2, AM3}. AM applications search for images that maximize the response of a neuron or channel in a trained network, which is equivalent to searching for local modes in the Gibbs distribution defined by neuron response. In particular, the model which we focus on in Section \ref{sec:exp} is nearly identical to the DGN-AM model \cite{DGNAM}, where a generator neural network is used to facilitate exploration of a complex neural network energy function. We learn our generator and energy network jointly using the method of Xie et al. \cite{xie_coop}, while the DGN-AM model uses separate, pre-trained generator and energy networks. 

Our work differs from the AM literature in several important ways. Previous AM works only identify a handful of local minima in the energy landscape, and do not attempt to systematically identify the basins of attraction and the structure among these basins. We show not only that neural networks can learn realistic image memories, but also that structure of image memories in the energy landscape reflects human visual intuition. AM applications generally use neurons from pre-trained classifier neural networks as the energy function, while we train our networks specifically to learn an energy function (and generator network) for a training dataset of our choice.

\section{Attraction-Diffusion}\label{sec:AD}
\subsection{Introduction to Attraction-Diffusion}\label{subsec:intro_AD}
We propose a new method for characterizing the relative stability of local minima of an energy function, which we call \emph{Attraction-Diffusion} (AD). Given an energy function $E$ and two local minima, one minima is designated as the starting location $X_0$ and the other as the target location $X^\ast$. An MCMC sample is initiated from $X_0$ using an altered density
\begin{equation}
p_{\, T, \alpha , X^\ast} (X) = \frac{1}{Z_{\, T, \alpha , X^\ast}} \, \exp\left\{-\left( E(X)/T +\alpha || X - X^\ast||_2 \right)  \right\} \label{eqn:AD_main}
\end{equation}
whose energy function is the sum of the original energy $E$ and a ``magnetization" term penalizing the distance between the current state and the target location. $T$ gives the temperature of the system, while $\alpha$ is the strength of the ``magnetic field" penalizing distance from the target minimum. The roles of starting and target location are arbitrary and diffusion in both directions is possible. The space of $X$ can be continuous or discrete.

\begin{figure}[h]
	\centering
	\includegraphics[width=0.45\textwidth]{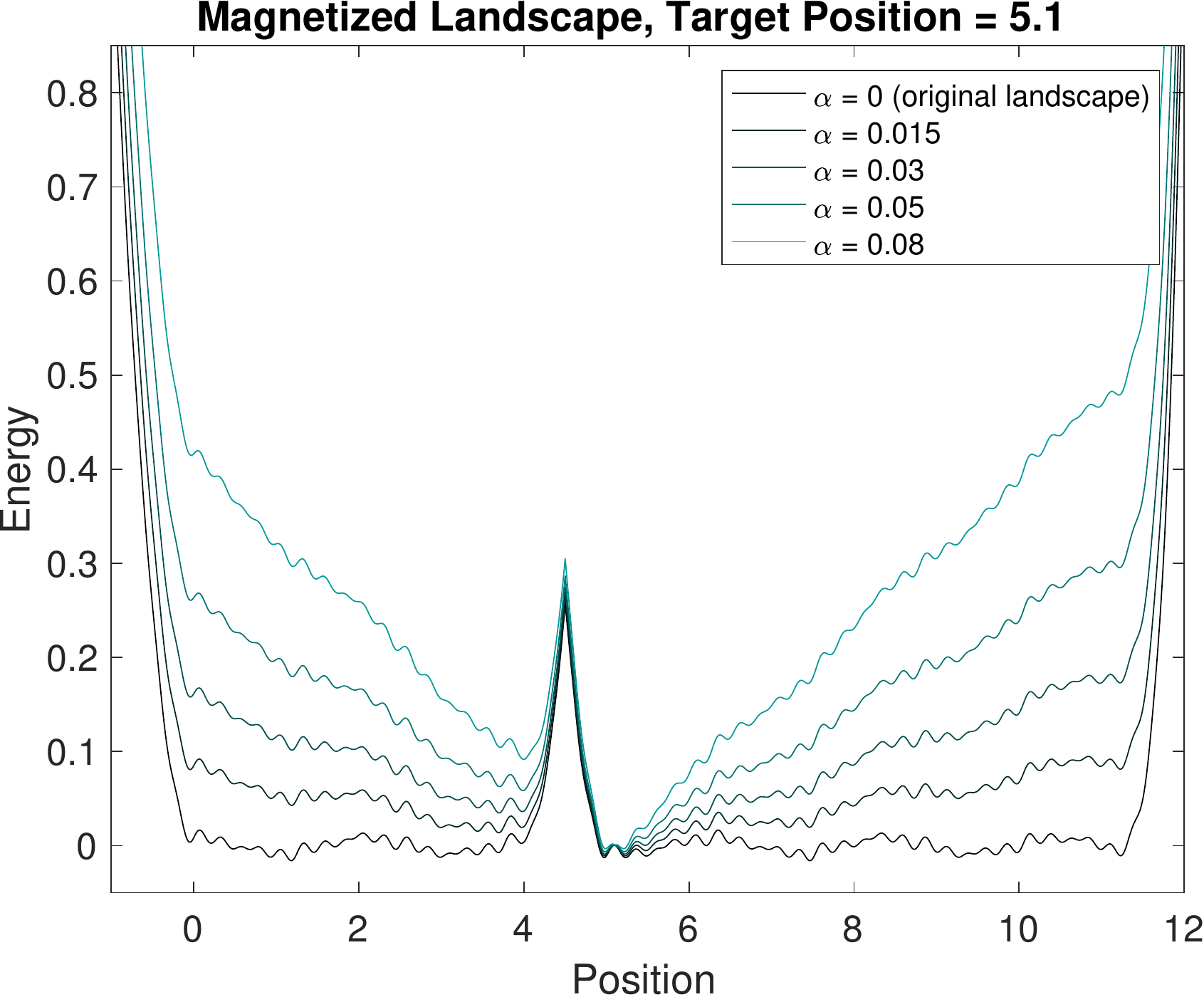}$\quad\quad$\includegraphics[width=0.45\textwidth]{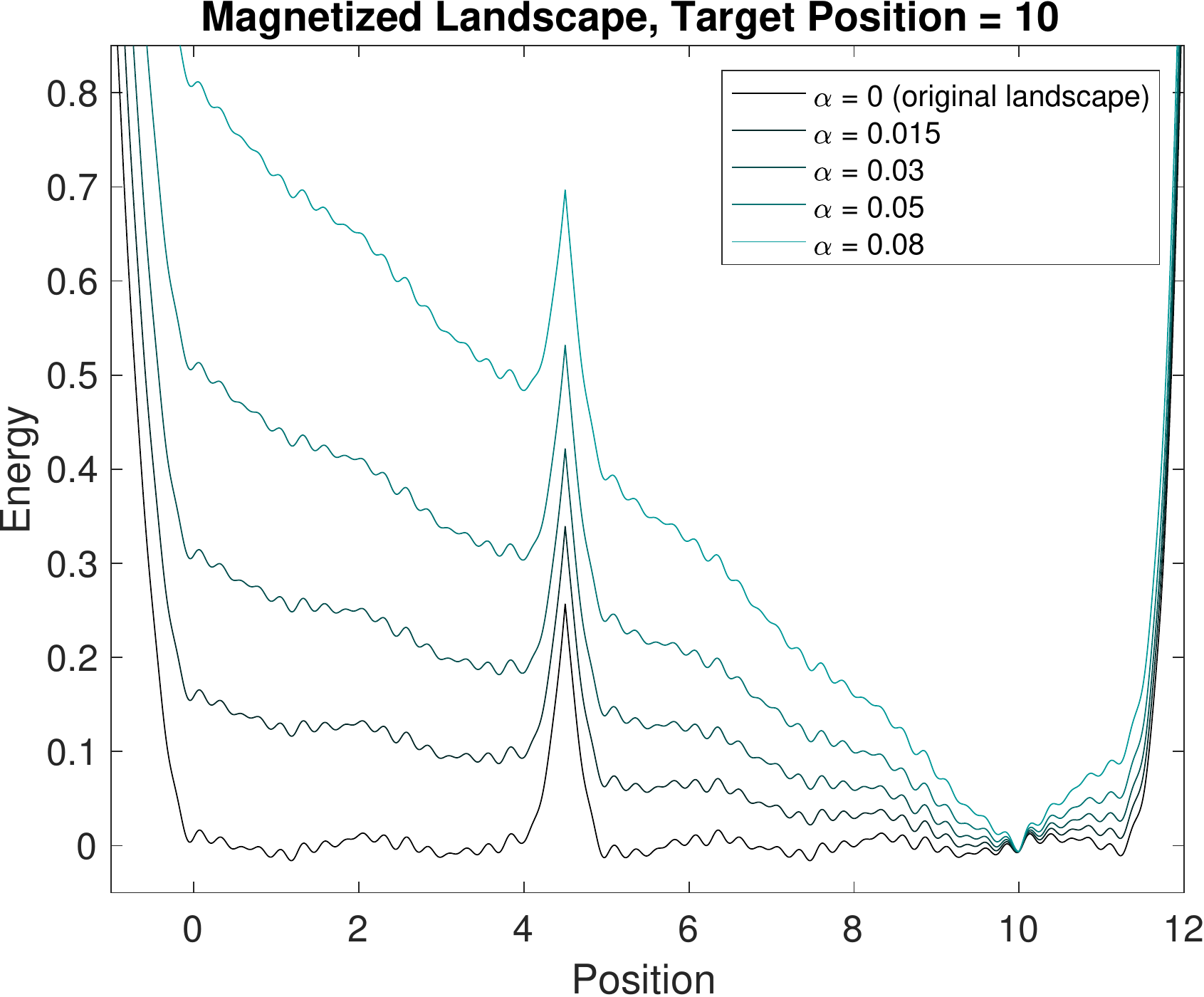}
	\caption{Magnetization of a toy 1D landscape with target positions $X=5.1$ (left) and $X=10$ (right). The original landscape has two flat and noisy basins. Both target positions belong to the same basin, even though they are distant in Euclidean space. The magnetized landscapes have easily identifiable minima, and preserve the large barrier separating the two basins. Since diffusion in the left-hand landscape from initiated from $X=10$ will reach $X=5.1$, and vice-versa in the right-hand landscape, these points belong to the same basin. Low-temperature diffusion initiated from the left of the barrier will be unable to reach the target position in either landscape.}
	\label{fig:mag}
\end{figure}

By adjusting the value of $\alpha$ and $T$, the altered landscape can be tuned so that a diffusion path can overcome small obstacles in the original landscape while remaining trapped in strong basins. If the Markov chain comes within a close distance of the target state, then the starting state belongs to the same energy basin as the target state at an energy resolution implicitly defined by the strength of magnetization. If the chain cannot improve on the minimum distance between the previous states of the chain and the target state for $M$ consecutive iterations, then there must be an energy barrier between the starting and target location that is stronger than the force of the magnetization. Figure \ref{fig:mag} demonstrates the basic principles of AD in a simple 1D landscape with two global basins.

AD can also be used to estimate the MEP and the energy barrier between minima, since the maximum energy along a successful diffusion path is an upper bound for the minimum barrier height. This estimate can be refined by setting $\alpha$ just above the threshold where the diffusion path fails to reach the target.  By using a \emph{local} MCMC method such as Random-Walk Metropolis-Hastings, Component-Wise Metropolis Hastings, Gibbs sampling, or Hamiltonian Monte Carlo \cite{neal}, one can limit the maximum Euclidean distance between points in the diffusion path and ensure that the step size is small enough so that the 1D landscape between successive images is well-behaved. An AD chain moves according to geodesic distance in the magnetized landscape, which should be similar to geodesic distance in the raw landscape as long as the strength of magnetization is not too strong.

The choice of the $L_2$-norm as the magnetization penalty is motivated by the observation that $\frac{d}{dX} ||X||_2 = X/||X||_2$, which means that the AD magnetization force points towards the target minimum with uniform strength $\alpha$ throughout the energy landscape. This can be seen in the Langevin Equation
\begin{equation}
dX(t) = -\left(  \nabla E(X(t))/T + \alpha \, \frac{X(t)- X^\ast}{|| X(t)- X^\ast ||_2} \right) dt + \sqrt{2} \, dW(t) \label{eqn:mag_langevin}
\end{equation}
associated with the magnetized dynamics. An $L_1$ penalty would probably give similar results. The penalty $\alpha ||X-X^\ast||_2^2$ would \emph{not} have desirable properties because the strength of magnetization would depend on the distance between the points, and the magnitude of alteration would vary throughout the landscape.

The magnetization term in (\ref{eqn:AD_main}) is similar to the spring term from the chain-of-states objective (\ref{eqn:chain_of_states}), except that our magnetization force is always pointing to the target minimum $X^\ast$ with uniform strength $\alpha$, while the spring force points to the next image in the chain and gets stronger when the distance between images increases. Despite the similarity in the energy functions, AD is most naturally formulated \emph{not} as a way of estimating the MEP between minima, but as a way of detecting \emph{metastability} (see Section \ref{subsec:intro_ising} and Section \ref{subsec:intro_metastable}) in an energy landscape.

\subsection{Magnetization of the Ising Model}\label{subsec:intro_ising}
The AD penalty term is closely related to the magnetization term found in energy functions from statistical physics. Consider the $N$-state magnetized Ising energy function
\begin{equation}
E_{T , H} (\sigma ) = - \frac{1}{T} \sum_{(i,j)\in \mathcal{N}} \sigma_i \sigma_j - H \sum_{i=1}^N \sigma_i 
\label{eqn:ising}
\end{equation}
where $\sigma_i = \pm 1$, $\mathcal{N}$ is the set of neighboring nodes, $T > 0$ gives the temperature, and $H$ gives the strength of an external magnetic field. This energy function is sometimes parameterized by the slightly different form $E_{T , H} (\sigma ) = \frac{1}{T}\left(-\sum \sigma_i \sigma_j - H \sum \sigma_i  \right)$, but the same properties and diagrams hold either way. The first term $- \frac{1}{T}\sum \sigma_i \sigma_j$ is the energy function of the standard Ising model, and $-H \sum \sigma_i$ represents a uniform magnetic field with strength $H$ acting on each node. When $H>0$, the field has a positive magnetization, encouraging every node to be in state +1. In this case, $E_{T, H}$  can be rewritten as
\begin{eqnarray*}
	E_{T, H}^\ast ( \sigma ) &=& E_{T, H} (\sigma ) + NH \\
	&=& - \frac{1}{T}\sum_{(i,j)\in \mathcal{N}} \sigma_i \sigma_j + H \sum_{i=1}^N (1-\sigma_i ) \\
	&=& -\frac{1}{T} \sum_{(i,j)\in \mathcal{N}} \sigma_i \sigma_j + H || \sigma - \sigma^+ ||_1 
\end{eqnarray*}
where $\sigma^+$ is the state with $\sigma_i^+ = 1$ for all nodes. The probability distribution defined by $E_{T, H}^\ast$ is the same as the distribution defined by $E_{T, H}$ because they differ only by a constant. Similarly, when $H<0$ and the magnetic field is negative, the energy function can be rewritten as
\[
E_{T, H}^\ast (\sigma) = - \frac{1}{T} \sum_{(i,j)\in \mathcal{N}} \sigma_i \sigma_j + |H|\, || \sigma - \sigma^- ||_1 
\]
where $\sigma^-$ is the state with all $\sigma_i^- = -1$. This shows that the role of $H$ in the magnetized Ising model is the same as the role of $\alpha$ in (\ref{eqn:AD_main}),  because $E_{T, H}^\ast$ is the sum of the unmagnetized Ising energy and a term that penalizes distance to either $\sigma^+$ or $\sigma^-$, the mirror global minima. Introducing the magnetization term upsets the symmetry of the standard Ising energy function and causes either $\sigma^+$ or $\sigma^-$ to become the sole global minimum, depending on the sign of $H$. 

\begin{figure}[h]
	\centering
	\hspace*{-.5cm}
	\includegraphics[width=0.4\textwidth]{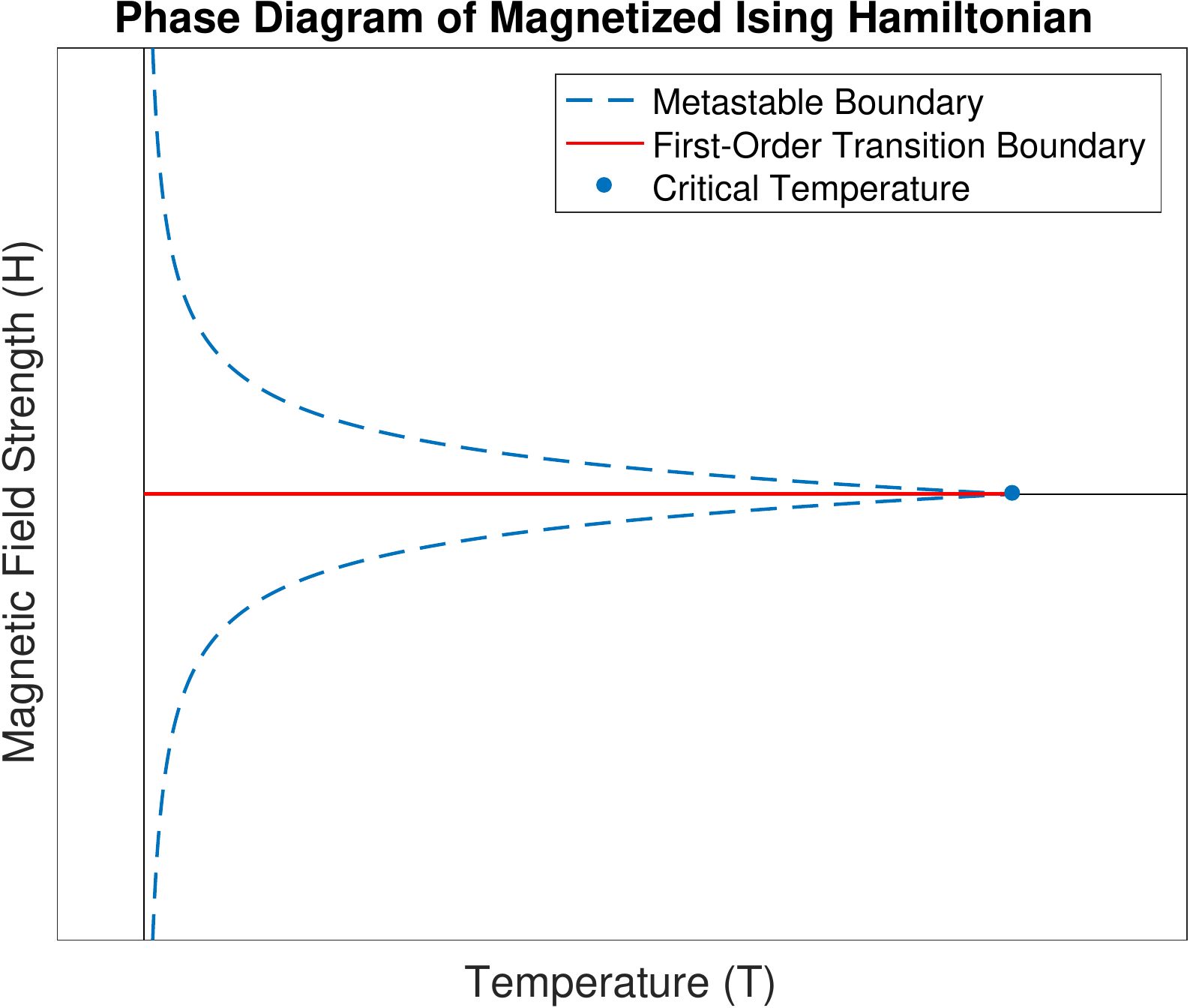}
	\hspace{0.03\textwidth}
	\includegraphics[width=0.4\textwidth]{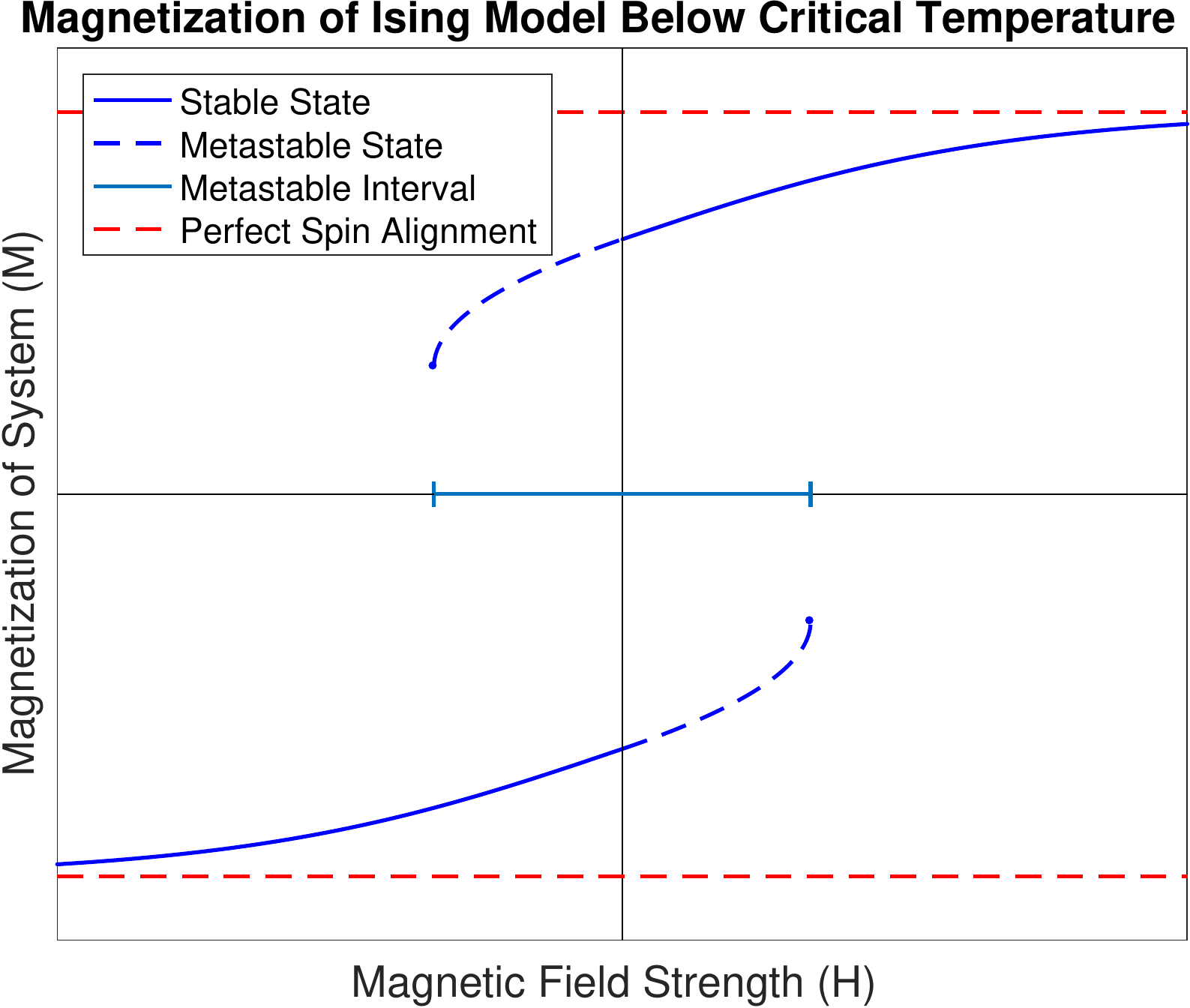}
	\caption{\emph{Left:} Phase diagram of the magnetized Ising model. Below the critical temperature, sweeping the magnetic field $H$ from positive to negative (or vice-versa) results in a jump between the basins of $\sigma^+$ and $\sigma^-$. However, if the magnetization force is weak, states in the opposite basin can remain stable for long time periods. \emph{Right:} Magnetization $M = \sum_i \sigma_i$ as a function of $H$ for a fixed $T^\ast<T_c$. The metastable interval is the region between the dashed lines along the vertical line $T=T^\ast$ in the left figure.}
	\label{fig:ising}
\end{figure}

The behavior of the system with respect to the parameters $(T, H)$ can be represented by the simple phase diagram in Figure \ref{fig:ising}. The dot is the critical temperature of the system, and the solid line is a first-order phase transition boundary. When the parameters of the system are swept across the first-order transition boundary, a discontinuous change in the state space occurs as the system flips from a predominantly positive state to a predominantly negative state, or vice-versa. On the other hand, sweeping the magnetic field $H$ across 0 above the critical temperature results in a smooth transition where positive and negative nodes coexist \cite{landau_ch2}.

Let $H> 0 $ be a weak magnetic field, and suppose the temperature $T$ is below the critical temperature $T_c$. In this situation, a phenomenon known as metastability can occur. If the system is initialized from a random configuration (each node $+1$ or $-1$ with probability 1/2), the influence of the magnetic field will cause the system to collapse to $\sigma^+$, or a nearby predominantly positive region of the state space, with high probability.  However, if the system is initialized from $\sigma^-$, and if $H$ is sufficiently small, the system will exhibit metastability, because magnetic force $H$ will be unable to overcome the strength of the bonds in $\sigma^-$, which are very strong below the critical temperature. The system will stay in a stable, predominantly negative state for a long period of time, even though the global minimum of the energy landscape is $\sigma^+$, because the magnetic field force cannot overcome the barriers between $\sigma^+$ and $\sigma^-$ in the raw Ising energy landscape \cite{landau_ch2}.

\subsection{Attraction-Diffusion and Metastability}\label{subsec:intro_metastable}
Metastability can be observed in any multi-modal energy landscape. Let $X^\ast$ be a local minimum of an energy function $E$. Even if $X^\ast$ is a shallow minimum, the temperature can be lowered so that the basin of attraction of $X^\ast$ is strong enough to trap a local diffusion process. To be more precise, for $T$ less than a critical temperature $T_{X^\ast}$, a Markov chain initialized from $X^\ast$ using a \emph{local} reversible sampling method according to the density
\[
p_T(X) = \frac{1}{Z_T} \exp\{-E(X)/T\}
\]
will remain trapped in a $\delta$-ball around $X^\ast$ for a large number of sampling iterations with high probability. The chain becomes trapped in the local mode because the behavior of MCMC is very similar to gradient descent when sampling at low temperature. The acceptance probability for proposals to higher energy regions of the landscape is virtually zero, and any movement away from $X^\ast$ has high probability of being reversed. To be considered a local sampling method, the probability of displacement in a single step of the sampler must be virtually 0 above some maximum tolerated step size $\varepsilon$ that is small relative to the scale of landscape features. Most standard MCMC methods, such as Random-Walk/Component-Wise Metropolis-Hastings, Gibbs sampling, and Hamiltonian Monte Carlo, are local, or can be tuned to be local.

Now consider two minima $X_1^\ast$ and $X_2^\ast$ and suppose $T< \textrm{min}(T_{X_1^\ast},T_{X_2^\ast})$. Since the diffusion temperature is less than the critical temperature for both minima, an MCMC sample of $p_T$ initiated from either $X^\ast_1$ or $X^\ast_2$ should remain in its original basin for a long period of time. Consider the altered density
\begin{equation}
p_{T,\alpha_1,\alpha_2} (X) = \frac{1}{Z_{T,\alpha_1,\alpha_2}} \exp \left\{ -\left( E(X)/T + \alpha_1 || X - X_1^\ast ||_2+ \alpha_2 || X - X_2^\ast ||_2 \right) \right\} \label{eqn:AD_2mins}
\end{equation}
for magnetization strengths $\alpha_1, \alpha_2 \ge 0$. 

Suppose that a sample is initialized from $X_2^\ast$ according to density $p_{T,\alpha_1,0}$ (i.e. set $\alpha_2 = 0$). If $\alpha_1$ is sufficiently small, the role of the magnetization term is negligible and the dynamics of the altered distribution are nearly identical to the original distribution. In this case, since $T< T_{X_2^\ast}$, the sample should remain trapped in the local energy basin of $X^\ast_2$ and unable to approach $X_1^\ast$ for a long period of time. On the other hand, it is clear that as $\alpha_1 \rightarrow \infty$, $X_1^\ast$ becomes the sole global minimum of the energy landscape of $p_{T,\alpha_1,0}$ and that an MCMC method initialized from $X_2^\ast$ would quickly travel to a $\delta$-ball around $X_1^\ast$ and stay within that ball indefinitely. The same properties hold when the roles of $X^\ast_1$ and $X^\ast_2$ are reversed and $\alpha_1=0$.

The above observations show that the phase space of $p_{T,\alpha_1,\alpha_2} $ with respect to the non-negative parameters $(T,\alpha_1,\alpha_2)$ in the quarter-planes $(T,\alpha_1,0)$ and $(T,0,\alpha_2)$ has properties similar to the phase space of the magnetized Ising energy $E_{T,H}$ with respect to $(T,H)$. The latter model has only two parameters because of the symmetry in the Ising model where $|| \sigma - \sigma^+||_1 = 2n - || \sigma - \sigma^- ||_1$, so the magnetization penalties for both $\sigma^+$ and $\sigma^-$ use the same parameter $H$.

An important difference between the magnetized Ising model and the AD model in a general energy landscape is the asymmetry in the stability of local minima that can occur in the latter case. Detecting asymmetry in the phase space is an essential feature of AD. When the metastable region of one minimum is significantly smaller than the metastable region of the other, this can be evidence that the former minimum belongs to a high-energy region of a large scale funnel, and that the latter minimum is located deeper within the funnel, as in the protein folding model discussed in Section \ref{subsec:complex_systems}. See Figure \ref{fig:meta_prac} for a practical demonstration of asymmetry in the AD phase space.

\begin{figure}[h]
	\centering
	\includegraphics[width=0.5\textwidth]{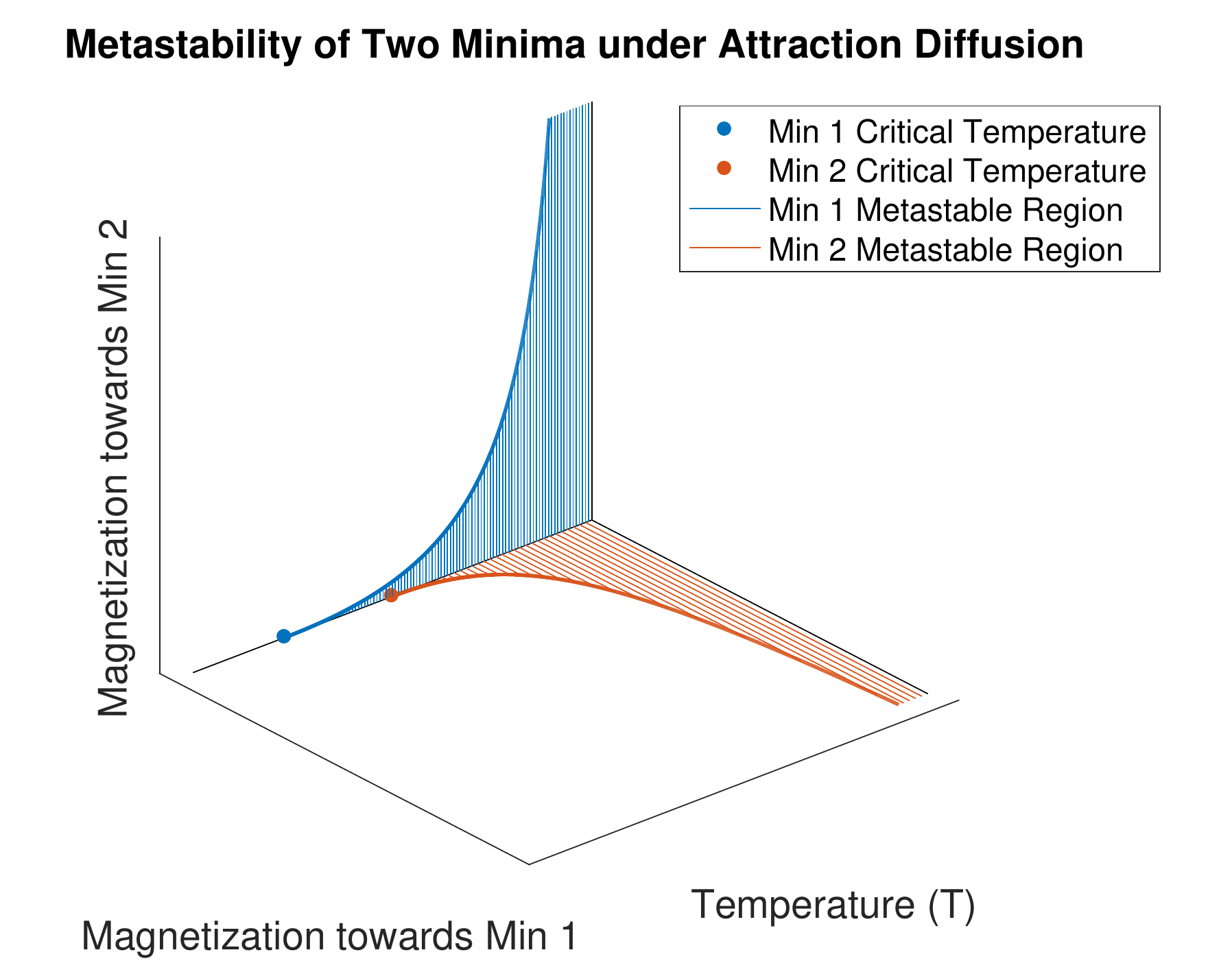}
	\caption{Metastable regions of the density (\ref{eqn:AD_2mins}) in the parameter space $(T,\alpha_1,\alpha_2)$. The system behavior in the quarter-planes $(T,\alpha_1,0)$ and $(T,0,\alpha_2)$ is similar to the upper and lower half of the Ising phase diagram Figure \ref{fig:ising}, except that the system is not symmetric. The diagram shows that Minimum 1 is more stable, because it has a higher critical temperature and larger metastable region. See Figure \ref{fig:meta_prac} for a practical example of this behavior.}
	\label{fig:meta_theory}
\end{figure}

The properties of the phase space can be analyzed with local MCMC methods. Such methods are often criticized for their tendency to become trapped in local minima, and for their inability to travel freely throughout the state space. In AD, this ``shortcoming" is exploited as a tool for measuring landscape features. When an MCMC sample of $E$ is initiated from a local mode, the correlation over time between the MCMC states and the initial local mode is an \emph{order parameter} that can be used to detect critical phenomena \cite{landau_ch2}. If temperature is low and an MCMC sample initialized from a local mode is unable to escape, the system is in an \emph{ordered} phase, and the Markov chain remains highly correlated with the local mode indefinitely (i.e. the order parameter remains non-zero). When the temperature is high enough to permit escape from the local mode, correlation with the local mode will decay quickly over time (i.e. the order parameter vanishes to 0), representing a \emph{disordered} phase. By examining whether an induced magnetization force disrupts or preserves an ordered phase, it is possible to discover landscape features. 

As discussed earlier, a major goal of the present work is to identify macroscopic landscape structures while ignoring noisy local structure. A natural way to accomplish this goal is to shift the focus of the mapping from basins of attraction under gradient descent, the standard practice in ELM applications, to regions of the landscape that are metastable under an MCMC flow, as presented in \cite{bovier}. Bovier divides the landscape into basins where the time-scale of the mixing within basins is exponentially small relative to the time-scale of mixing between basins. Local minima separated only by minor energy barriers belong to the same metastable region. This results in a simple landscape description that directly reflects the dynamics implied by the energy function. 

Unfortunately, it is not possible to identify the metastable regions of a landscape simply by initiating MCMC chains from two minima and waiting for the chains to meet, because the ``short" time-scales of mixing within basins are far too long for efficient simulation. The magnetization term in AD is meant to accelerate the short mixing time-scales within basins while still respecting the long mixing time-scales between basins. In this way, we can computationally identify the metastable regions described in \cite{bovier}, because the metastable regions of the magnetized landscape should be very similar to the metastable regions of the original landscape as long as $\alpha$ is not too strong.

In the worst case scenario, for any temperature $T$, all local minima collapse into a single mode above a threshold $\alpha_T$, while an essentially infinite number of minima can be found when the magnetization is below $\alpha_T$. However, if the energy landscape has a manageable number of macroscopic basins, there should be a critical range of $(T,\alpha)$ that will allow movement across the small noisy barriers within metastable basins while restricting movement across the large barriers between basins.

\subsection{Attraction-Diffusion in an Image Landscape}
We demonstrate the principles of AD using an energy function defined over 16$\times$16 grayscale images of the digits 0, 1, 2, and 3. In this experiment, we perform AD directly in the 256-dimensional image space. Although the images are small, the number of dimensions is quite large for an ELM application, which typically deal with landscapes of at most 100 dimensions. The energy function and minima are taken from the first experiment in Section \ref{subsec:exp_gen} -- Minimum A is Minimum 5, Minimum B is Minimum 4, and Minimum C belongs to the group represented by Minimum A.

\begin{figure}[h]
	\centering
	\medskip
	\begin{tabular}{c|c|c}
		Min A & Min B & Min C \\
		\hline \includegraphics[scale = 2]{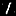} & \includegraphics[scale = 2]{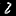} & \includegraphics[scale = 2]{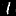} 
	\end{tabular}
	\medskip
	\includegraphics[width=0.3\textwidth]{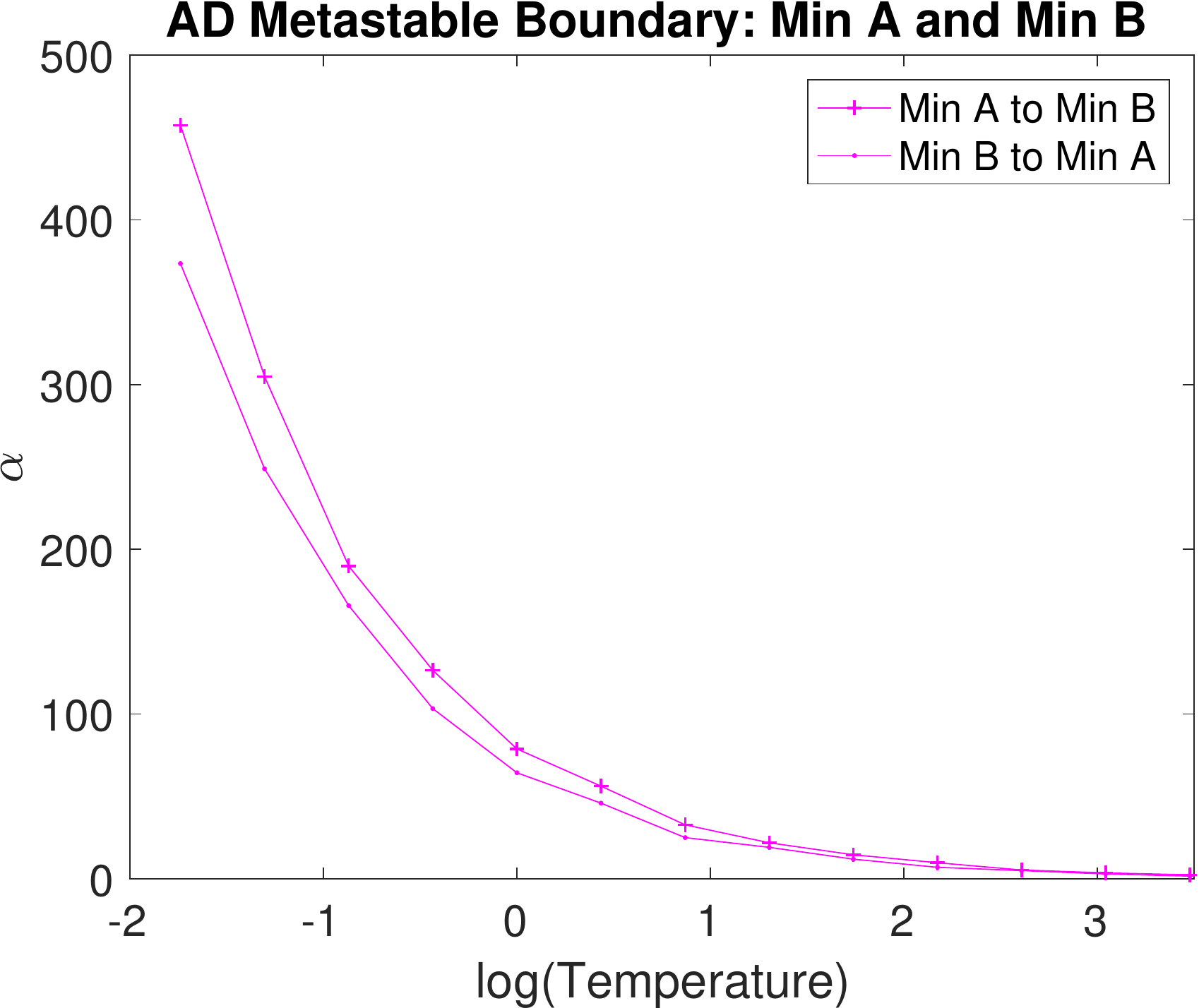}\includegraphics[width=0.3\textwidth]{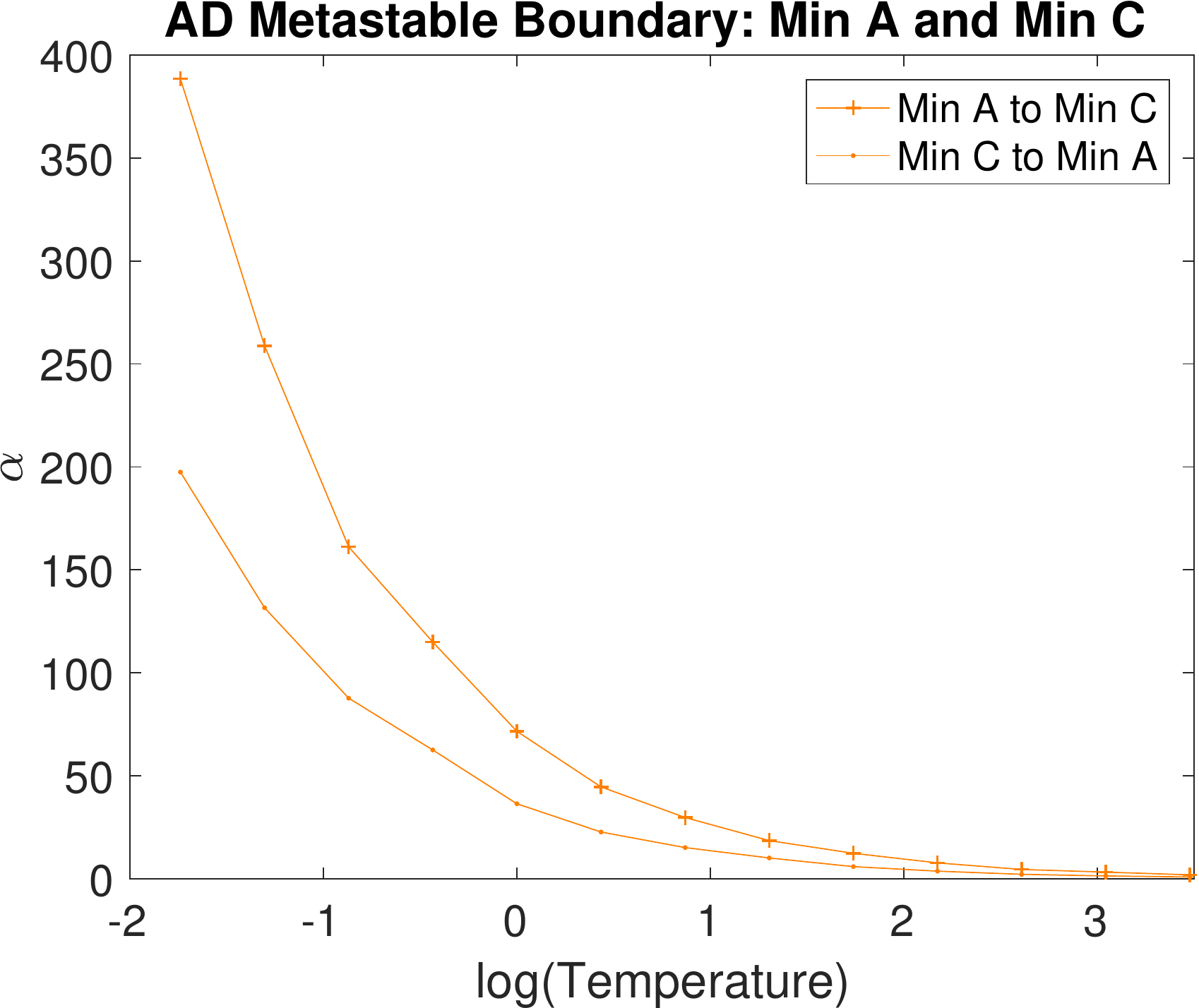}\includegraphics[width=0.3\textwidth]{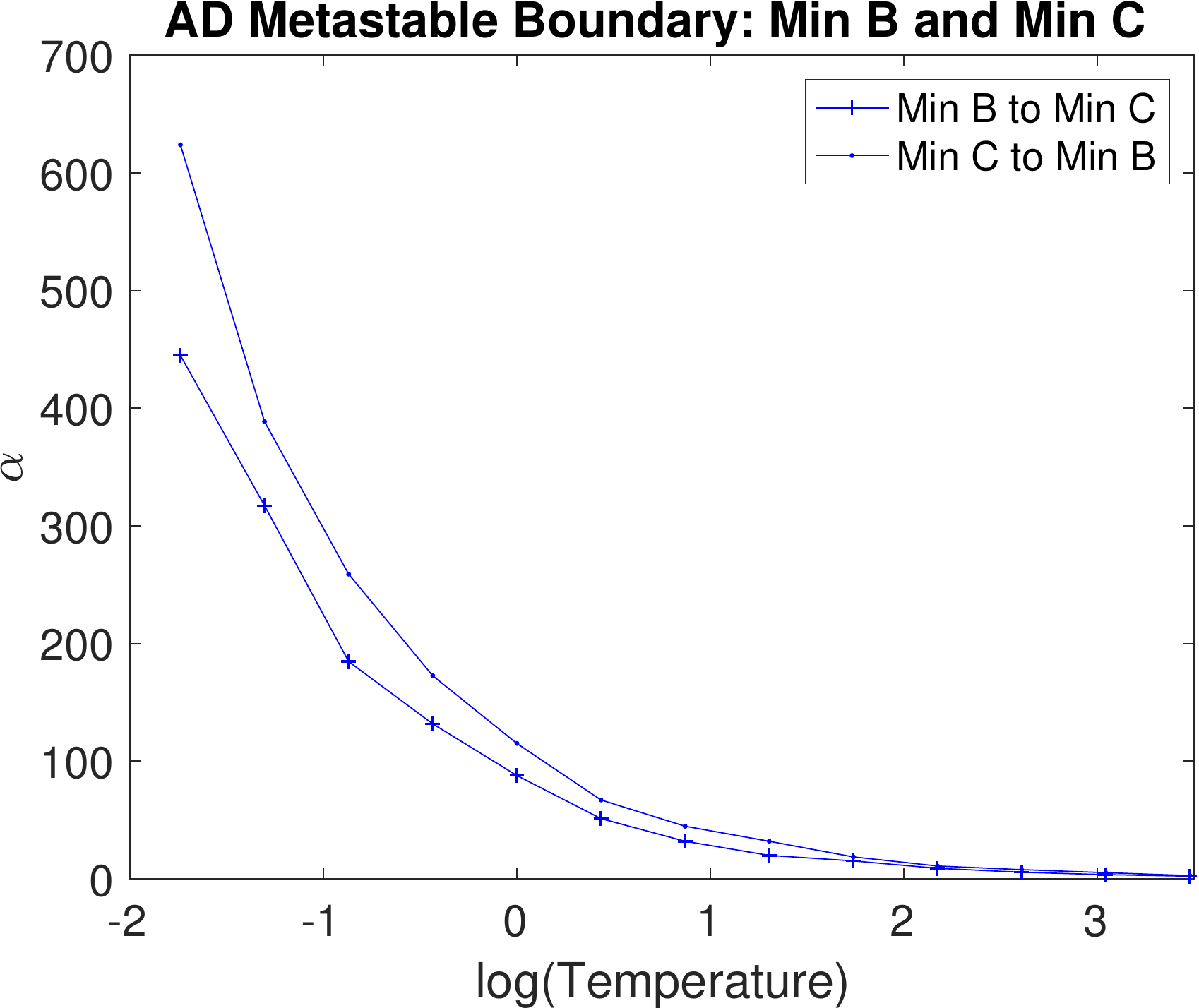} \\
	\includegraphics[height=5.2cm]{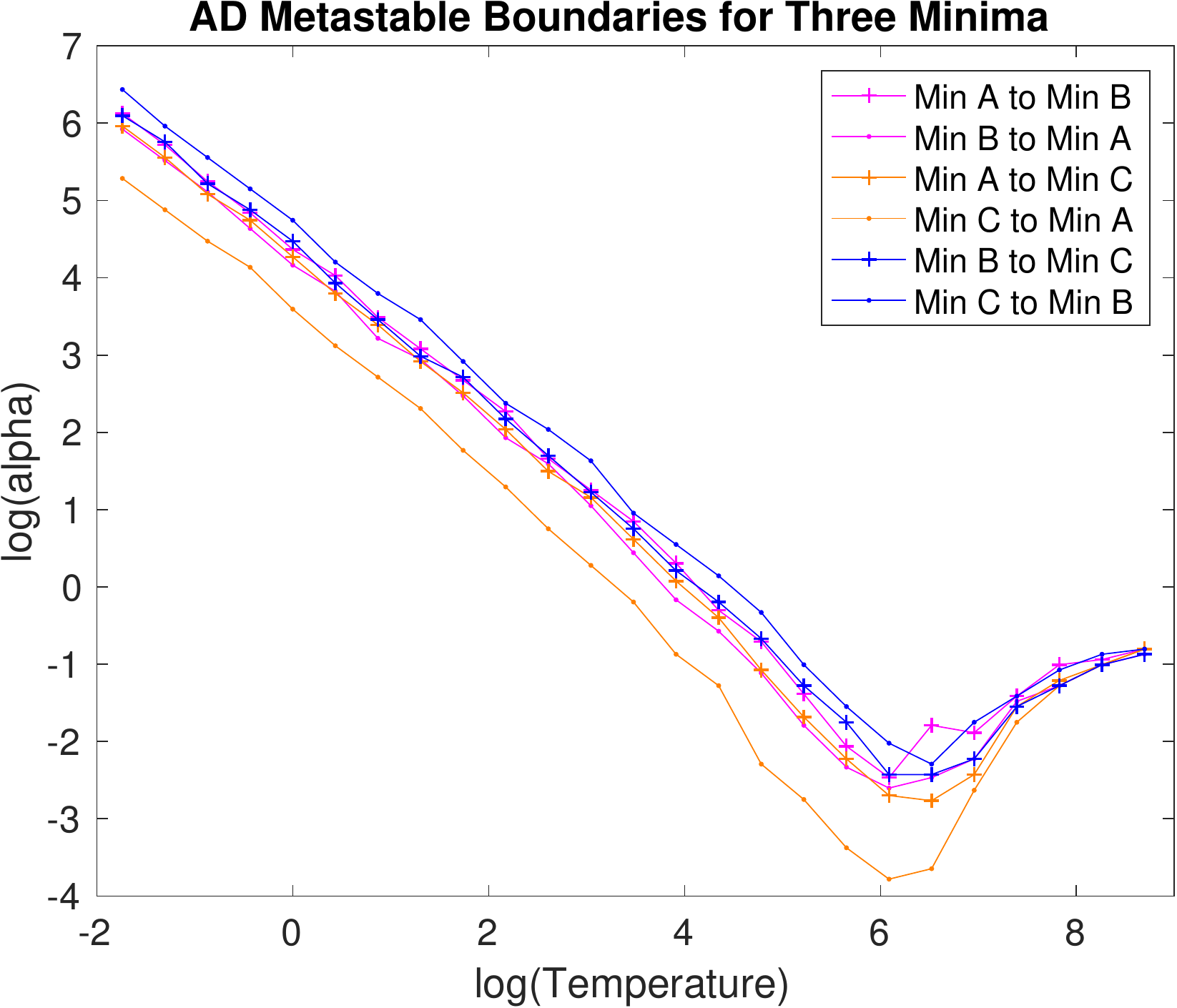}$\quad\;$\includegraphics[height=5.2cm]{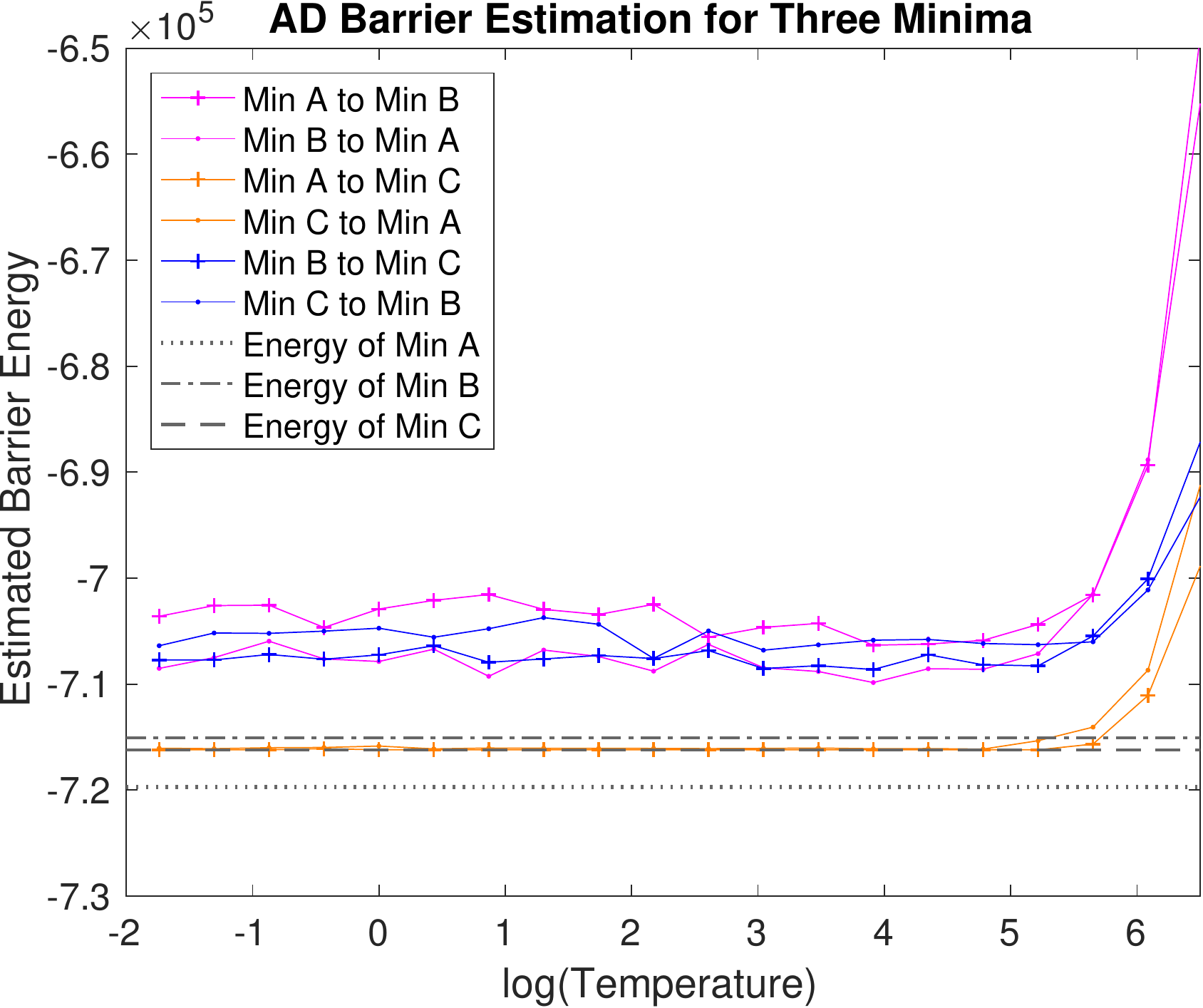}
	\caption{\emph{Top:} The three minima tested. \emph{Middle:} Metastable regions for the minima pairs AB, AC, and BC respectively. These plots are a superimposition of the two planes from Figure \ref{fig:meta_theory}. \emph{Bottom Left:} Comparison of metastable boundaries. Min C merges with Min A at a low $\alpha$, while the other minima merge at around the same energy level. The relation is approximately linear, and the upward turn reveals the critical temperature. \emph{Bottom Right:} Barrier estimates across $T$.}
	\label{fig:meta_prac}
\end{figure}

The metastable regions of each minima pairing in the parameter space $(T,\alpha)$ can be mapped using AD, and the results are similar to the phase space of the magnetized Ising function, as described in Section \ref{subsec:intro_ising}. We used an improvement limit $M = 20$ (one Gibbs sweep is a single iteration) and distance resolution $\delta=150$ (each pixel has a value from 0 to 255, so this resolution is quite strict). For a range of temperatures spaced evenly on log scale, we estimated the metastable threshold of $\alpha$ by searching for the point where diffusion just failed to reach the target. We started at a high value of $\alpha$, and attempted 20 AD trials for each pairing. If any of these trials were successful, we decreased the value of $\alpha$ by 3\% and ran another 20 trials, and repeated until none of the trials were successful. The minimum energy barrier found during the search was recorded. The minima played both roles in each pairing, so there were 6 tests in total. The plots, shown in Figure \ref{fig:meta_prac}, validate the AD principles discussed in Section \ref{subsec:intro_metastable} and are evidence that the autocorrelation of an MCMC sample can be used as a reliable metric for metastable phenomena in an energy landscape.

Figure \ref{fig:meta_prac} also gives an idea of how AD can be used to group minima. The plots show that Minimum C collapses to Minimum A in a region of the parameter space where the other minima are highly stable. Moreover, the barrier found along the AD path between Minimum A and Minimum C is almost 0, despite the fact that the minima are distant in Euclidean space and are separated by an energy barrier along the 1D interpolation path. This is evidence that Minimum C is located along the side of a ``funnel" of the energy basin represented by Minimum A, much like an intermediate state in protein folding.

\begin{figure}[h]
	\centering
	\begin{tabular}{cc}
		\includegraphics[width=0.45\textwidth]{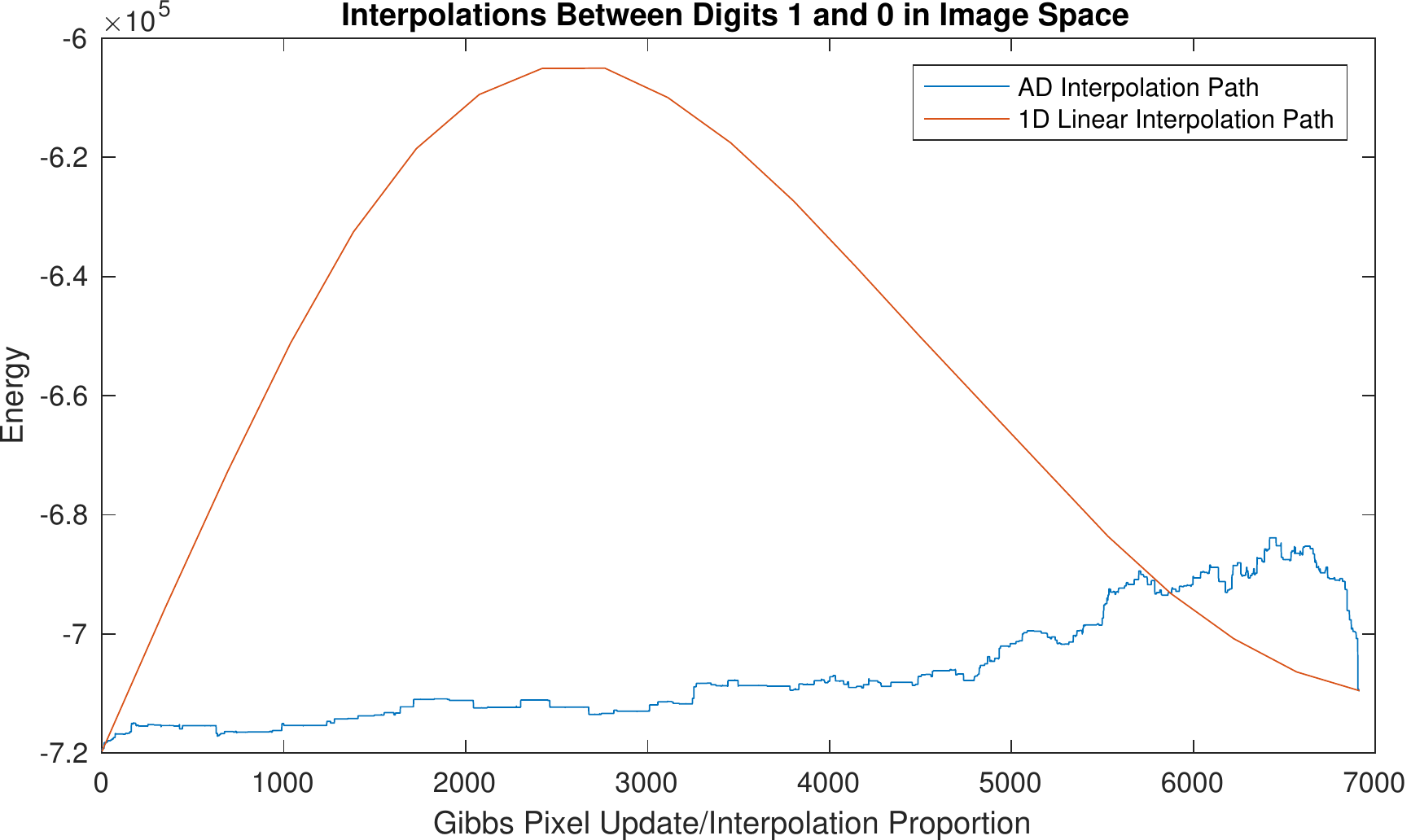} & \includegraphics[width=0.45\textwidth]{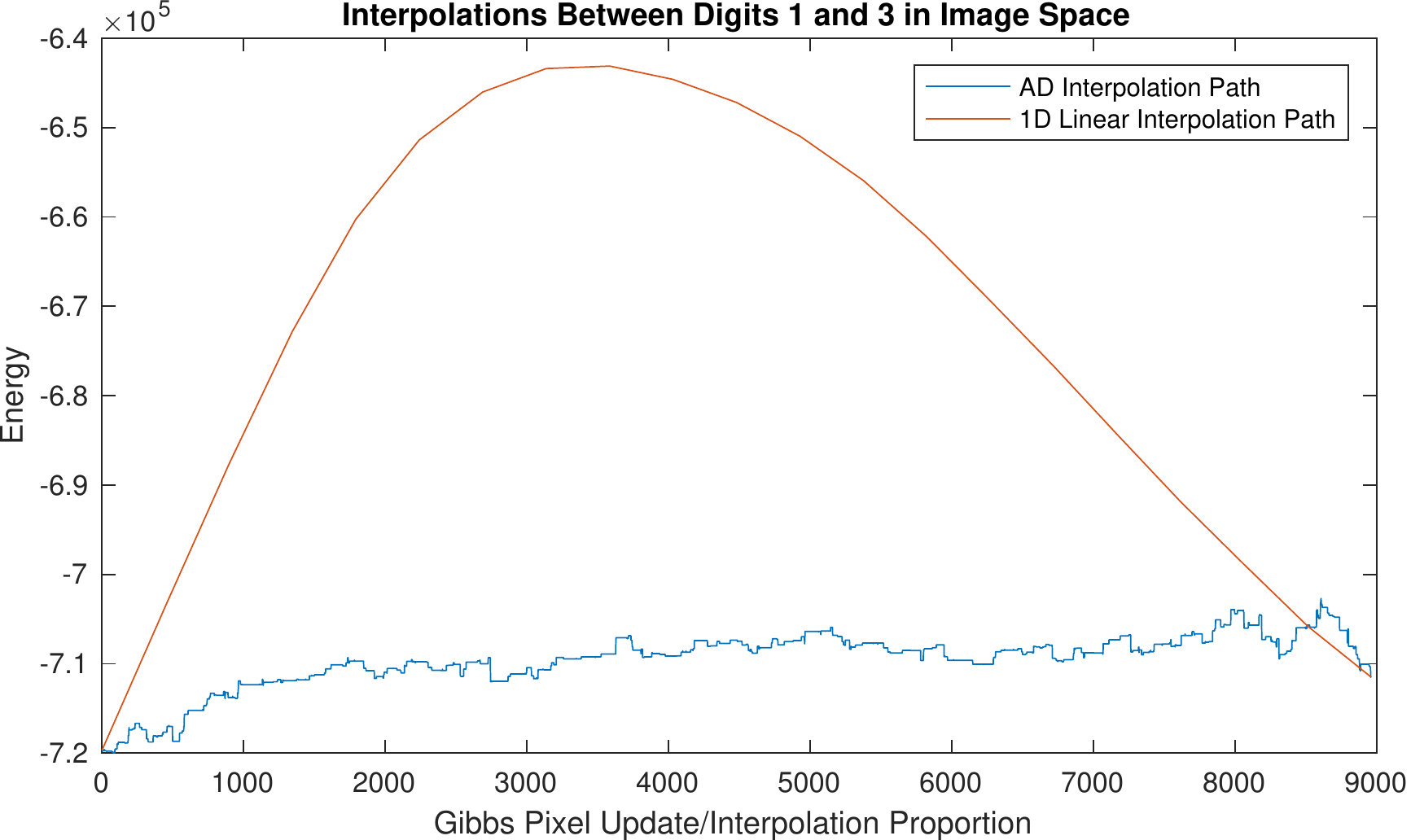}\\
		\includegraphics[width=0.45\textwidth]{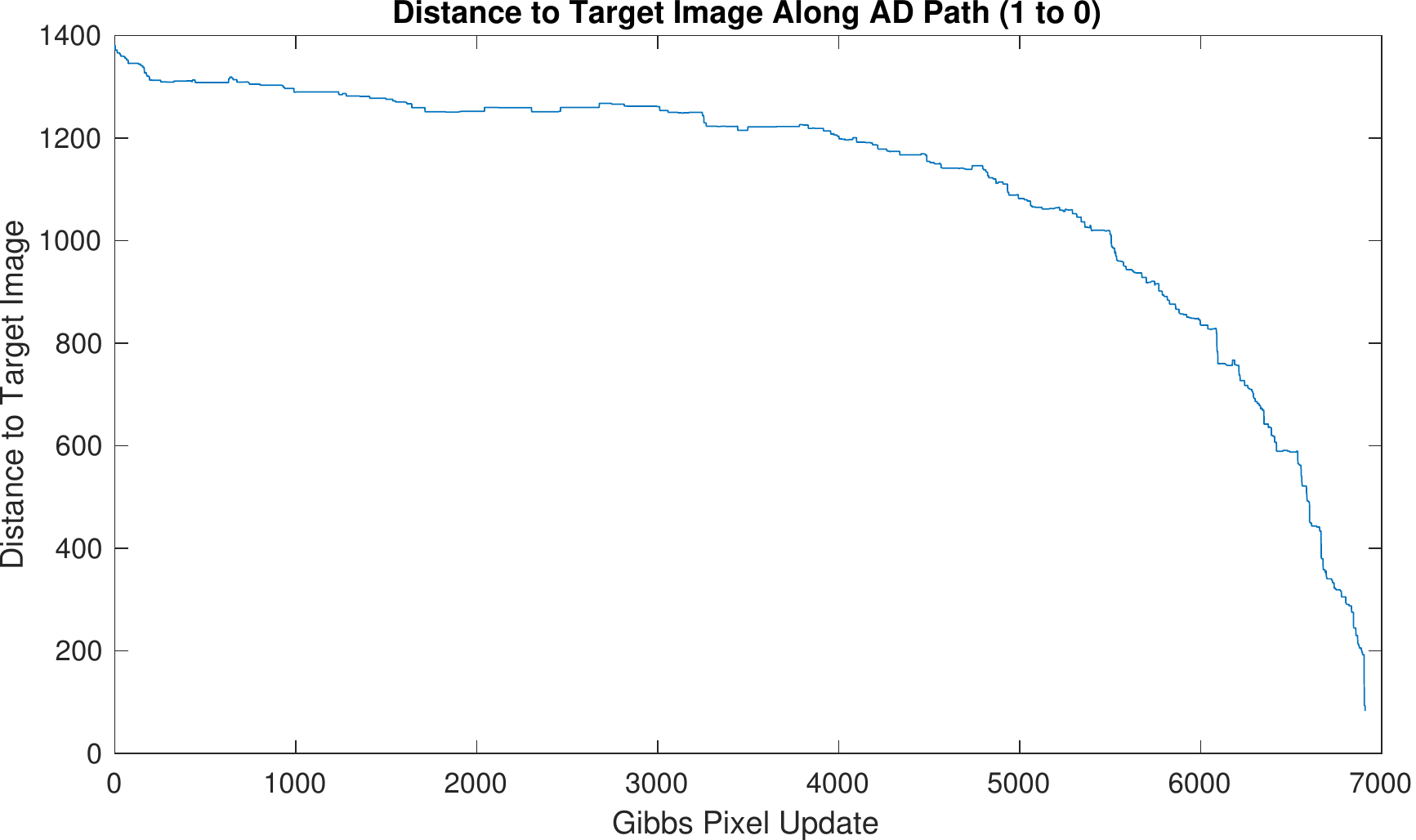}& \includegraphics[width=0.45\textwidth]{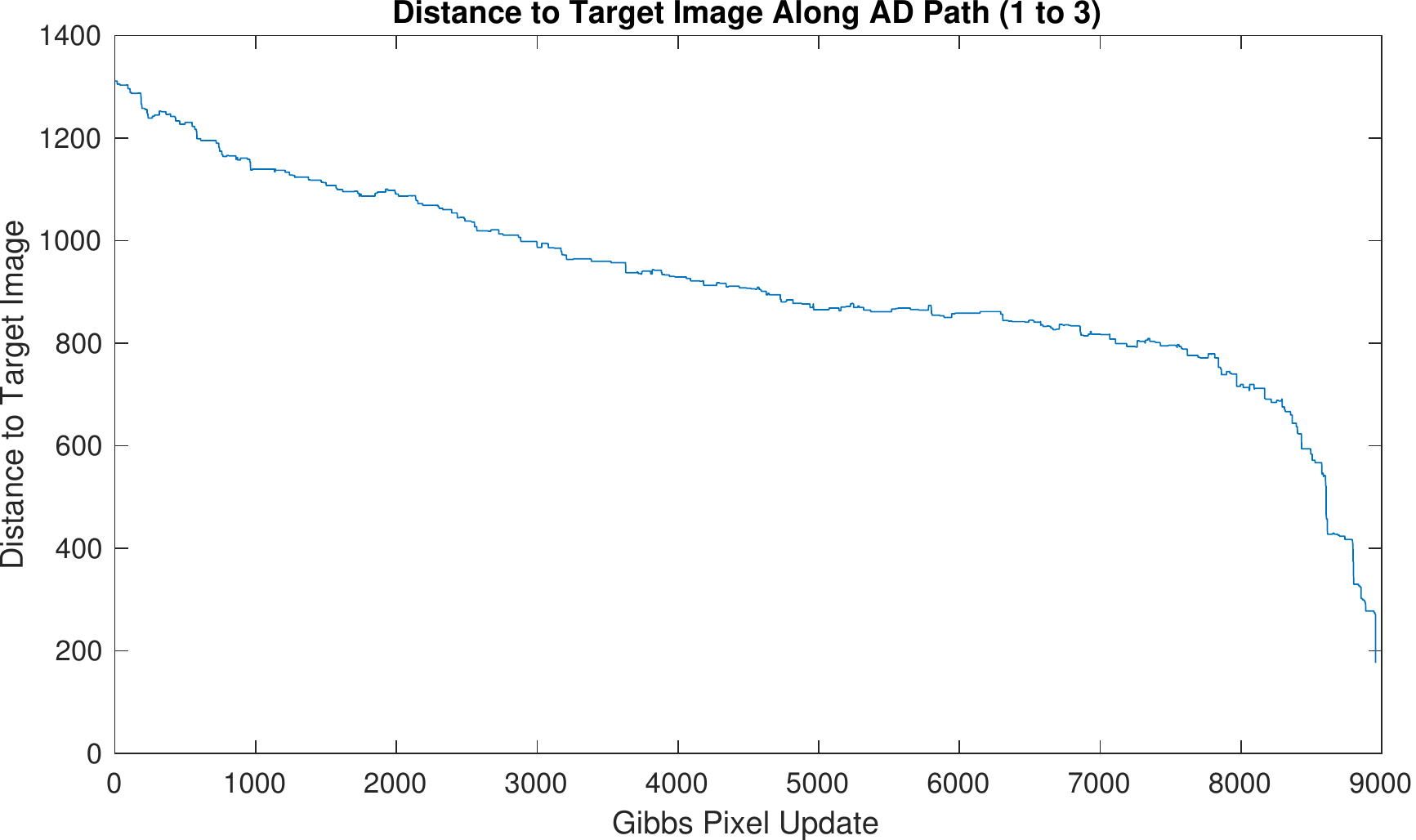}\\
		\includegraphics[width=0.48\textwidth]{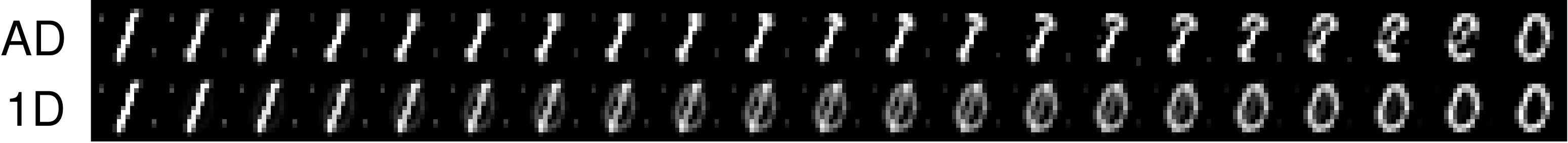}&
		\includegraphics[width=0.48\textwidth]{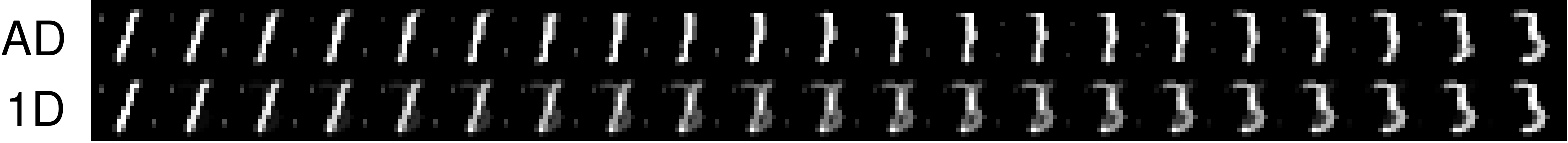}
	\end{tabular}
	\medskip
	\caption{\emph{Top:} 1D Interpolation Barrier vs. AD Barrier for diffusion from digit 1 to digit 0, and from digit 1 to digit 3. The AD barriers are much lower, and the AD paths are quite flat. \emph{Middle:} Distance from target minima vs. Gibbs Sweep. \emph{Bottom:} Visualization of interpolations. The AD paths are able to move along the image manifold using only an energy network.}
	\label{fig:meta_prac2}
\end{figure}

AD can also be used as a method for estimating the MEP between minima. When finding MEP estimates, it is best to run AD chains below critical temperature using a magnetization $\alpha$ that is just above the metastable boundary. Running AD chains at or above critical temperature yields poor results because the chains will not be restricted to the lowest-energy regions of the landscape. When $\alpha$ is too strong, the interpolations will be very close to the 1D linear interpolation, because the chain will ignore landscape features and simply travel straight to the target. When $\alpha$ is too low, the chain will never reach the target and no barrier estimate can be obtained. In a small critical region above the metastable boundary, the magnetization force and energy features have equal magnitude and jointly encourage the chain to travel to the target while respecting landscape structure.

Figure \ref{fig:meta_prac2} shows interpolations performed in a 16$\times$16 image space using the energy network from the first experiment in Section \ref{subsec:exp_gen}. The red curve gives the barrier along the 1D linear path between minima in the image space, while the blue curve shows the energy of a successful AD path between the minima. The barriers estimated by AD are drastic reductions of the 1D estimates. Visualizing the images in the AD path shows that the chains diffuse along the image manifold to find non-linear interpolations using only an energy function. AD can also be used to refine pathways in a latent space of a generator network, as we show in Figure \ref{fig:neb_comp}.

\section{Mapping the Energy Landscape Using Attraction-Diffusion}\label{sec:ELM}
\subsection{Three Essential Steps of ELM}\label{subsec:ELM_intro}
ELM methods have three basic exploration steps:

\begin{center}
	\begin{enumerate}
		\item Get a state $X$ as the starting point for a minima search.
		\item Find a local minimum $Y$ starting from $X$.
		\item Determine if $Y$ is grouped with a previously found minima basin or if $Y$ starts a new minima basin. 
	\end{enumerate}
\end{center}

\noindent These steps are repeated until no new local minima are found for a certain number of iterations. After the local minima are identified, the barriers between the minima are estimated. 

Step 2 can be accomplished with standard gradient descent methods, and the GWL Algorithm provides a principled way to propose $X$ in Step 1. Previous ELM methods lack a reliable way to tackle Step 3. Traditionally, ELM studies have attempted to enumerate \emph{all} basins of attraction of the energy landscape (or the $N$ lowest-energy minima), no matter how shallow \cite{wales_ELM1, zhou, wales_ELM2, wales_ELM3}. Minima are only grouped together if they are identical in discrete spaces, or if they are extremely close in continuous spaces. This approach is doomed to failure in all but the simplest cases, because the number of distinct local minima grows exponentially with landscape complexity and/or dimension. On the other hand, for some families of energy functions, the \emph{macroscopic} structure might remain unchanged as landscape complexity/dimension increases. For example, the Ising energy landscape will always have two global basins, regardless of neighborhood structure or number of nodes.

Instead of dividing up the state space according to basins of attraction under gradient flow, we follow the approach of Bovier \cite{bovier} and divide the state space according to disjoint regions which are metastable under the flow induced by a reversible MCMC process. This results in a much simpler description of the landscape, because the metastable regions will merge basins of attraction which are only separated by small barriers. If the magnetization $\alpha$ used in AD is weak, the metastable regions of the altered landscape should roughly correspond to the metastable regions of the original landscape, and the success or failure of an AD trial can be used as an indicator of membership in a given metastable region. Mapping regions that are metastable under an MCMC process rather than basins of attraction under gradient flow is essential for the success of ELM in complex landscapes.

\subsection{Attraction-Diffusion ELM Algorithm}\label{subsec:ELM_ADELM}
We now present an Attraction-Diffusion Energy Landscape Mapping (ADELM) Algorithm. Steps 1 and 2 do not involve AD and the implementation details are left open-ended.

\small{
	\begin{center}
		\begin{algorithm}[H]
			\SetKwInOut{Input}{input} \SetKwInOut{Output}{output}
			\DontPrintSemicolon
			\Input{Target energy $E$, local MCMC sampler $S$, temperature $T>0$, magnetization force $\alpha>0$, distance resolution $\delta>0$, improvement limit $M$, number of iterations $N$}
			\Output{States $\{X_1, \, \dots , \, X_N\}$ with local minima $\{Y_1, \, \dots , \, Y_N\}$, minima group labels $\{l_1, \, \dots , \, l_N\}$, and group global minima $\{Z_1, \, \dots , \, Z_L\}$, where $L = \textrm{max} \{ l_n\} $}
			\;
			\For{$n = 1:N$}{
				
				\smallskip
				
				1. Get proposal state $X_n$ for minima search. (Random initialization, or a GWL MCMC proposal)\;
				2. Start a local minimum search from $X_n$ and find a local minimum $Y_n$.\;
				3. \If{$n=1$}{ 
					Set $Z_1 = Y_1$ and $l_1=1$.} 
				\Else{
					Determine if $Y_n$ can be grouped with a known group using AD. Let $L_n = \textrm{max}\{l_1, \, \dots, \, l_{n-1} \}$, and let minimum group membership set $G_n = \emptyset$. \;
					
					\smallskip
					
					\For{$j= 1:L_n$}{
						
						\smallskip
						
						a) Set $C = Y_n$, $X^\ast = Z_j$, $d_1 = || C - X^\ast ||_2$, $d^\ast = d_1$, and $m=0$. \;
						\While{$(d_1 > \delta) \; \& \; (m <  M)$}{
							
							\smallskip
							
							\noindent Update $C$ with a single step of sampler $S$ using the density
							\[
							P(X) = \frac{1}{Z} \, \exp\{ -(E(X)/T + \alpha || X - X^\ast || _2 )\}
							\]
							\noindent and find the new distance to the target minimum: $d_1 \leftarrow || C - X^\ast ||_2$.\;
							
							$\;$  \textbf{If} {$d_1 \ge d^\ast$} \textbf{then} {$m \leftarrow m+1 $}, \textbf{else} {$m \leftarrow 0$ and $d^\ast \leftarrow d_1$.}\;
						}\;
						
						b) Set $C = Z_j$, $X^\ast = Y_n$, $d_2 = || C - X^\ast ||_2$, $d^\ast = d_1$, and $m=0$, and repeat the loop in Step a).\;
						
						c) If $d_1 \le \delta$ or $d_2 \le \delta$, then add $j$ to the set $G_n$, and let $B_j$ be the barrier along the successful path. If both paths are successful, let $B_j$ be the smaller of the two barriers.\;
					}
					\If{$G_n$ is empty}{
						$Y_n$ starts a new minima group. Set $l_n = \textrm{max}\{l_1, \, \dots, \, l_{n-1} \} +1 $, and $Z_{l_n} = Y_n$.
					}
					\Else{
						$Y_n$ belongs to a previous minima group. Set $l_n = \textrm{argmin}_j B_j $.\;
						\If{$E(Y_n) < E(Z_{l_n})$}{Update the group global minimum: $Z_{l_n} \leftarrow Y_n$.}
					}
				}
			}
			\caption{Attraction-Diffusion ELM (ADELM)}
		\end{algorithm}
	\end{center}
}

The MCMC sampler $S$ should be local in the sense that displacement after a single step is small relative to landscape features with high probability. MCMC methods with step size parameter $\varepsilon$ such as Metropolis-Hastings with a Gaussian proposal or HMC/Langevin Dynamics are local samplers, since $\varepsilon$ can be tuned to limit displacement. Gibbs sampling is also local, because only a single dimension is changed in each update. The requirement that $S$ is local is needed to ensure that a Markov chain updated using $S$ cannot escape from local modes at low temperatures. Usually, this is considered an undesirable feature of MCMC methods, but in AD it is essential that the Markov samples remain trapped in the absence of magnetization. Upsetting this baseline behavior by introducing a magnetic field enables the discovery of landscape features.

In the ADELM algorithm, the global minima $Z_j$ of each basin are used as the targets for AD trials. One reason for this choice is the intuition that, for the same strength $\alpha$, an AD chain should be more likely to successfully travel from a higher-energy minimum to a lower-energy minimum than vice-versa. While not true in general, in practice the intuition holds in most cases, especially for very deep minima. A more nuanced implementation could consider multiple candidates from the same basin as targets for diffusion instead of just the global minimum.

Correct tuning of $T$ and $\alpha$ is essential for good results. The temperature $T$ must be set low enough so that movement is restricted to the current mode, but not so low that the chain becomes totally frozen. The magnetization strength $\alpha$ must be strong enough to overcome the noisy shallow barriers in the landscape while respecting the large-scale barriers. One simple method of tuning $T$ and $\alpha$ requires two minima $X$ and $Y$ which are believed to be in separate basins. By perturbing the minima, two relatives $X'$ and $Y'$ can be found which should be in the same basin as $X$ and $Y$ respectively. The parameters $T$ and $\alpha$ can then be tuned so that only diffusion between the original and perturbed copies is successful. Figure \ref{fig:meta_prac} shows that the behavior of AD is quite consistent across a range of $T$ below the critical temperature. Choosing $\alpha$ seems to be the most important tuning decision.

Ideally, in each step of the ADELM Algorithm, diffusion to only one basin representative $Z_j$ should be successful. Successful diffusion to a large number of previously found basins is a sign of poor tuning --- in particular, either the value of $T$ or $\alpha$ (or both) is too high, causing leakage between basins. On the other hand, some leakage between minima is usually inevitable, because there are often plateau regions that sit between stronger global basins. This is not too much of a problem as long as the basin representatives remain separated. The global basin representatives $\{Z_j\}$ should be checked periodically to make sure they remain well-separated at the current parameter setting. If an AD chain successfully travels between two of the $\{Z_j\}$, these minima should be consolidated into a single group. This is especially important in the early stages of mapping, when good basin representatives have not yet been found. A single basin can split into multiple groups if the early representatives are not effective attractor states for the entire basin. When consolidating minima, the lower-energy minimum is kept as the group representative. 

The ADELM algorithm has two computational bottlenecks: the local minima search in Step~2, and the AD grouping in Step~3. The computational cost of Step~2 is unavoidable for any ELM method, and the MCMC sampling in Step 3 is not unreasonable as long as it has a comparable running time. In our experiments, we find that the running time for local minimum search and a single AD trial are about the same. Step 3 of the ADELM algorithm involves AD trials between a new minimum and several known candidates, and the efficiency of ADELM can be greatly increased by running the AD trials in parallel.

\subsection{Barrier Estimation and Landscape Visualization}\label{subsec:ELM_barrier}
AD can be used to estimate the energy barriers and the MEP between local minima after exploration is over. This is done by fixing the temperature $T$ and tuning $\alpha$ to find a threshold where successful travel between minima is just barely possible. The AD barrier estimates are lowest when $\alpha$ is just above the metastable border in the AD phase space, and will increase as $\alpha$ increases. In the limit $\alpha \rightarrow \infty$, the AD barriers are identical to the 1D linear barriers, because the MCMC samples will simply move in a straight line towards the target. Estimated barrier height appears consistent for a range of $T$ below critical temperature, as in Figure \ref{fig:meta_prac}. In our mappings, we are primarily interested in the energy barriers between the global basin representatives, which are the most significant features of the macroscopic landscape. 

Disconnectivity graphs, or DG's (see Section \ref{subsec:dg} and Figure \ref{fig:dg}), have been used in many previous ELM studies as a method for visualizing the energy landscape. Construction of a DG is straightforward once the minima have been identified by ADELM and the barriers have been estimated by running AD trials between the basin representatives. Our ELM visualizations introduce two new elements to the standard DG format. First, we draw circles around the minima nodes of the DG, sized in proportion to the number of local minima sorted into the corresponding global basin. Second, when mapping image landscapes, we display the global basin representatives in a row at the bottom of the DG. Above the basin representatives, we display randomly selected examples of minima images sorted into each basin, sorted from top to bottom in order of decreasing energy. See Figure \ref{fig:elm_0123_des} for an example.

\section{Experiments} \label{sec:exp}
\subsection{Mapping an SK Spin Glass} \label{subsec:exp_SK}
In our first ADELM experiment, we map the structure of a sample from the 100-state SK spin glass model. The $N$-state SK spin glass is a generalization of the standard $N$-state Ising model where the coefficients for couplings unspecified. The energy function for the $N$-state SK spin glass is 
\begin{equation}
E (\sigma ) = - \frac{1}{T} \sum_{1\le i < k \le N} J_{ik} \, \sigma_i \sigma_k 
\label{eqn:SK_glass}
\end{equation}
where $\sigma_i = \pm 1$, $T>0$ is the temperature, and $J_{ik}$ are couplings. In standard Ising model, the coupling coefficients are either 1 (i.e. the nodes are adjacent) or 0 (i.e. the nodes are not adjacent). The energy landscape of an SK spin glass contains multiple well-separated global basins that have noisy local structure. Like the Ising model, the landscape is exactly symmetric, since $E(\sigma) = E(-\sigma)$.

Computationally mapping the local minima structure of an SK spin glass is a challenging task, because exhaustive search of the state space is infeasible for $N>30$, and the landscape structure is highly non-convex. Zhou \cite{zhou} has shown that the GWL algorithm can accurately identify the lowest-energy minima and barriers for as many as $N=100$ states. Mapping a 100-dimensional SK spin glass is a good setting for validating our ADELM algorithm because the results of our mapping can be compared with the results of a GWL mapping, which are very close to the ground truth. The symmetry of SK spin glass landscapes is also useful for evaluating our method, because we can compare the mappings of the mirror basins.

\begin{figure}[h]
	\centering
	\includegraphics[width=\textwidth]{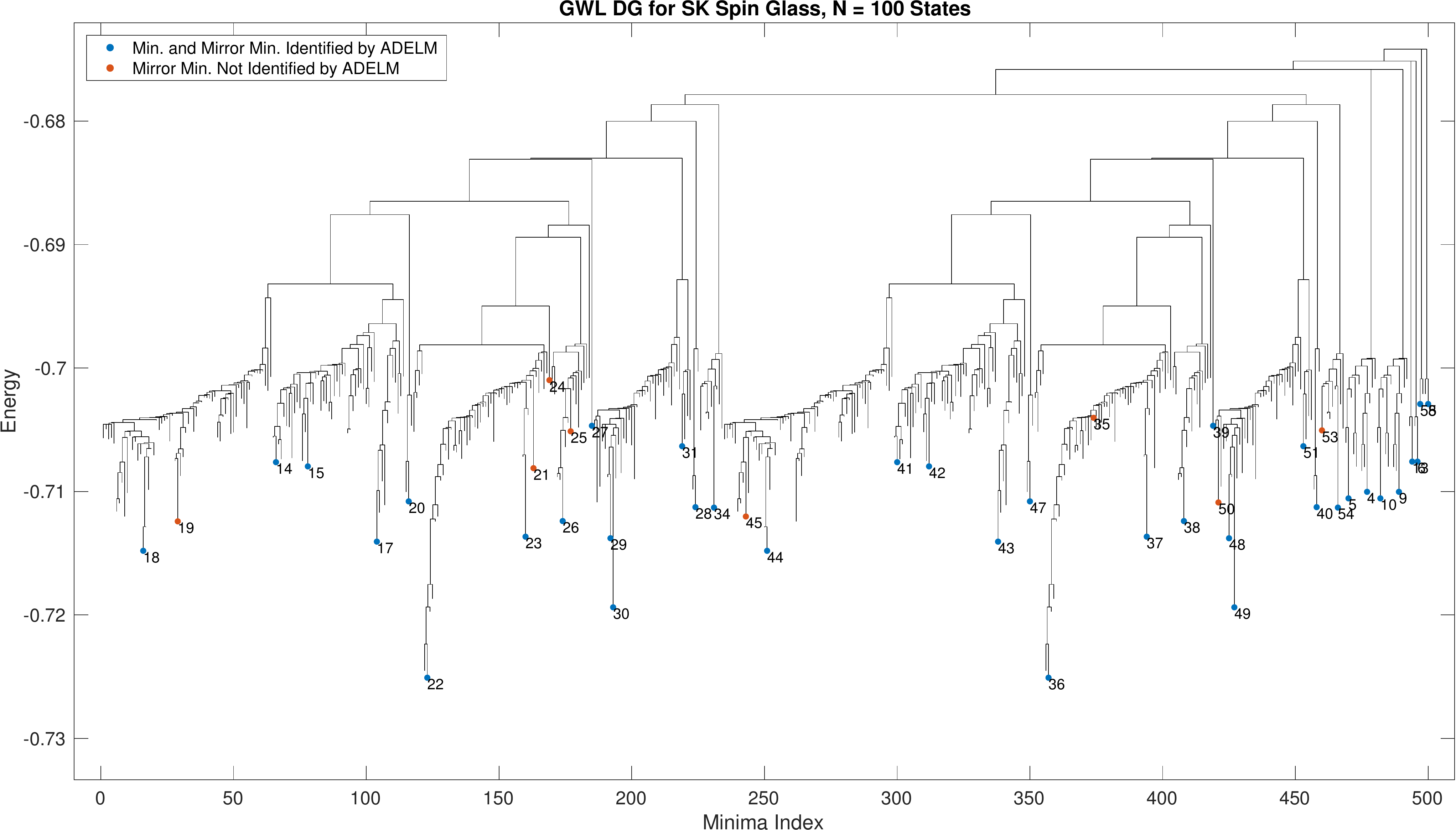}
	\caption{GWL DG for SK Spin Glass. The map records the 500 lowest energy minima in the landscape and the tree is nearly symmetric. The blue and orange dots indicate basins that were identified by our ADELM mapping. The blue dots show minima whose mirror state was also identified during the ADELM run, while the orange dots show minima whose mirror state was not identified. The blue dots cover the most important features of the landscape, which are very stable for the AD parameters $(T=0.1,\alpha = 1.35)$, while the orange dots record substructures within stronger basins that are close to the metastable border.}
	\label{fig:SK_GWL}
\end{figure}

We replicated the GWL mappings in \cite{zhou}, and the result is shown in Figure \ref{fig:SK_GWL}. The couplings $J_{ik}$ are independent Gaussians with mean 0 and variance $1/N$, as in the original experiment. We ran our mapping for $5\times 10^8$ iterations using the same GWL parameters described in the original paper, and searched for the 500 lowest minima in the landscape. The number of local minima in an SK spin glass is far more than 500 even with only $N=100$ states, but previous mappings show that the 500 lowest-energy local minima capture the main landscape features. In more complex landscapes or larger spin glasses, even the lowest-energy regions can contain an astronomical number of local minima, making the GWL approach problematic.

After running the GWL mapping, the 500 lowest minima identified were exactly symmetric, meaning that for each minima discovered we also identified its mirror state as a minima in the mapping. We used two methods to estimate the barriers between minima, and recorded the lower result as the energy barrier. The first method is the one described in \cite{zhou}, which involves identifying transitions between minima basins along the GWL MCMC path and refining these transition states by ridge descent to identify the barrier between the minima.  

The second method is a greedy algorithm for interpolation in discrete spaces where we change a starting state to a target state by iteratively choosing, among the spins differing from the target state, the change that causes either the smallest increase or the greatest decrease in energy. Suppose $\sigma$ and $\tau$ are two states, and let $\mathcal{I} = \{ i : \sigma_i \neq \tau_i \}$. Let 
\[
\sigma^{(i)}_j = \left\{ \begin{matrix} \sigma_j \quad \textrm{if } j \neq i \\  \tau_j \quad \textrm{if } j = i \end{matrix} \right.
\]
 for $1 \le j \le N$, and $i^\ast = \textrm{argmin}_{i\in \mathcal{I}} \; E(\sigma^{(i)}) - E(\sigma)$. Update the state $\sigma \leftarrow \sigma^{(i^\ast)}$ and repeat until $\sigma = \tau$. This procedure is not symmetric, so the roles of $\sigma$ and $\tau$ should also be reversed, and the lower barrier of the two paths recorded.

In nearly all cases, the barriers estimated by the second method were significantly lower than the barriers estimated by the first method. Even with the GWL penalty, most MCMC crossings between basins occur well above the minimum energy barrier that separates the basins. This is corroborated by the observation that the GWL mapping exhibited very poor mixing when we changed the energy spectrum from [-0.8, -0.35], as in the original experiment, to [-0.8, -0.55], which is still well above the maximum barrier between any of the lowest 500 minima. It appears that the global basins of the SK spin-glass model influence the energy landscape in regions that have significantly higher energy than the energy barrier at which the basins merge, and we encounter the same behavior in our other ELM experiments. In this case, it is more appropriate to describe the landscape in terms of metastability, as we are doing in ADELM, rather than barrier height between basins, because the barrier along the MEP is not representative of the energy level at which a diffusion process is affected by a basin of attraction.

We mapped the same energy landscape using ADELM to compare results and to see if ADELM can reliably identify the most important features of the landscape. We used the temperature $T=0.1$, which is well below the critical temperature $T_c = 1$, and magnetization strength $\alpha = 1.35$ as the AD parameters. Setting $T$ exactly at the critical temperature yielded poor results, because the energy fluctuation of the chains in the absence of magnetization was greater than the depth of the landscape features, and a colder system is needed to restrict diffusion to the lowest energy levels. We ran our algorithm for 5,000 iterations, set the AD improvement limit to $M=100$ Gibbs sweeps of all states, and set our distance resolution $\delta = 0$, which requires that AD chains travel exactly to their target for a successful trial. 

\begin{figure}[h]
	\centering
	\includegraphics[width=\textwidth]{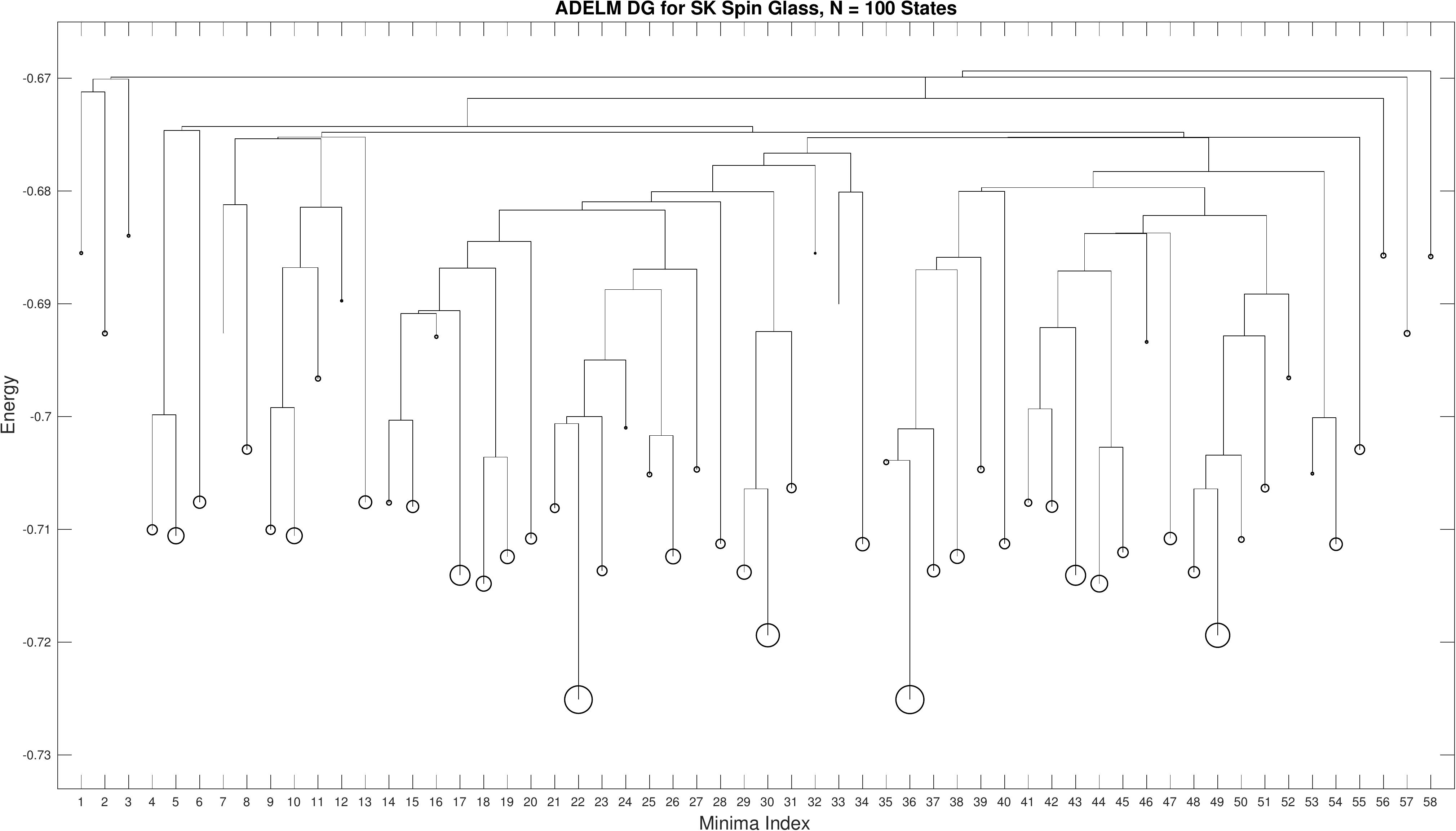}
	\caption{AD DG for SK Spin Glass. The AD diagram is quite symmetric (see Figure \ref{fig:SK_GWL}) and the structure of the DG is very consistent with the DG created from the GWL mapping (see Figure \ref{fig:ising_overlay}). 44 of the AD minima are also located by GWL, while 14 of the ADELM minima are not among the 500 lowest energy minima. The GWL mapping, which records only lowest-energy minima, misses significant stable features in higher-energy regions. The size of circles around minima nodes is proportional to the number of minima sorted to each basin, as described in Section \ref{subsec:ELM_barrier}. }
	\label{fig:SK_AD}
\end{figure}

\begin{figure}[h]
	\centering
	\includegraphics[width=\textwidth]{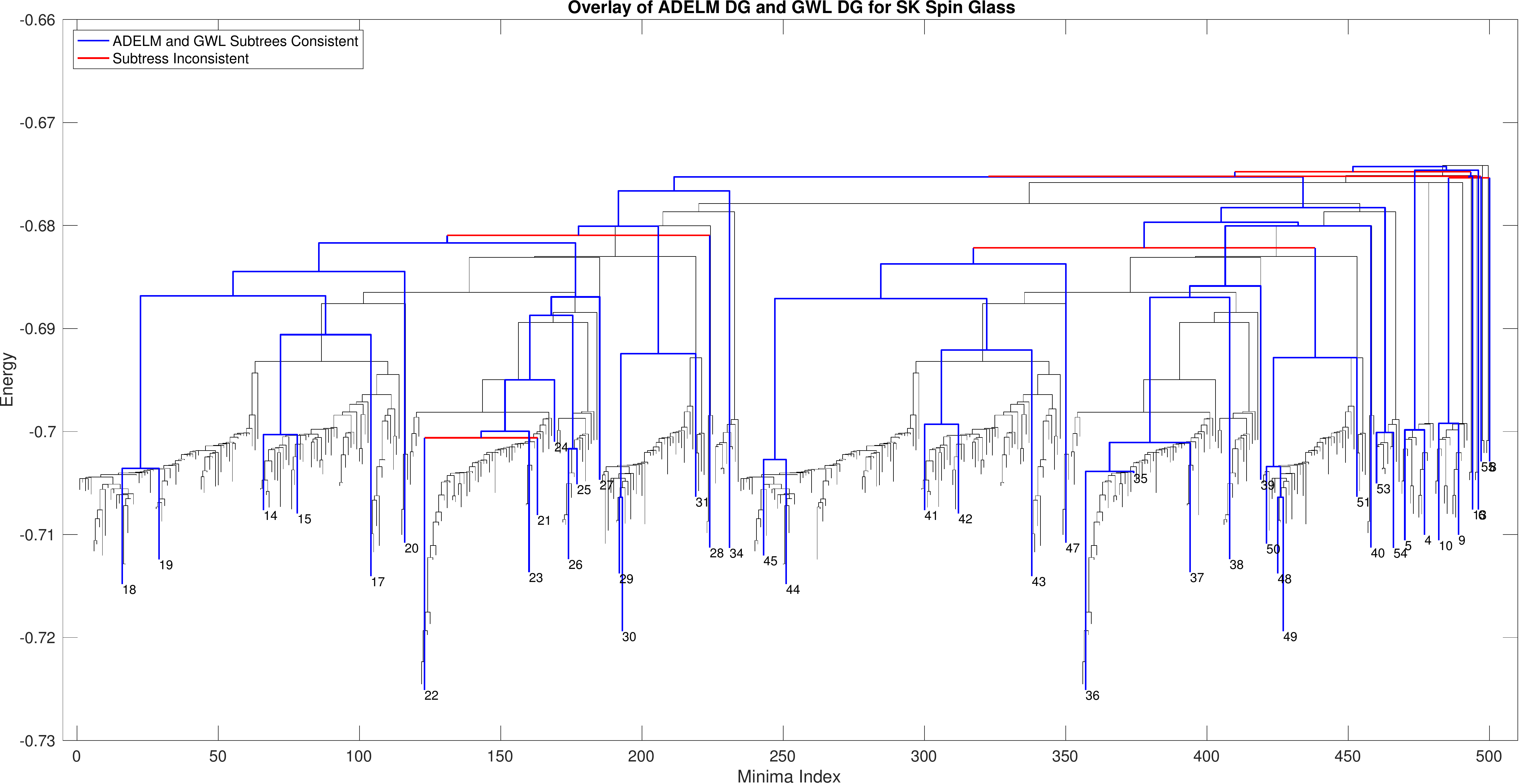}
	\caption{Overlay of Ising AD and GWL mapping. Blue horizontal lines indicate nodes of the ADELM DG where branch merges are consistent with the GWL DG. Red horizontal lines indicate nodes where the ADELM DG and GWL DG merge branches in a different order. The inconsistencies are minor and mostly occur in higher energy regions. Most inconsistencies only occur for a single merging, and are corrected by the next merge. The ADELM mapping effectively captures the macroscopic features of the GWL mapping.}
	\label{fig:ising_overlay}
\end{figure}

Our ADELM result is shown in Figure \ref{fig:SK_AD}, and a side-by-side comparison of the ADELM and GWL mappings is shown in Figure \ref{fig:ising_overlay}. The ADELM mapping identifies the lowest energy minima for all of the major basins of the landscape, as well as substructures within the basins. ADELM is also able to identify a number of basins which are stable but not recorded by the GWL mapping, since these local minima are not among the 500 lowest-energy minima in the landscape. Overall, 44 of the AD basins were also included in the GWL mapping, while 14 stable basins identified by AD were beyond the energy threshold of inclusion in the GWL mapping. 

\begin{figure}[h]
	\centering
	\includegraphics[width=0.6\textwidth]{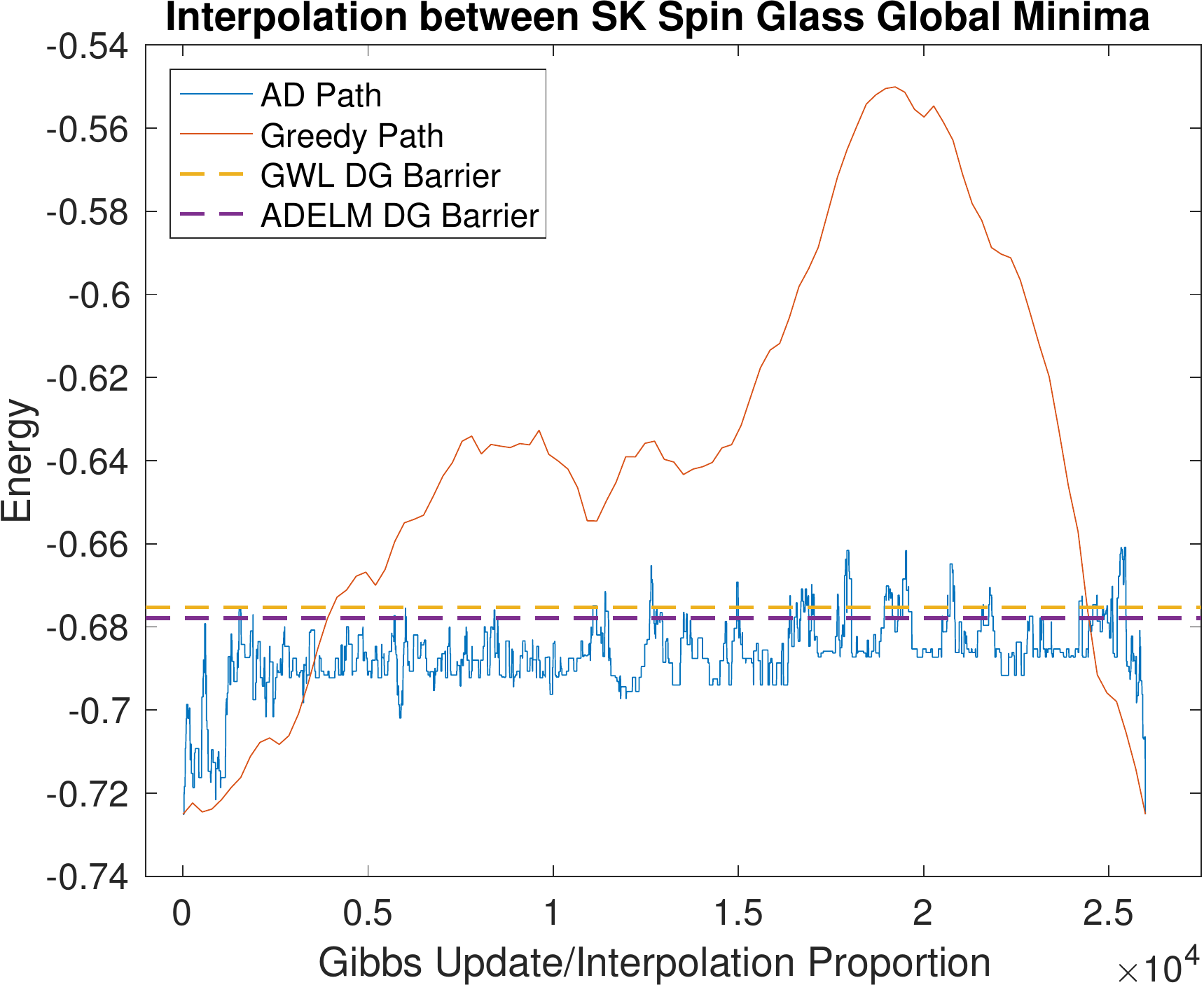}
	\medskip
	\caption{Interpolation between SK spin-glass global minima. The AD path travels in an energy spectrum very close to the barriers in the GWL and ADELM DG's.}
	\label{fig:ising_interpolation}
\end{figure}

The barriers estimated by the GWL mapping and the ADELM mappings are very similar, although in most cases the GWL barriers are slightly lower than the barriers estimated by AD. This shows that using a large number of minima during barrier estimation can be helpful, because shallow minima can help bridge the gap between stronger basins of attraction. Even though nearly all of the individual barriers identified by GWL are higher than the barriers identified by AD (see Figure \ref{fig:ising_interpolation}), the total information of barrier estimates between 500 minima can lead to overall barriers that are lower than the estimates obtained using only 58 minima. On the other hand, it might not be possible to exhaustively identify all of the relevant lowest-energy minima in other landscapes, and it is important to be able to accurately estimate barriers between distant minima without many shallow intermediate minima to connect the basins. Figure \ref{fig:ising_interpolation} shows an AD path between the two global minima of the SK spin-glass. The maximum energy along the path is only slightly above the barrier identified in GWL and ADELM DG's. This is evidence that AD can provide reliable interpolations between distant locations.

\subsection{Mapping an Energy Function of Image States}\label{subsec:exp_des}
For the rest our experiments, we use the ADELM Algorithm to map the energy landscape of ConvNet functions which are trained to model real image data. In this section, the target density has the form of the DeepFRAME Model \cite{lu, xie}
\begin{equation}
p(X | W) = \frac{1}{Z} \, \exp\{ F(X |  W)\} q(X)\label{eqn:deepframe_density}
\end{equation}
where $q$ is the prior distribution $\textrm{N}(0, \sigma^2 I_n)$, and $F(\cdot | W)$ is a ConvNet function with weights $W$. The target energy function has the form
\begin{equation}
E(X | W) = - F(X|W) + \frac{1}{2\sigma^2} || X ||_2^2. \label{eqn:deepframe_energy}
\end{equation}
All experiments are performed in Matlab using the MatConvNet package \cite{vedaldi} for ConvNet implementation.\\

\subsubsection{0-3 ELM in Image Space}
In the first ADELM experiment on image models, we map the energy landscape of a DeepFRAME function directly. The training data are the handwritten digits 0, 1, 2, and 3 from the first half of the MNIST \cite{lecun_cnn} testing set (according to the MNIST documentation, these digits are easier to classify). Each digit has about 500 training examples. The images were resized to $16 \times 16$ pixels, and each pixel intensity was discretized to $8$ values from $0$ to $255$. This was done so that a Gibbs sampler could be used as the sampling scheme $S$ in ADELM. We used a Gibbs sampler in this experiment because DeepFRAME landscapes learned with Langevin Dynamics have serious defects in the local mode structure which make mapping impossible. We hope to address this issue in future work, and eventually we would like to use Langevin Dynamics in the image space as our sampling procedure. 

In this experiment, the image space has 256 dimensions. This is larger than the spaces explored in the majority of past ELM experiments, which typically have at most 100 dimensions \cite{zhou, wales_ELM2, wales_ELM3}. However, use of a Gibbs sampler restricts the size of the image space, since Gibbs sampling scales poorly as dimension increases. We address this problem in later experiments by introducing a generator network, which has a low-dimensional latent space that facilitates movement in the image space of the DeepFRAME energy landscape. Composing a generator network and a DeepFRAME energy provides a way to map the pattern manifold for images of realistic size (see Section \ref{subsec:exp_coop}).

The weights $W$ are learned using the same method as \cite{xie} \emph{except} that Gibbs Sampling was used instead of Langevin Dynamics to synthesize images, for reasons explained above. The scoring function $F(\cdot | W)$ had a convolutional layer of 100 size 5$\times$5 filters followed by fully connected layers with 50 and 10 filters respectively. Each layer used the ReLu activation function. The weights were trained for 300 epochs with a learning rate $\gamma = 0.00007$ and $T=10$ Gibbs updates of the synthesized images for each 32-image batch.

We set the improvement limit to $M = 20$ and the distance resolution to 150 (each pixel has intensity between 0 and 255, so about half a pixel). The AD parameters were $T = 30$ and $\alpha = 1.05$. The proposal in Step 1 of the ADELM used random initialization. The mapping was done in two stages: a burn-in stage of 500 iterations, and a testing stage of 2000 iterations. After the burn-in stage, the global minima were consolidated by performing AD on all pairs of global minima using the same parameters as during mapping. This is done weed out extraneous minima that appear early during mapping when good global minima for each basin have not been found. In the testing stage, no new basins were identified, indicating that the mapping procedure has identified the main landscape features.

\begin{figure}[!ht]
	\centering
	\includegraphics[width=.4\textwidth]{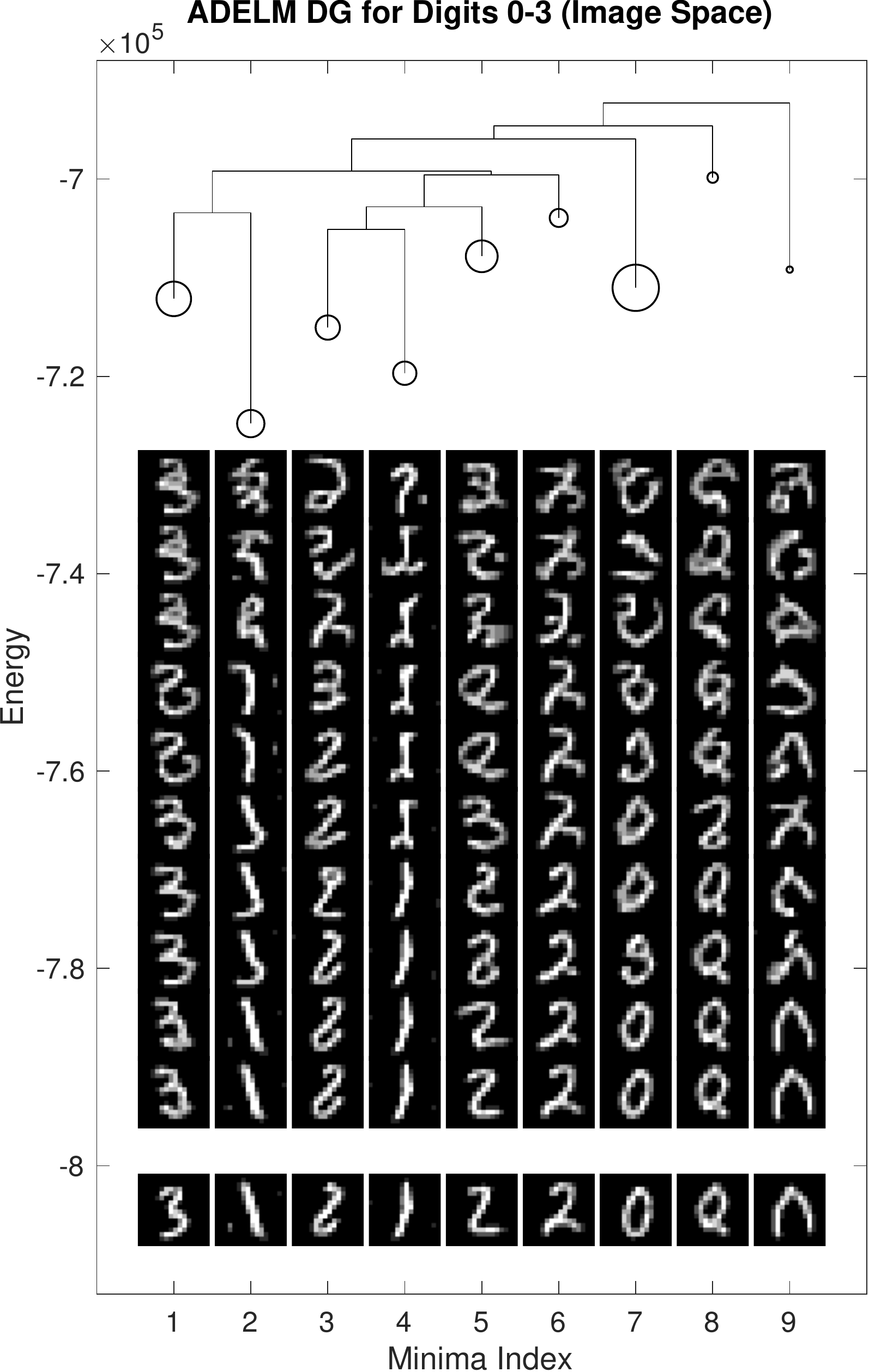}
	\caption{DG of Digits 0-3 ELM in Image Space. The left-tilted 1 digit and the 3 digit merge at low energy, as do the right-tilted digit 1 and different versions of the digit 2, while the digit 0 remains separate and merges at a higher energy. The images within basins mostly represent the same digit, although many noisy images are identified. The lower-energy images within the basins are well-formed digit images, while the high-energy images are not reliable digit representations. The circles and the organization of the images on the DG are explained in Section \ref{subsec:ELM_barrier}.}
	\label{fig:elm_0123_des}
\end{figure}

The ADELM results are shown in Figure \ref{fig:elm_0123_des}. The digits 0 and 3 are represented by a single minima, while the digit 1 was split between two basins according to direction of tilt and the digit 2 divided into three groups. We also found two stable basins that do not represent digits. See Section \ref{subsec:ELM_barrier} for an description of the DG layout.

\subsection{Mapping an Energy Function with Generator Proposals} \label{subsec:exp_gen}
The results in the previous section show that it is possible to map a DeepFRAME energy function by moving through the image space directly. However, many of the local minima identified during mapping are severely distorted digits, or images that do not resemble digits at all. While the DeepFRAME Model builds strong modes that approximate the manifold of the digit data, it also creates many accidental, higher-energy modes that warp the features of the true digit modes. In order to reduce the number of accidental modes discovered in the landscape, one could restrict the proposals in Step 1 of the ADELM algorithm to a region that is close to the true data manifold, instead of using a random image as in the previous section. Since the DeepFRAME energy function has only been trained to model a very small subset of the image space (a consequence of any learning algorithm based on Contrastive Divergence), restricting proposals to a region close to the data manifold reveals the structure of the landscape in the regions where the structure is most meaningful.

One way to restrict proposals to a well-formed region of the image landscape is to use a generator ConvNet \cite{goodfellow, han} as the proposal mechanism. As discussed in Section \ref{subsec:pattern_model}, an energy function $E$ can be trained jointly with a generator network $g$ using the Co-op Net Algorithm \cite{xie_coop}, so it is natural to use $g$ as a proposal mechanism for exploring $E$. This helps to limit exploration to a region of the image space where $E$ has learned to reliably model the image data. Moreover, mapping the local minima of proposals from the generator network provides a novel way to map the structure of the latent space of the generator networks. Although many authors have made observations about non-linear interpolations in the image space that occur when moving linearly through the latent space, there is no previous work that systematically maps the concepts of a latent space. Following the terminology of \cite{xie_coop}, we sometimes refer to the DeepFRAME energy $E$ as a \emph{descriptor} network.

In this section, we only use the generator network to propose new images as a starting point for local minima search. Local minima search and AD are both performed in the 16$\times$16 image space using only an energy network. We extend the role of the generator network further in Section \ref{subsec:exp_coop} by mapping a function of the form $E_{W_1,W_2} (Z) = E(g(Z|W_2) |W_1)$, where the energy $E(\cdot |W_1)$ is evaluated over the range of $g(\cdot | W_2)$. Since the latent space is much smaller than the image space, this formulation provides a way to efficiently map DeepFRAME energy functions defined over images of realistic size. Interestingly, it appears that the barriers in the landscape of the concatenated energy are more meaningful than the barriers in the raw DeepFRAME landscape, even though the local minima images are very similar (see Figures \ref{fig:elm_tree_spots1} and \ref{fig:elm_tree_spots2}).\\

\subsubsection{0-3 ELM in Image Space with Generator Proposals}

In our next experiment, we train a Co-Op Network to model the training images of the digits 0, 1, 2, and 3 from the previous section. The generator network uses a standard bivariate Gaussian $\textrm{N}(0,I_2)$ as the latent distribution and has three layers: two upsampling layers with the ReLu activation function sized 4$\times$4 and 5$\times$5 with 100 and 50 filters respectively, and a final upsampling layer of size  5$\times$5 with a tanh activation. An upsampling factor of 2 was used between all layers. The learning rate for the generative layer was 0.0003. The descriptor energy was initialized as the energy from Section \ref{subsec:exp_des} and the learning rate was very low, so the structure of the energy landscape should be similar to that of the previous experiment. Gibbs sampling was used to update the generator images instead of Langevin dynamics, as discussed in the previous section. 

In each iteration of the ADELM algorithm, we draw a random variable $Z$ from the latent distribution, find the image $g(Z|W_2)$ associated with the latent vector, and use this image as the starting point for local minima search. We used the same ADELM parameters as in Section \ref{subsec:exp_des}. We ran a burn-in sample of 500 iterations, consolidated the minima, and ran a test sample of 2000 iterations to obtain the results shown below.

\begin{figure}[h]
	\centering
	\includegraphics[width=.25\textwidth]{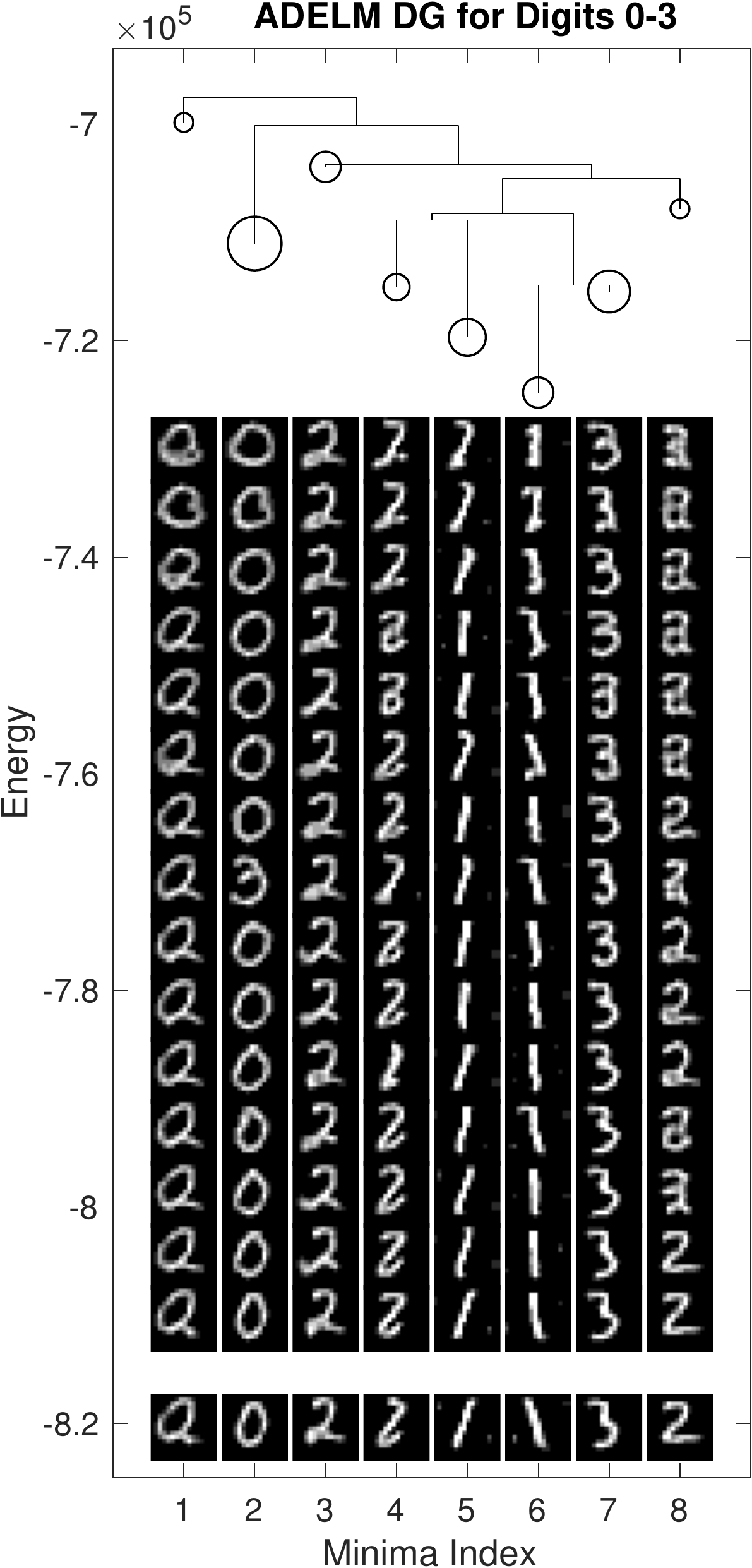}
	\includegraphics[width=.45\textwidth]{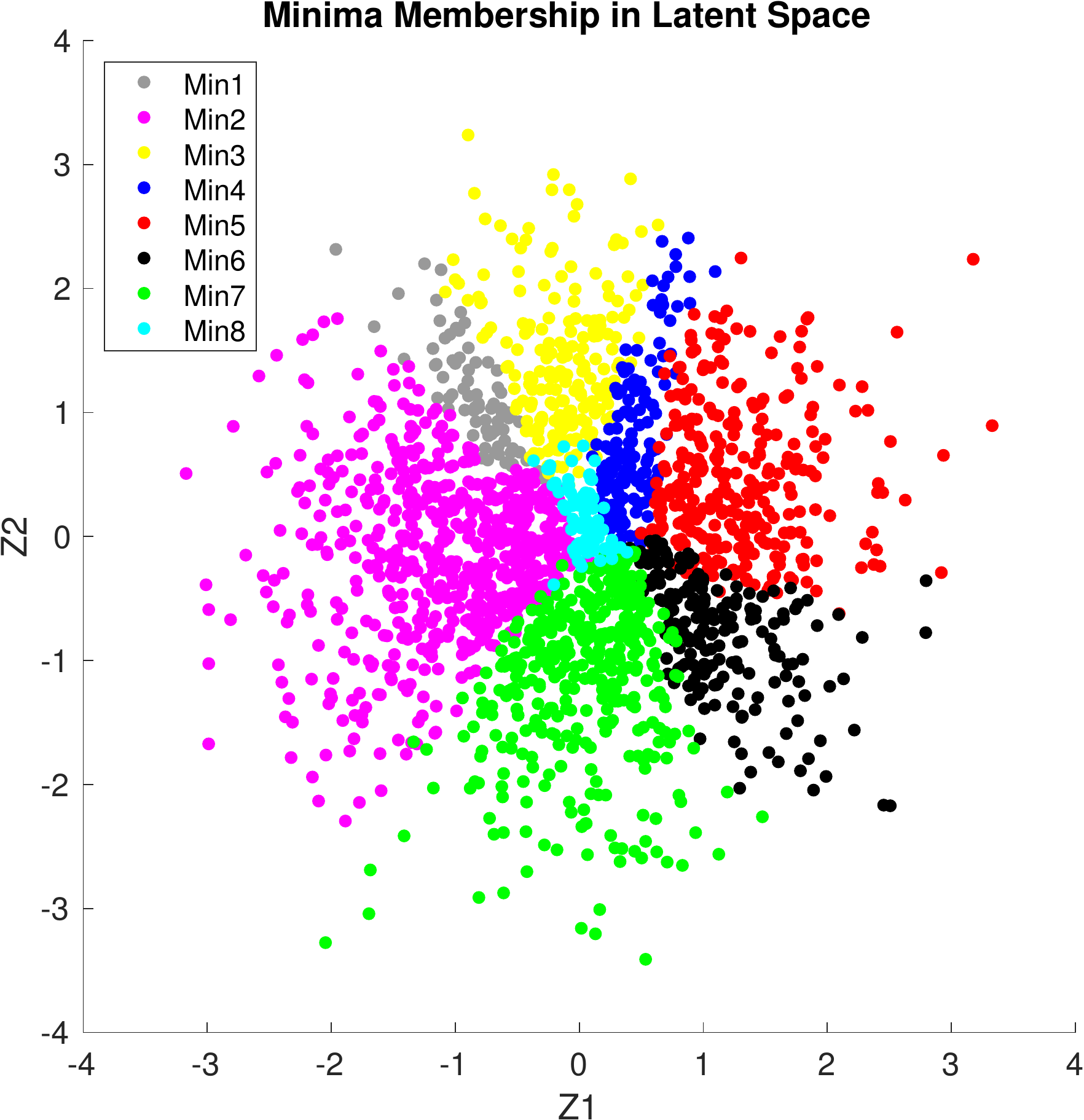} \\
	\hspace*{-.2cm}
	\includegraphics[width=.5\textwidth]{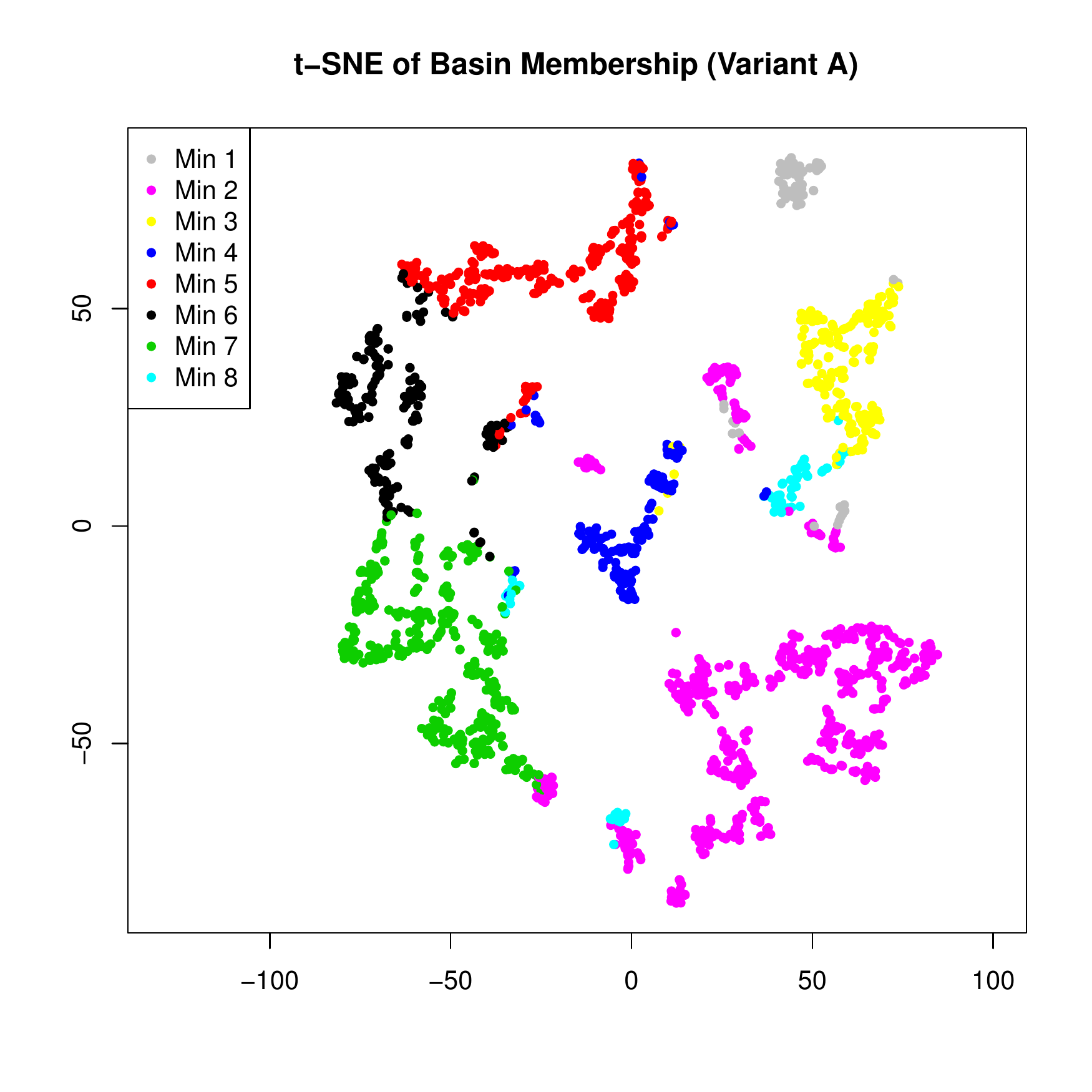}
	\includegraphics[width=.5\textwidth]{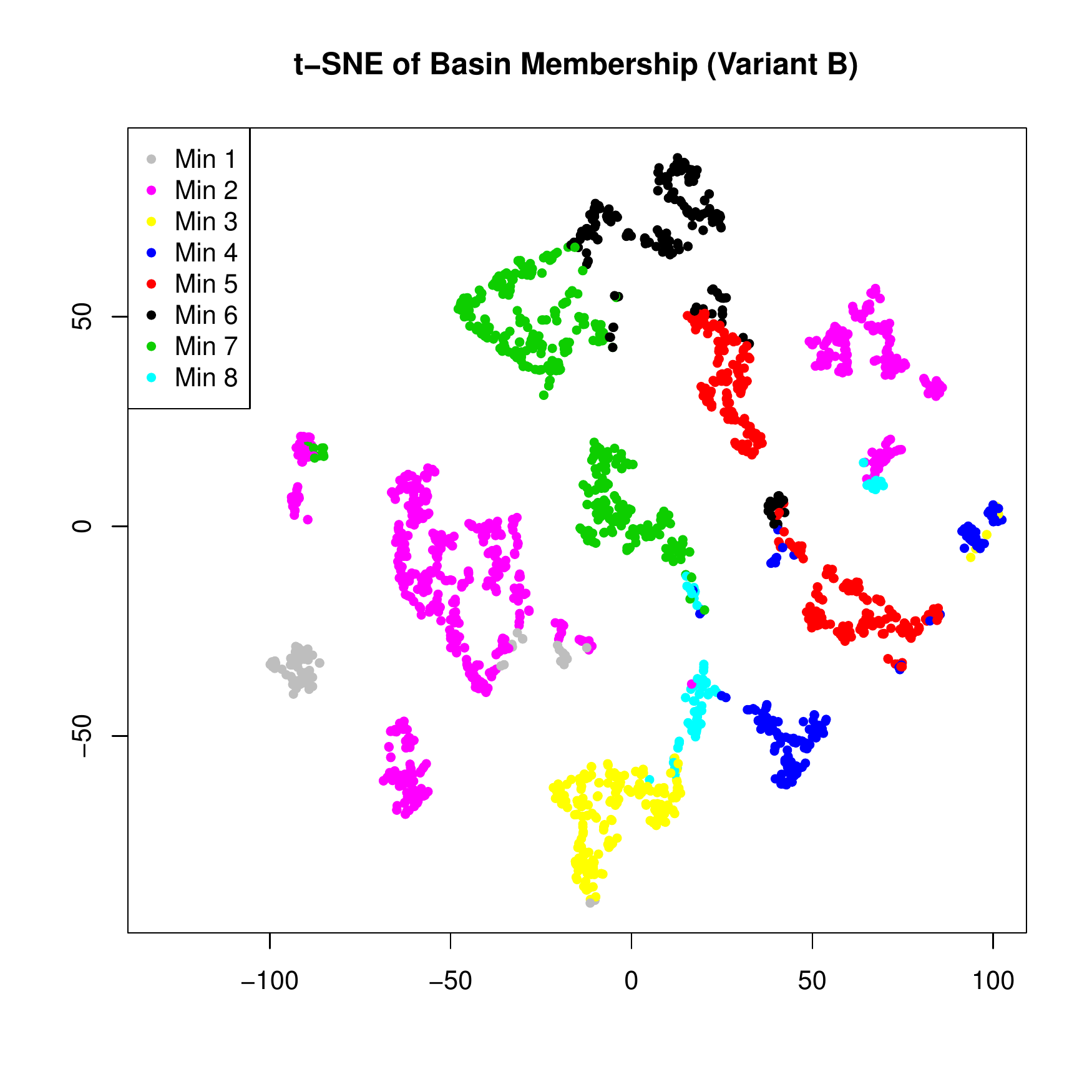}
	\caption{\emph{Top Left:} DG for Digits 0-3 ELM using generator proposals. \emph{Top Right:} Latent space $\textrm{N}(0, I_2)$ colored by basin membership found by ADELM. \emph{Bottom:} $t$-SNE visualizations of local minima found in ADELM colored by basin membership.}
	\label{fig:03gen}
\end{figure}

Figure \ref{fig:03gen} displays the mapping results. The minima are more consistent with the training data than those found when searching the image space from random initialization. The members of the image basins are coherent, and all but single basin correspond to recognizable digits. The DG structure and basin representatives are very similar to the results in Section \ref{subsec:exp_des}. The DG shows that the basins of the left-tilted 1 and the digit 3 merge at a relatively low energy, and the right-tilted 1 and the skinny 2 merge at a slightly higher energy. Two of the barriers found in the diagram are quite shallow. Nonetheless, these minima are still well-separated under AD at the parameter setting used during mapping. This is evidence that metastability rather than barrier height is best suited for grouping minima, especially if there are stable but flat basins in the landscape. Raw barrier height is not always representative of the dynamics of the system, and global basins influence the landscape well above the energy at which basins merge.

As noted earlier, it is possible to use the ADELM groupings to map the structure of the generator network. Figure \ref{fig:03gen} shows the latent vectors used to find proposal images, colored according to the ADELM groupings. The ADELM group labels form well-defined clusters in the latent space, and images representing the same digit are adjacent. Moreover, the arrangement of the latent space reflects the structure of the energy landscape. For example, the group of left-tilted 1's borders the group of the digit 3 in the latent space, and the group of right-tilted 1 borders the group of the skinny 2 digits. We also visualize the minima groupings using t-SNE embeddings. Since $t$-SNE is a random algorithm, two different results are given. The minima labels match well with the clusters found by $t$-SNE in both variants. \\

\subsubsection{Spots and Stripes ELM in Image Space with Generator Proposals}

\begin{figure}[h]
	\centering
	\includegraphics[width=.35\textwidth]{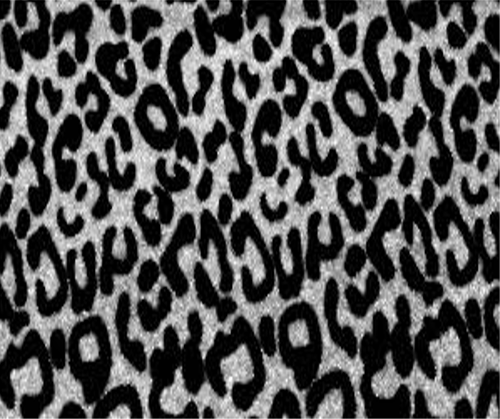}
	\hspace{.02\textwidth}
	\includegraphics[width=.35\textwidth]{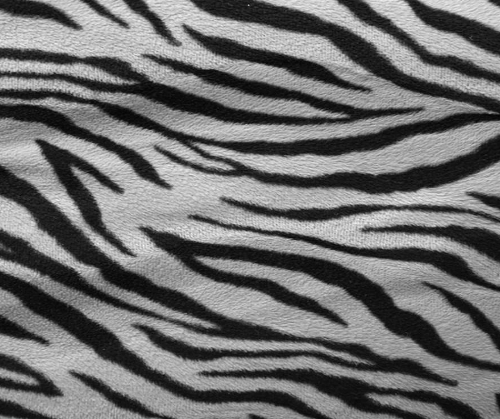}
	\caption{Spots and Stripes training images. 400 random image patches were taken from each image and resized to $16\times16$ for use as training data.}
	\label{fig:spotstripes}
\end{figure}

Next, we map a new descriptor and generator network trained to model small patches from texture images. The textures are shown Figure \ref{fig:spotstripes}. 500 small random patches were taken from each texture image and resized to 16$\times$16 pixels. We used the same network structure and training parameters as in the digits 0-3 ELM experiments, except that the latent space of the generator has 4 dimensions rather than 2. Gibbs sampling was used as the method for updating the synthesized images during training. The AD parameters were $T = 45$ and $\alpha = 1.3$. The other ADELM parameters and procedures were the same as in the previous two experiments. 

\begin{figure}[h]
	\centering
	\includegraphics[width=.9\textwidth]{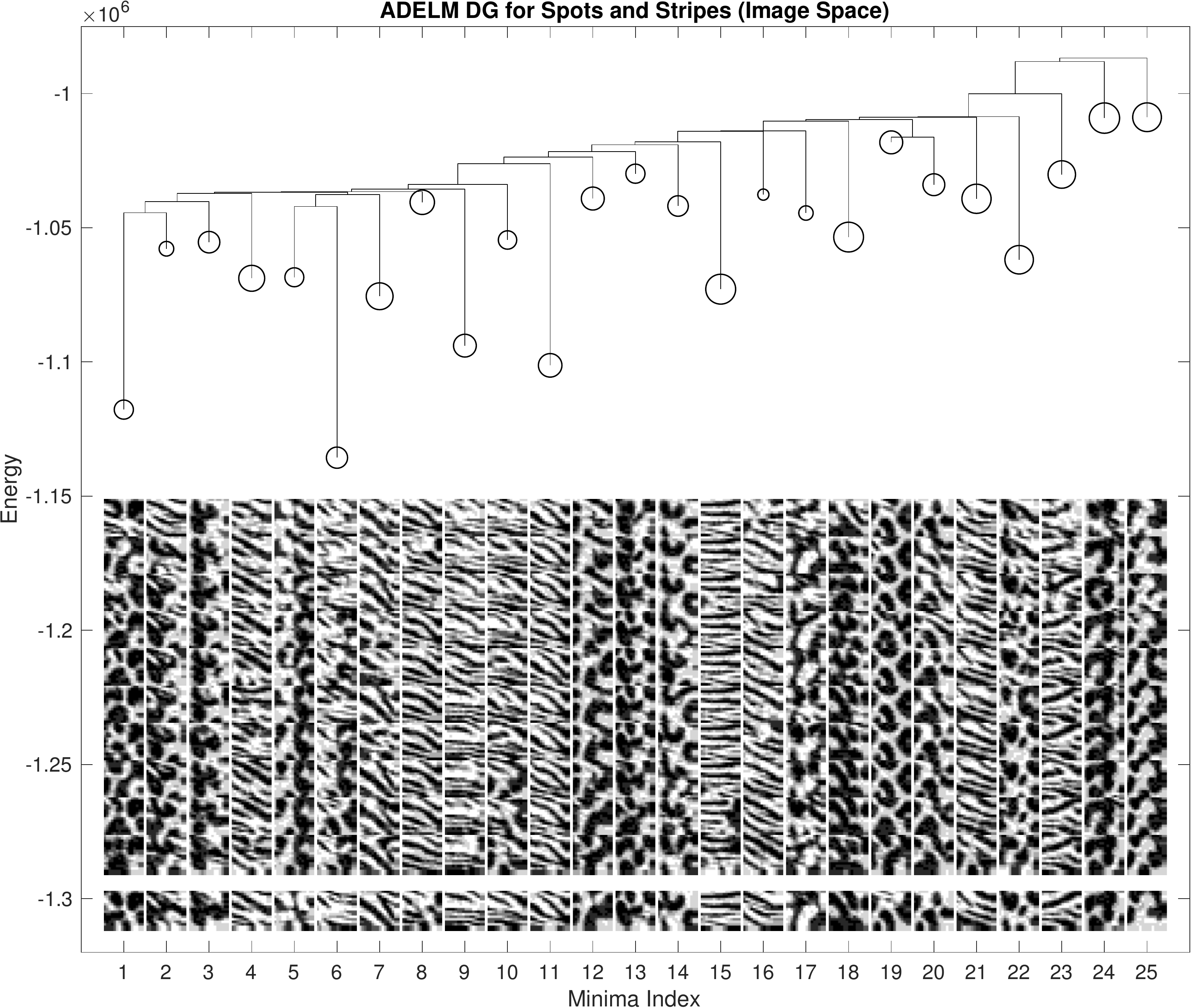}
	\caption{DG of Spots/Stripes ELM with generator proposals. The tree has a trivial structure where all minima merge along a single branch, and spots and stripes cannot be distinguished in the landscape. This happens because the DeepFRAME function creates accidental low energy regions between modes while it creates the modes. Introducing a generator network helps resolve this problem (see Figure \ref{fig:elm_tree_spots2}).}
	\label{fig:elm_tree_spots1}
\end{figure}

The results of the Spots and Stripes ELM with generator proposals are shown in Figure \ref{fig:elm_tree_spots1}. Although the appearance of the images within the minima groupings are consistent, the DG has a trivial structure. All minima merge into a single main branch, and the spots and stripes do not form separate regions of the energy landscape. This happens because the descriptor landscape has many accidental low energy regions that are formed as by-product of CD-style training which obscure the relations between the global basins. Diffusion paths travel through the accidental regions, creating low-energy connections throughout the landscape instead of meaningful barriers. We address this problem in the next section.

\subsection{Mapping Energy Functions over a Latent Space}\label{subsec:exp_coop}
The ideas of the previous section can be taken a step further by defining an energy function 
\begin{equation}
E_{W_1,W_2} (Z) = E(g(Z|W_2) |W_1)\label{eqn:coop_en}
\end{equation}
over the latent generator variable $Z \in \mathbb{R}^n$, where $E(\cdot | W_1)$ and $g(\cdot | W_2)$ are learned according to the standard Co-Op Net Algorithm. The energy $E_{W_1,W_2}$ is very similar to the energy used in the DGN-AM model \cite{DGNAM}, except that we train our generator and descriptor networks jointly to model a dataset of our choosing, while the DGN-AM experiments use a pre-trained GAN for the generator and a pre-trained classification neuron for the descriptor energy. Langevin Dynamics can be used to update the synthesized images during training, because the image space is never sampled directly during AD trials. In previous studies the latent space has a few hundred dimensions at most, and the experiments in Section \ref{subsec:exp_SK} through Section \ref{subsec:exp_gen} show that ADELM can handle such spaces using standard Gibbs sampling or Metropolis-Hastings sampling. The proposal in Step 1 of ADELM can be obtained by sampling from the latent distribution of the generator network. The formulation in (\ref{eqn:coop_en}) provides a way to efficiently map DeepFRAME functions defined over images of realistic size using ADELM.\\

\subsubsection{Spots and Stripes ELM in Latent Space}

\begin{figure}[h]
	\centering
	\includegraphics[width=.95\textwidth]{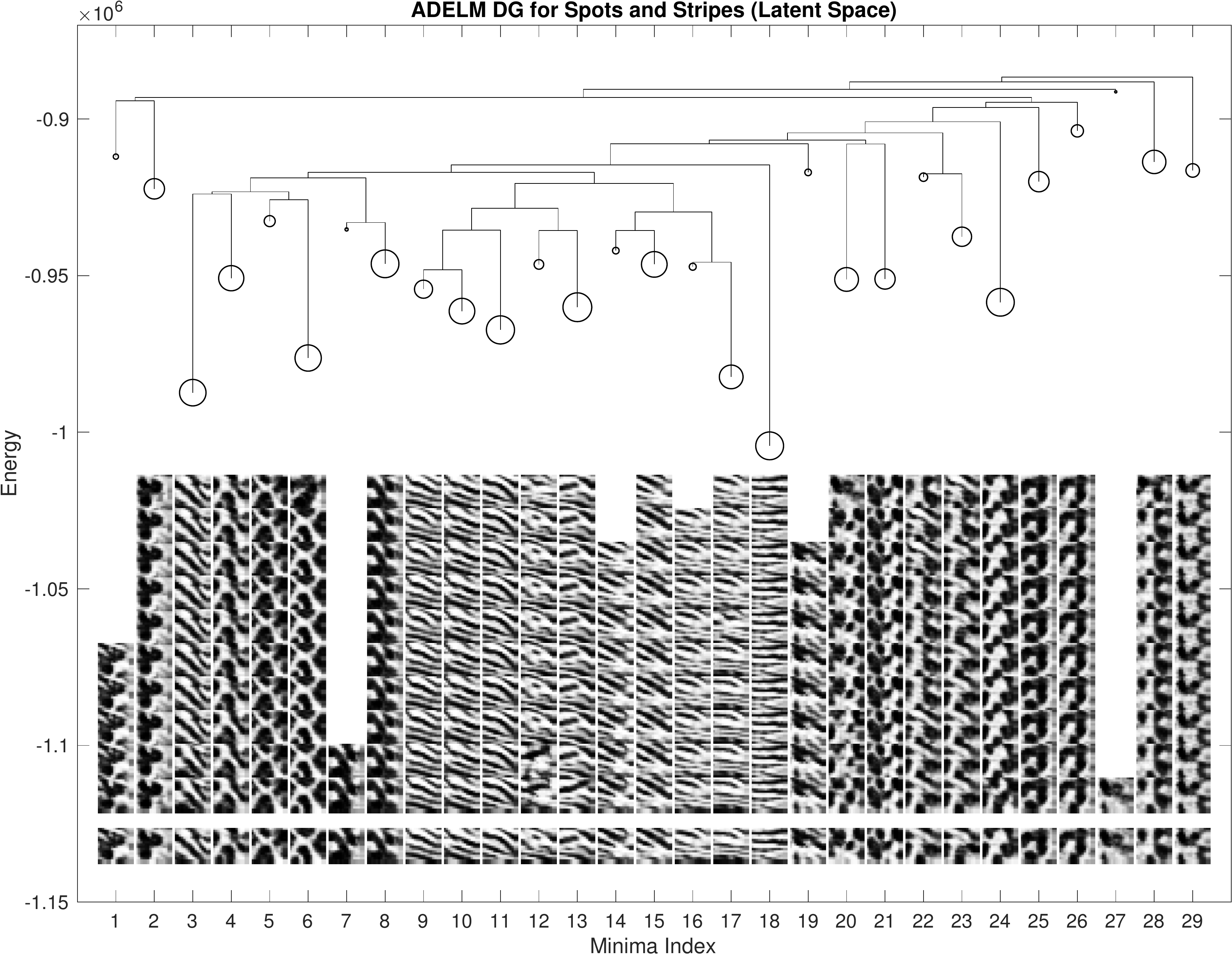}
	\caption{DG of Spots/Stripes ELM in latent space. The structure is much more complex than the landscape of the same networks mapped using AD directly in the image space, because the generator network has well-defined barriers between its clear images. The barriers generally respect the difference between the spots and stripes categories. On the other hand, the stripes images have lower energy than the spots images, causing the some of the spots images to merge with the main branch instead of forming their own grouping.}
	\label{fig:elm_tree_spots2}
\end{figure}

We use the same Spots and Stripes Co-Op Networks from previous section and implement ADELM in the 4-dimensional latent space of the generator network to map the energy function (\ref{eqn:coop_en}). We use Metropolis-Hastings with Gaussian proposals and a step size $\varepsilon = 0.025$ as our sampler $S$, and we set $M = 150$ and $\delta = 0.3$. The AD parameters are $T = 75$ and $\alpha = 300$. We ran 500 burn-in iterations, consolidated the minima, and ran 2000 testing iterations. The proposals in Step 1 of ADELM were drawn from the latent distribution $\textrm{N}(0,I_4)$. The testing results are shown in Figure \ref{fig:elm_tree_spots2}.

The minima images shown in Figure \ref{fig:elm_tree_spots2} are very similar to the images in Figure \ref{fig:elm_tree_spots1}, so ADELM recovers about the same global basins in both the image space and latent space. Minima found in the latent space are much more regular than the images from the previous experiment. The latent space DG has a more complex and meaningful structure than the trivial DG from the spots and stripes mapping in the image space. Unlike the raw descriptor energy over image space, the joint energy over the latent space contains definite boundaries between the basins of well-formed images.

Figure \ref{fig:elm_tree_spots2} shows some separation between the spots and stripes. Minima 4-8 merge at a low energy and represent spots (although Minimum 3, an oddball stripes image, belongs to the same subtree), while Minima 9-18 are all stripes. The remaining images are mostly spots, which merge with these two main subtrees at a higher energy. The algorithm for DG construction is greedy, because branches are merged at the lowest possible energy. This can cause the lower energy minima (the stripes) to disrupt the structure among the higher-energy minima (the spots) in the DG plot. A more nuanced visualization method which groups minima by minimizing the barriers within groups while maximizing the barrier outside of groups in the style of a community-detection algorithm might be able to separate the two categories even more effectively.\\

\subsubsection{Digits 0-9 ELM in Latent Space}

Next, we apply ADELM to map the energy (\ref{eqn:coop_en}) of Co-Op Networks modeling all of the digits of MNIST. We used the first half of the MNIST testing set as our training data (about 500 examples of each digit). This time, we increase image size to $64\times 64$ pixels. Since we will only sample in the \emph{latent} space, which has low dimension, we can use realistically-sized images during training. 

\begin{figure}[h]
	\centering
	\includegraphics[width=.9\textwidth]{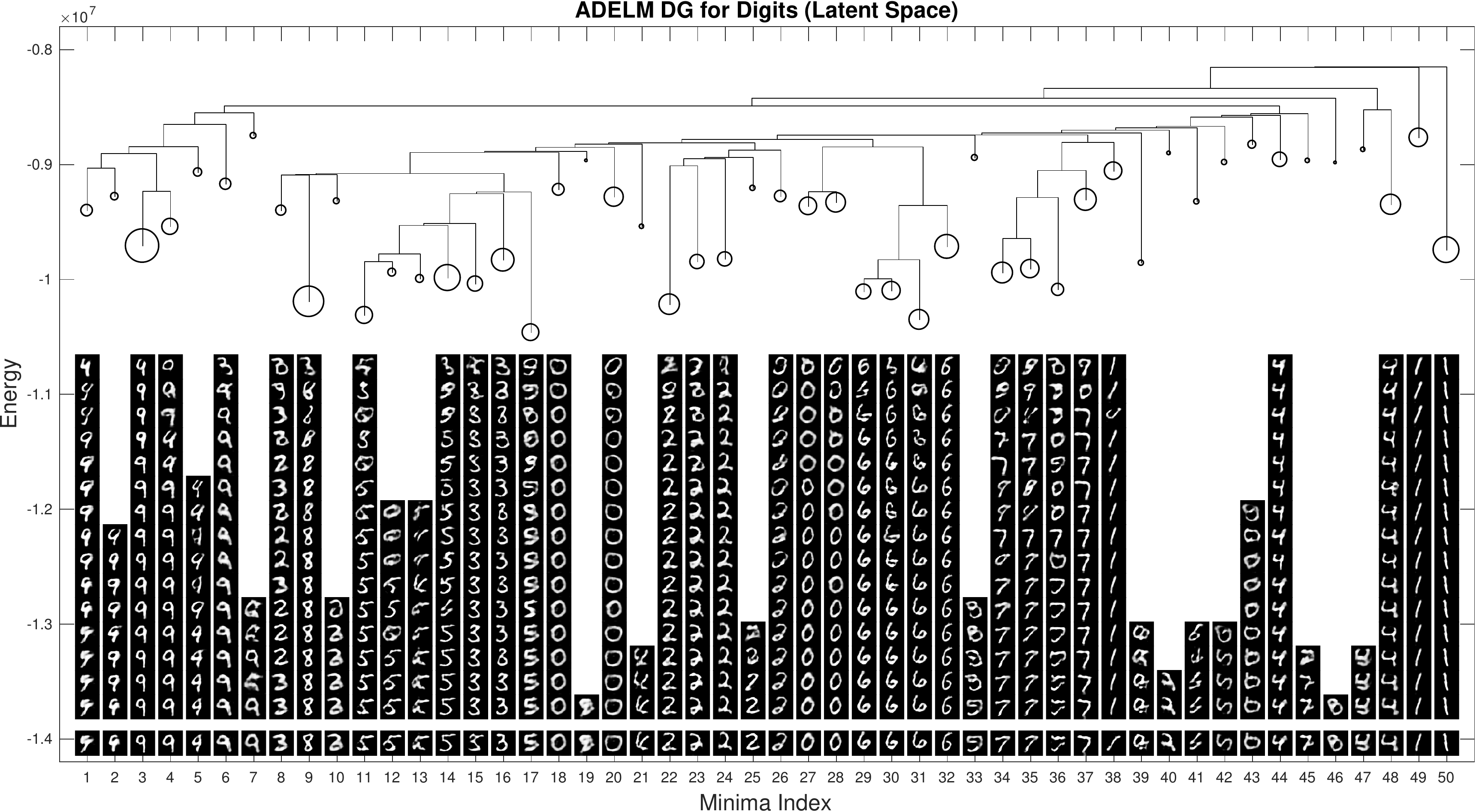}
	\caption{DG of Digits 0-9 ELM in latent space. The descriptor network is over $64\times 64$ images, but the generator latent space has only 8 dimensions, allowing for efficient mapping. Remarkably, all 10 digits have at least one well-separated branch in the DG. Minima representing the same digit generally merged at low energy levels. }
	\label{fig:elm_tree_digit}
\end{figure}

The descriptor network had three layers: two convolutional layers sized 5$\times$5 of 200 filters and 100 filters, and a fully-connected layer of 10 filters. Each layer is followed by a ReLu activation function. The latent generator distribution was the 8-dimensional normal $\textrm{N}(0,I_8)$. The generator network had three layers of size 4$\times$4, $7\times$7, and 7$\times$7 with 200, 100, and 10 filters respectively. ReLu was used as the activation for the first two layers, and tanh was used as the activation of the last layer. An upsampling factor of 4 was used after each generator layer. The AD parameters were $T = 1200$ and $\alpha =230$. The other ADELM parameters used were the same as in the Spots/Stripes Latent Space ELM. For mapping, 500 burn-in iterations and 5000 testing iterations were used, and the results are shown in Figure \ref{fig:elm_tree_digit}.

The ELM in Figure \ref{fig:elm_tree_digit} has many strong, well-separated energy basins. A close look at the DG shows that all 10 digits are represented by at least a single strong minima basin. The basin members and the global structure of the DG both match closely with human visual intuition. 

\begin{figure}[h]
	\centering
	\hspace*{-.25cm} \begin{tabular}{cc}
		\includegraphics[width=.5\textwidth]{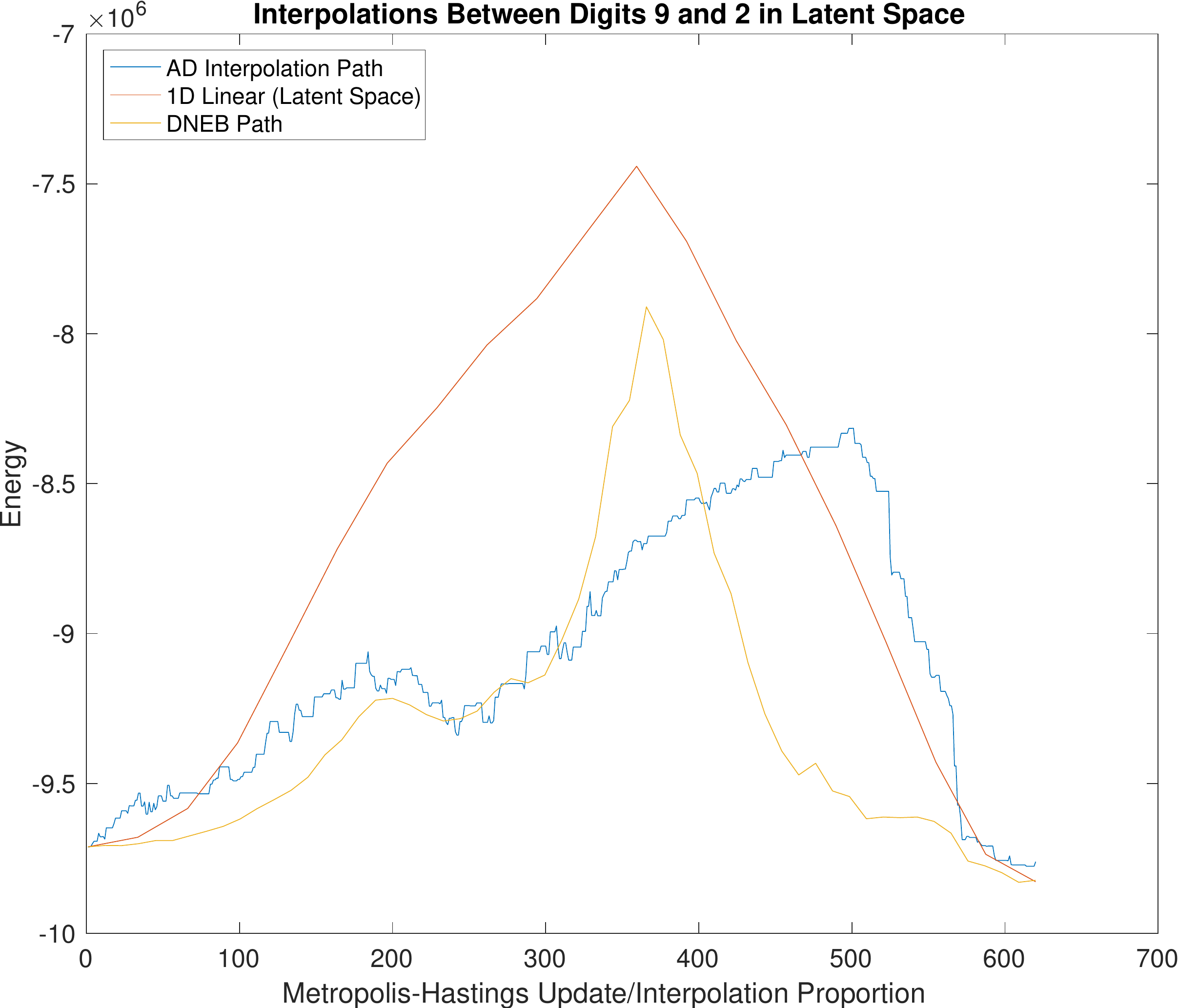} &
		\includegraphics[width=.5\textwidth]{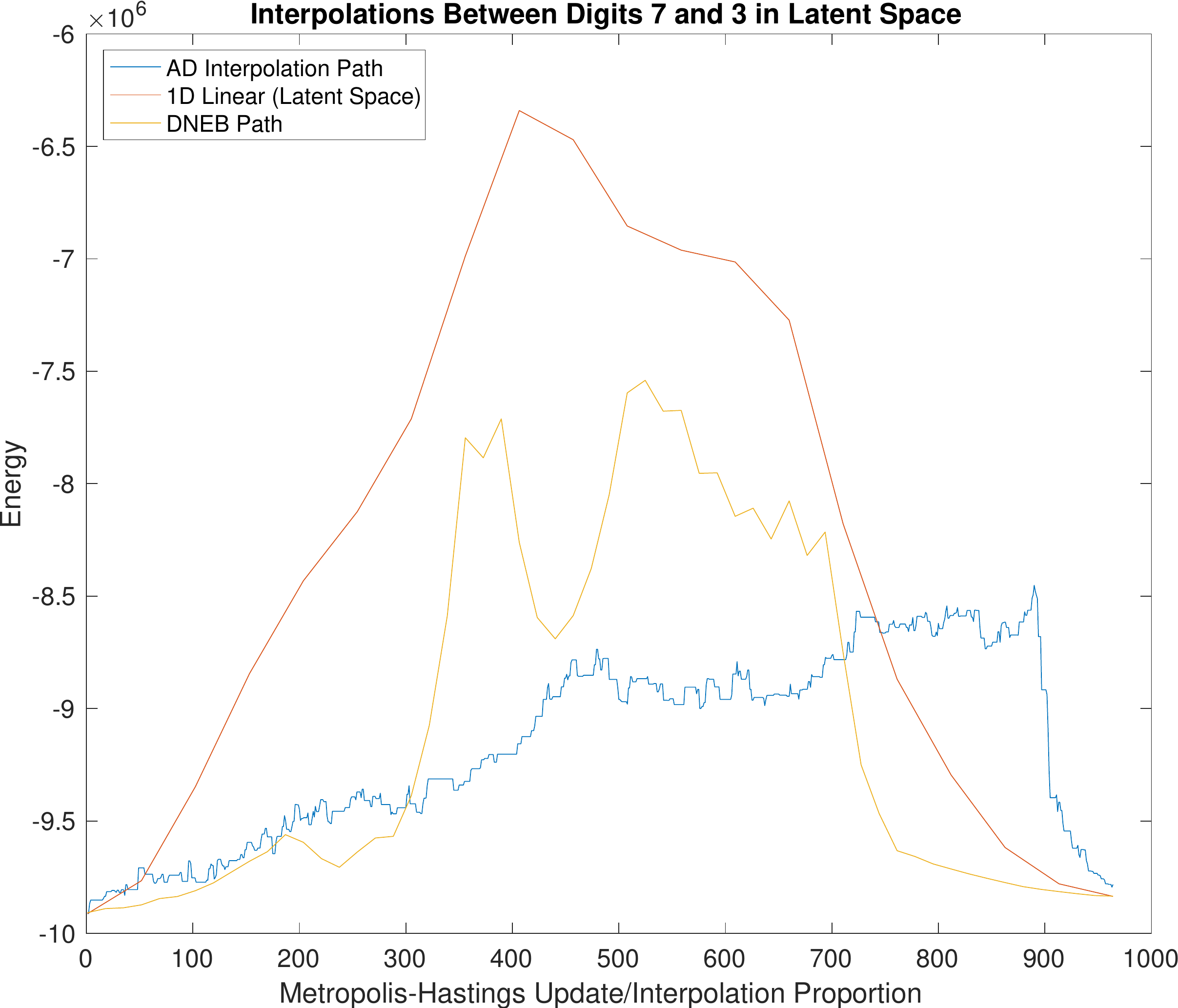}\\
		\hspace{-.2cm}\includegraphics[width=.5\textwidth]{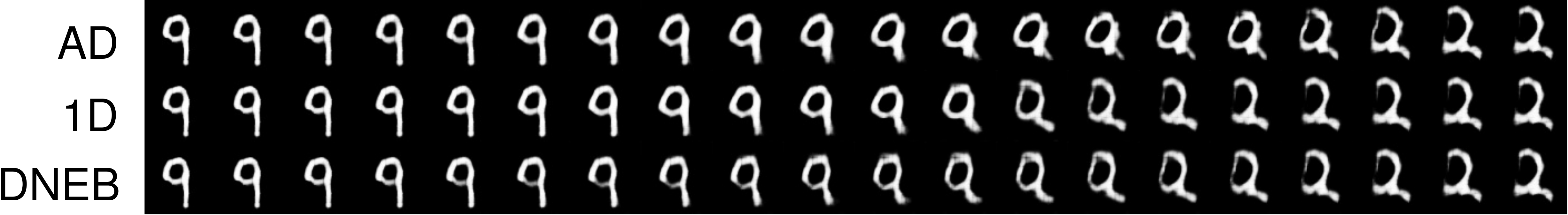} &
		\hspace{-.2cm}\includegraphics[width=.5\textwidth]{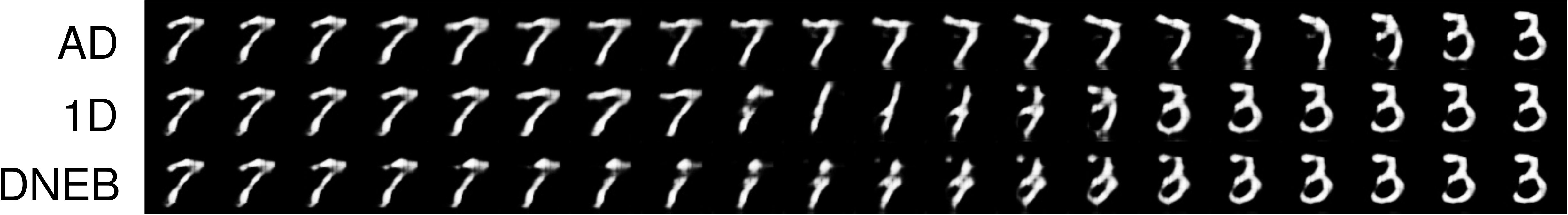}
	\end{tabular} 
	\caption{Comparison of barrier estimation between AD, DNEB \cite{wales_neb}, and 1D linear interpolation. \emph{Top:} Barrier estimation with 3 different methods. Both AD paths have lower energy than the 1D linear path and the DNEB path found by refining the 1D path. \emph{Bottom:} Visualization of interpolations. The DNEB interpolation is almost identical to the 1D interpolation, while AD finds a latent-space interpolation that differs from the 1D linear interpolation in appearance.}
	\label{fig:neb_comp}
\end{figure}

We compare MEP estimates from the DNEB \cite{wales_neb} method with MEP estimates from AD. The latent space has only 8 dimensions, so this landscape is a manageable test setting. DNEB uses the 1D linear space between minima as the initial path for further refinement, and Figure \ref{fig:neb_comp} shows that the DNEB image paths appear similar to the initial 1D path. On the other hand,  the AD paths travel through a different, significantly lower energy region of the landscape. It is well-known that 1D interpolations in the latent space provide more intuitive paths between minima than 1D interpolations in Euclidean space. Figure \ref{fig:neb_comp} shows that AD can find interpolations of the latent space that are distinct from the 1D latent interpolation in terms of both energy and appearance. AD and DNEB can be used in conjunction, since AD can provide a rich variety of initialization paths for further refinement by DNEB, which is currently limited to linear 1D initialization.

\subsubsection{Ivy Texton ELM in Latent Space}

\begin{figure}[h]
	\begin{subfigure}{.49\textwidth}
		\centering
		\includegraphics[width=.72\textwidth]{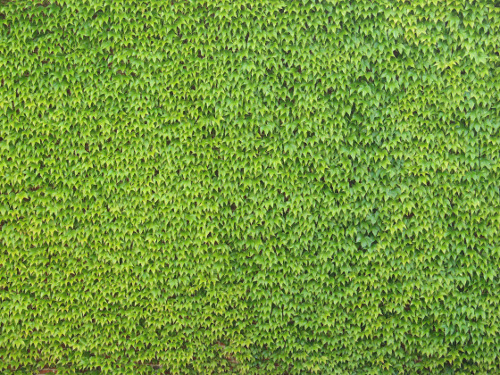}
	\end{subfigure}
	\hspace{-1cm}
	\begin{subfigure}{.49\textwidth}
		\centering
		\begin{tabular}{m{1cm}|m{.35cm}m{.35cm}m{.35cm}m{.35cm}m{.35cm}}
			Scale 1 & \includegraphics[scale = 0.25] {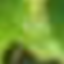} &\includegraphics[scale = 0.25]{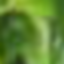} & \includegraphics[scale = 0.25]{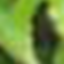} &  \includegraphics[scale = 0.25]{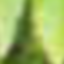} &  \includegraphics[scale = 0.25]{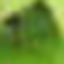} \\
			Scale 2 & \includegraphics[scale = 0.25]{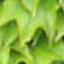} & \includegraphics[scale = 0.25]{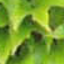} & \includegraphics[scale = 0.25]{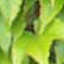} &  \includegraphics[scale = 0.25]{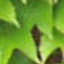} &  \includegraphics[scale = 0.25]{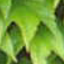} \\
			Scale 3 & \includegraphics[scale = 0.25]{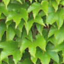}&  \includegraphics[scale = 0.25]{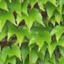} & \includegraphics[scale = 0.25]{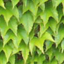} &  \includegraphics[scale = 0.25]{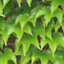} &  \includegraphics[scale = 0.25]{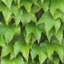}\\
			Scale 4 & \includegraphics[scale = 0.25]{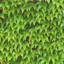} & \includegraphics[scale = 0.25]{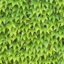} & \includegraphics[scale = 0.25]{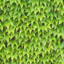} &  \includegraphics[scale = 0.25]{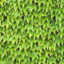} &  \includegraphics[scale = 0.25]{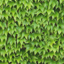}
		\end{tabular}
	\end{subfigure}
	\caption{Ivy texture image and image patches from four scales}
	\label{fig:ivy_ims}
\end{figure}

We now map a Co-Op Network trained on image patches from an ivy texture. At close range, ivy patches have distinct and recognizable structure, and the goal of the mapping is to identify the main patterns that recur in the ivy textons. Figure \ref{fig:ivy_ims} shows the entire ivy texture image along with image patches from the texture taken at four different scales. The networks in this experiment are trained to model 1000 image patches from Scale 2.

\begin{figure}[h]
	\centering
	\hspace*{-1cm} \includegraphics[height=8.5cm]{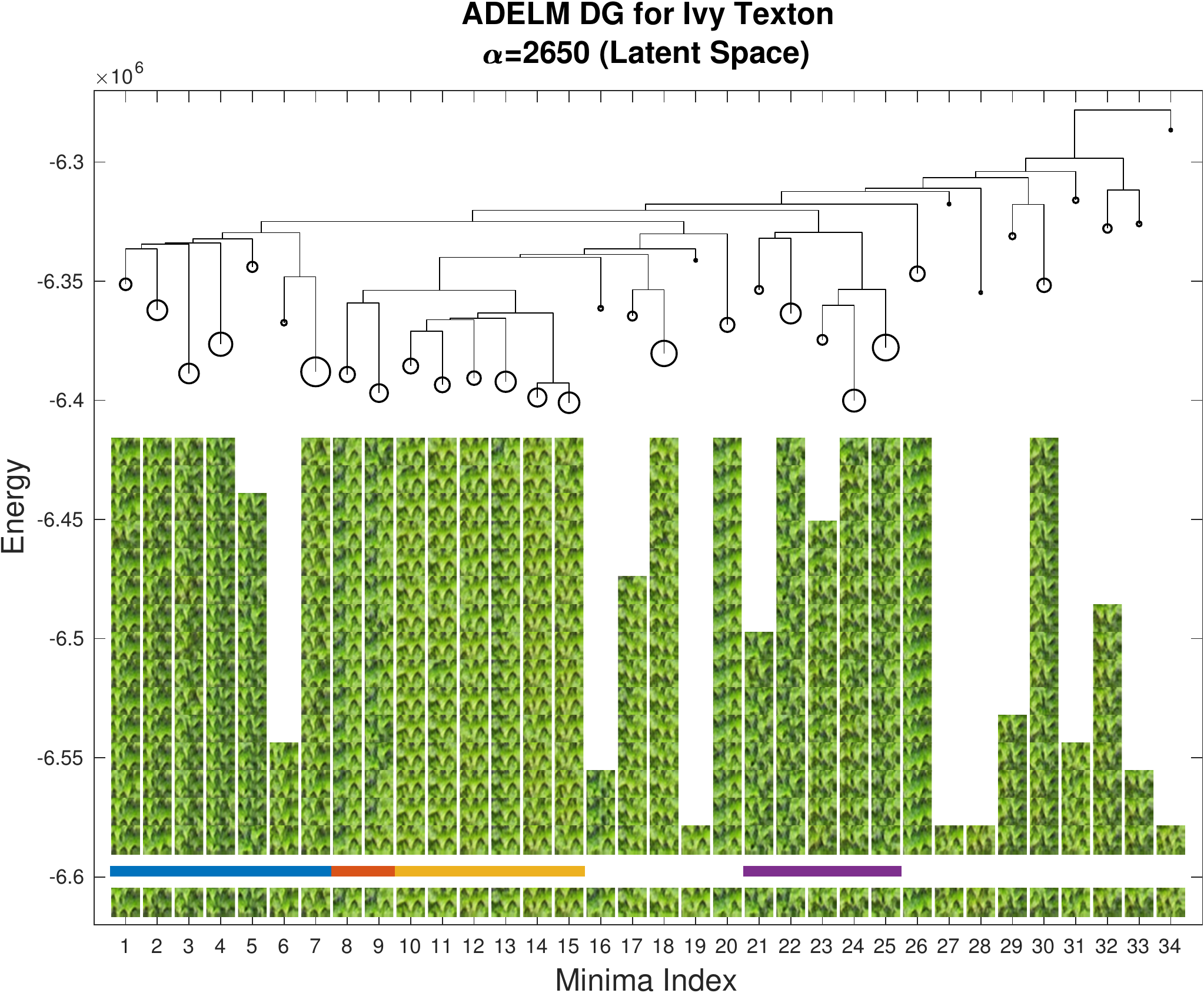}
	\includegraphics[height=8.5cm]{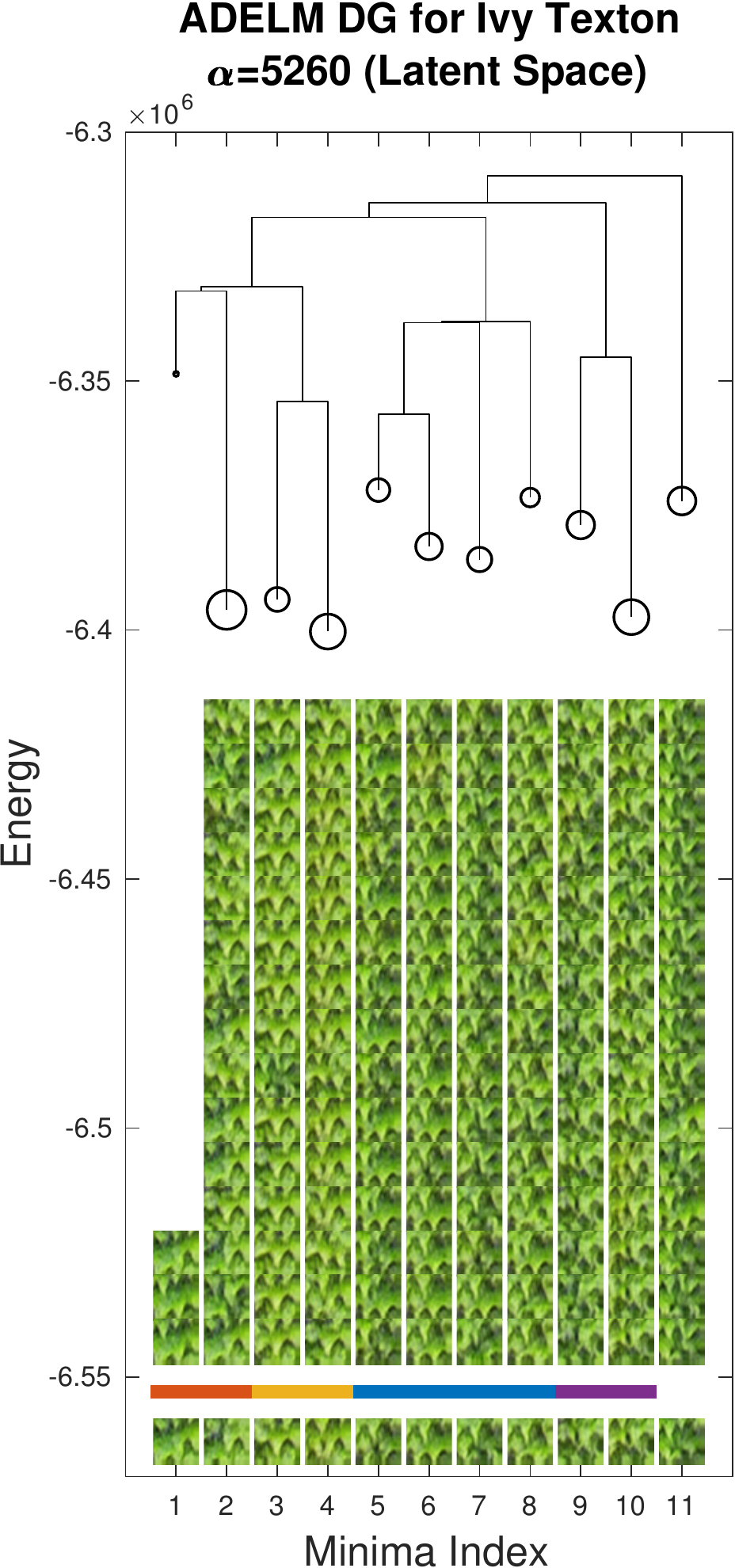}
	\caption{DG's of Ivy Textons for two different values of magnetization $\alpha$. Both mappings show 3 strong global basins and substructures within these basins that are stable at different magnetizations. There is no ground-truth grouping for texton image patches, so it is useful to map images structures at multiple resolutions to identify ``concepts" at different degrees of visual similarity. The colors below the basin representatives indicate regions that appear in both mappings.}\label{fig:dg_ivy}
\end{figure}

\begin{figure}[h]
\begin{center} 
	\textbf{Ivy Texton $\alpha=2650$ (Latent Space)} \\ 
	
	\hspace*{-.3cm}\begin{tabular}{M{.8cm}M{.8cm}M{10cm}M{1cm}} \toprule 
		Min. & Basin & Randomly Selected Members & Member \\ 
		Index & Rep. & (arranged from low to high energy) & Count \\ \midrule 
		1 & \includegraphics[scale = 0.5]{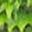} & \includegraphics[scale = 0.5]{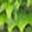} \includegraphics[scale = 0.5]{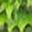} \includegraphics[scale = 0.5]{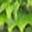} \includegraphics[scale = 0.5]{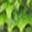} \includegraphics[scale = 0.5]{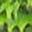} \includegraphics[scale = 0.5]{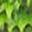} \includegraphics[scale = 0.5]{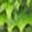} \includegraphics[scale = 0.5]{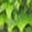} \includegraphics[scale = 0.5]{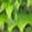} \includegraphics[scale = 0.5]{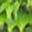} \includegraphics[scale = 0.5]{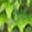} \includegraphics[scale = 0.5]{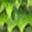} \includegraphics[scale = 0.5]{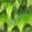} \includegraphics[scale = 0.5]{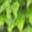} \includegraphics[scale = 0.5]{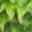}& 17\\ 
		3 & \includegraphics[scale = 0.5]{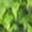} & \includegraphics[scale = 0.5]{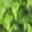} \includegraphics[scale = 0.5]{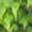} \includegraphics[scale = 0.5]{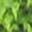} \includegraphics[scale = 0.5]{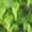} \includegraphics[scale = 0.5]{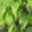} \includegraphics[scale = 0.5]{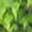} \includegraphics[scale = 0.5]{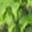} \includegraphics[scale = 0.5]{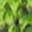} \includegraphics[scale = 0.5]{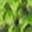} \includegraphics[scale = 0.5]{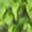} \includegraphics[scale = 0.5]{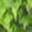} \includegraphics[scale = 0.5]{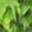} \includegraphics[scale = 0.5]{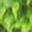} \includegraphics[scale = 0.5]{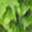} \includegraphics[scale = 0.5]{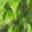}& 48\\ 
		8 & \includegraphics[scale = 0.5]{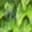} & \includegraphics[scale = 0.5]{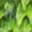} \includegraphics[scale = 0.5]{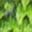} \includegraphics[scale = 0.5]{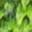} \includegraphics[scale = 0.5]{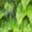} \includegraphics[scale = 0.5]{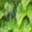} \includegraphics[scale = 0.5]{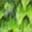} \includegraphics[scale = 0.5]{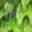} \includegraphics[scale = 0.5]{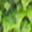} \includegraphics[scale = 0.5]{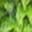} \includegraphics[scale = 0.5]{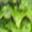} \includegraphics[scale = 0.5]{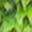} \includegraphics[scale = 0.5]{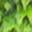} \includegraphics[scale = 0.5]{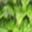} \includegraphics[scale = 0.5]{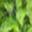} \includegraphics[scale = 0.5]{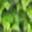}& 29\\ 
		9 & \includegraphics[scale = 0.5]{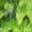} & \includegraphics[scale = 0.5]{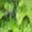} \includegraphics[scale = 0.5]{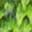} \includegraphics[scale = 0.5]{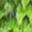} \includegraphics[scale = 0.5]{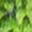} \includegraphics[scale = 0.5]{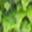} \includegraphics[scale = 0.5]{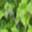} \includegraphics[scale = 0.5]{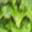} \includegraphics[scale = 0.5]{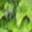} \includegraphics[scale = 0.5]{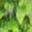} \includegraphics[scale = 0.5]{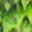} \includegraphics[scale = 0.5]{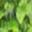} \includegraphics[scale = 0.5]{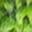} \includegraphics[scale = 0.5]{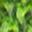} \includegraphics[scale = 0.5]{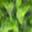} \includegraphics[scale = 0.5]{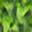}& 39\\ 
		10 & \includegraphics[scale = 0.5]{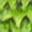} & \includegraphics[scale = 0.5]{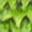} \includegraphics[scale = 0.5]{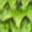} \includegraphics[scale = 0.5]{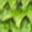} \includegraphics[scale = 0.5]{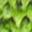} \includegraphics[scale = 0.5]{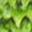} \includegraphics[scale = 0.5]{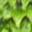} \includegraphics[scale = 0.5]{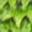} \includegraphics[scale = 0.5]{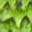} \includegraphics[scale = 0.5]{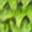} \includegraphics[scale = 0.5]{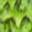} \includegraphics[scale = 0.5]{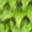} \includegraphics[scale = 0.5]{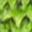} \includegraphics[scale = 0.5]{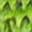} \includegraphics[scale = 0.5]{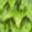} \includegraphics[scale = 0.5]{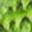}& 28\\ 
		15 & \includegraphics[scale = 0.5]{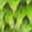} & \includegraphics[scale = 0.5]{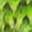} \includegraphics[scale = 0.5]{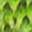} \includegraphics[scale = 0.5]{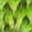} \includegraphics[scale = 0.5]{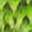} \includegraphics[scale = 0.5]{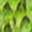} \includegraphics[scale = 0.5]{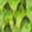} \includegraphics[scale = 0.5]{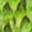} \includegraphics[scale = 0.5]{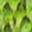} \includegraphics[scale = 0.5]{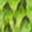} \includegraphics[scale = 0.5]{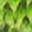} \includegraphics[scale = 0.5]{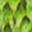} \includegraphics[scale = 0.5]{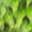} \includegraphics[scale = 0.5]{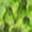} \includegraphics[scale = 0.5]{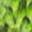} \includegraphics[scale = 0.5]{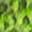}& 53\\ 
		22 & \includegraphics[scale = 0.5]{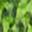} & \includegraphics[scale = 0.5]{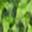} \includegraphics[scale = 0.5]{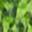} \includegraphics[scale = 0.5]{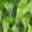} \includegraphics[scale = 0.5]{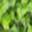} \includegraphics[scale = 0.5]{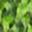} \includegraphics[scale = 0.5]{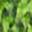} \includegraphics[scale = 0.5]{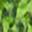} \includegraphics[scale = 0.5]{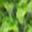} \includegraphics[scale = 0.5]{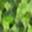} \includegraphics[scale = 0.5]{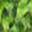} \includegraphics[scale = 0.5]{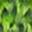} \includegraphics[scale = 0.5]{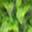} \includegraphics[scale = 0.5]{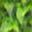} \includegraphics[scale = 0.5]{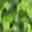} \includegraphics[scale = 0.5]{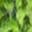}& 50\\ 
		25 & \includegraphics[scale = 0.5]{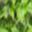} & \includegraphics[scale = 0.5]{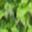} \includegraphics[scale = 0.5]{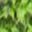} \includegraphics[scale = 0.5]{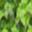} \includegraphics[scale = 0.5]{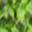} \includegraphics[scale = 0.5]{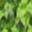} \includegraphics[scale = 0.5]{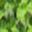} \includegraphics[scale = 0.5]{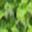} \includegraphics[scale = 0.5]{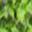} \includegraphics[scale = 0.5]{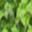} \includegraphics[scale = 0.5]{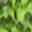} \includegraphics[scale = 0.5]{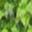} \includegraphics[scale = 0.5]{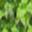} \includegraphics[scale = 0.5]{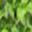} \includegraphics[scale = 0.5]{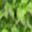} \includegraphics[scale = 0.5]{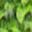}& 83\\ 
		\bottomrule
	\end{tabular} 
\end{center}
	\caption{Minima of Ivy texton with magnetization $\alpha=2650$ in latent space for the DG depicted in Figure~\ref{fig:dg_ivy}.} 	\label{fig:dg_ivy_table}
\end{figure}

The DG's for the ivy texton mapping in Figure \ref{fig:dg_ivy} show that the landscape is dominated by 3 or 4 global basins. The images within basins are very consistent, and the barriers between the basins are representative of visual similarity between the minima images. Unlike the digits mapping, there is no ground-truth for the minima groupings, so it is useful to explore the landscape at different energy resolutions to identify image groupings at different degrees of visual similarity. One major advantage of ADELM is the ability to perform mappings at different energy resolutions simply by changing the magnetization strength $\alpha$ used during the AD trials. Figure \ref{fig:dg_ivy} presents two mappings of the same landscape at different energy resolutions. The same landscape features appear in both mappings with more or less substructure depending on the magnetization strength.\\

\subsubsection{Multiscale Ivy ELM in Latent Space}

We continue our investigation of the ivy texture image from the previous section by mapping a Co-Op Network trained on 1000 image patches from each of the four scales shown in Figure \ref{fig:ivy_ims}. In this experiment, we want to investigate the differences in memory formation between the different scales. In particular, we are interested in identifying a relation between the \emph{metastability} of local minima in the landscape and the \emph{perceptibility} of visual difference among the minima. We expect to find fewer structures at the extreme scales. Image patches from Scale 1 are mostly solid-color images with little variety, which should form a few strong basins in the landscape. Image patches from Scale 4 have no distinct features and cannot be told apart by humans, so we expect these images will form a wide basin without much substructure. For the intermediate scales, we expect to find a richer assortment of stable local minima, because the intermediate scales contain more variation than Scale 1, but the variation still can be distinguished visually, in contrast to the Scale 4 images. 

Figure \ref{fig:dg_ivy_multi} shows the results of our mapping, and Figure \ref{fig:dg_ivy_multi_table} gives a closer look at basins from each scale. The structure of the landscape does indeed differ between the image scales. As expected, the memories from Scale 1 form a few strong and large basins. Scale 2 accounts for the majority of the basins in the landscape, since this scale contains the most variety of perceptible image appearances. The Scale 2 basins merge with the Scale 1 basins in the DG visualization, indicating that there are accessible low-energy connections between these regions of the landscape. The images from Scale 3 and Scale 4 each form a separate region of the energy landscape with little substructure. The mapping shows that the perceptibility threshold for ivy texture images (at least in terms of memories learned by the Co-Op Network) lies somewhere between Scale 2 and Scale 3. Above the perceptibility threshold, the network cannot reliably distinguish variation between images, and the landscape forms a single region with no significant substructure. It is difficult for a human to distinguish groups among images from Scale 3, so the perceptibility threshold for the network seems similar to that of humans.\\

\subsubsection{Cat Faces ELM in Latent Space}

\begin{figure}[h]
	\centering
	\includegraphics[width=\textwidth]{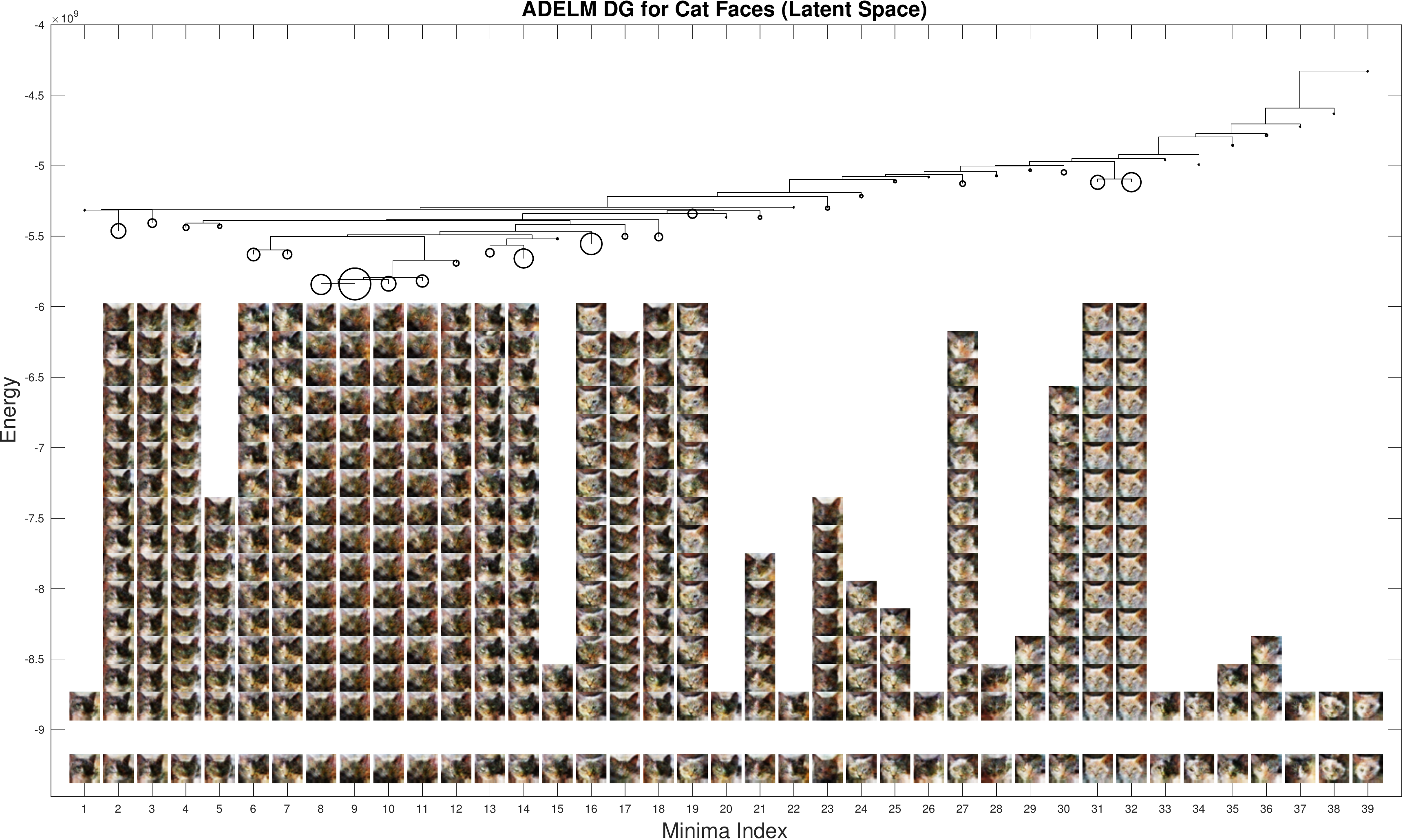}
	\caption{DG of Cat Faces in latent space. The landscape has a single global basin, likely because interpolations between cat faces that respect geometry and color constraints are easily found, unlike interpolations between digits, which must pass through a high-energy geometric configuration along the path. Despite the lack of overall landscape structure, AD is able to find meaningful image basins that show a variety of cat faces.} \label{fig:dg_cat}
\end{figure}

\begin{figure}[h]
\begin{center} 
	\textbf{Cat Faces (Latent Space)} \\ 
	
	\hspace*{-.3cm}\begin{tabular}{M{.8cm}M{.8cm}M{10cm}M{1cm}} \toprule 
		Min. & Basin & Randomly Selected Members & Member \\ 
		Index & Rep. & (arranged from low to high energy) & Count \\ \midrule 
		2 & \includegraphics[scale = 0.25]{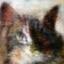} & \includegraphics[scale = 0.25]{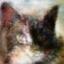} \includegraphics[scale = 0.25]{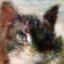} \includegraphics[scale = 0.25]{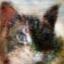} \includegraphics[scale = 0.25]{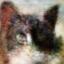} \includegraphics[scale = 0.25]{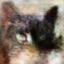} \includegraphics[scale = 0.25]{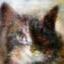} \includegraphics[scale = 0.25]{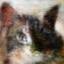} \includegraphics[scale = 0.25]{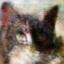} \includegraphics[scale = 0.25]{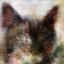} \includegraphics[scale = 0.25]{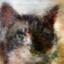} \includegraphics[scale = 0.25]{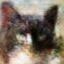} \includegraphics[scale = 0.25]{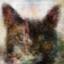} \includegraphics[scale = 0.25]{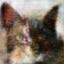} \includegraphics[scale = 0.25]{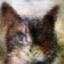} \includegraphics[scale = 0.25]{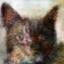}& 157\\ 
		4 & \includegraphics[scale = 0.25]{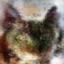} & \includegraphics[scale = 0.25]{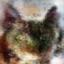} \includegraphics[scale = 0.25]{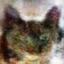} \includegraphics[scale = 0.25]{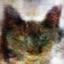} \includegraphics[scale = 0.25]{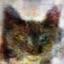} \includegraphics[scale = 0.25]{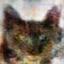} \includegraphics[scale = 0.25]{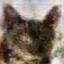} \includegraphics[scale = 0.25]{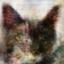} \includegraphics[scale = 0.25]{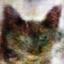} \includegraphics[scale = 0.25]{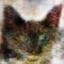} \includegraphics[scale = 0.25]{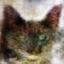} \includegraphics[scale = 0.25]{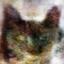} \includegraphics[scale = 0.25]{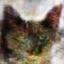} \includegraphics[scale = 0.25]{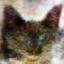} \includegraphics[scale = 0.25]{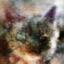} \includegraphics[scale = 0.25]{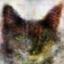}& 25\\ 
		6 & \includegraphics[scale = 0.25]{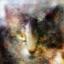} & \includegraphics[scale = 0.25]{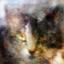} \includegraphics[scale = 0.25]{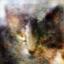} \includegraphics[scale = 0.25]{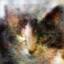} \includegraphics[scale = 0.25]{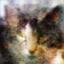} \includegraphics[scale = 0.25]{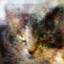} \includegraphics[scale = 0.25]{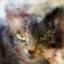} \includegraphics[scale = 0.25]{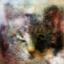} \includegraphics[scale = 0.25]{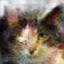} \includegraphics[scale = 0.25]{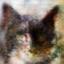} \includegraphics[scale = 0.25]{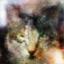} \includegraphics[scale = 0.25]{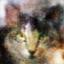} \includegraphics[scale = 0.25]{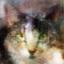} \includegraphics[scale = 0.25]{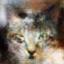} \includegraphics[scale = 0.25]{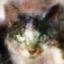} \includegraphics[scale = 0.25]{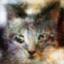}& 114\\ 
		9 & \includegraphics[scale = 0.25]{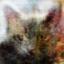} & \includegraphics[scale = 0.25]{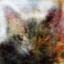} \includegraphics[scale = 0.25]{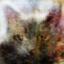} \includegraphics[scale = 0.25]{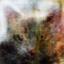} \includegraphics[scale = 0.25]{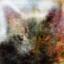} \includegraphics[scale = 0.25]{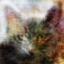} \includegraphics[scale = 0.25]{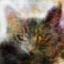} \includegraphics[scale = 0.25]{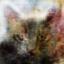} \includegraphics[scale = 0.25]{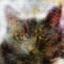} \includegraphics[scale = 0.25]{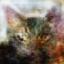} \includegraphics[scale = 0.25]{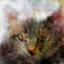} \includegraphics[scale = 0.25]{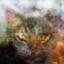} \includegraphics[scale = 0.25]{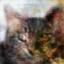} \includegraphics[scale = 0.25]{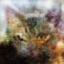} \includegraphics[scale = 0.25]{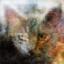} \includegraphics[scale = 0.25]{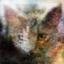}& 729\\ 
		16 & \includegraphics[scale = 0.25]{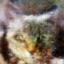} & \includegraphics[scale = 0.25]{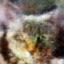} \includegraphics[scale = 0.25]{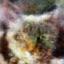} \includegraphics[scale = 0.25]{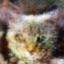} \includegraphics[scale = 0.25]{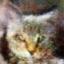} \includegraphics[scale = 0.25]{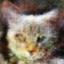} \includegraphics[scale = 0.25]{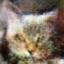} \includegraphics[scale = 0.25]{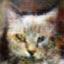} \includegraphics[scale = 0.25]{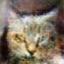} \includegraphics[scale = 0.25]{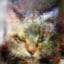} \includegraphics[scale = 0.25]{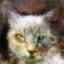} \includegraphics[scale = 0.25]{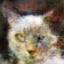} \includegraphics[scale = 0.25]{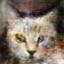} \includegraphics[scale = 0.25]{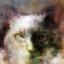} \includegraphics[scale = 0.25]{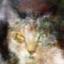} \includegraphics[scale = 0.25]{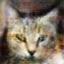}& 335\\ 
		18 & \includegraphics[scale = 0.25]{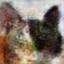} & \includegraphics[scale = 0.25]{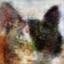} \includegraphics[scale = 0.25]{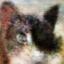} \includegraphics[scale = 0.25]{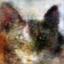} \includegraphics[scale = 0.25]{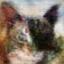} \includegraphics[scale = 0.25]{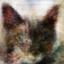} \includegraphics[scale = 0.25]{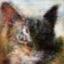} \includegraphics[scale = 0.25]{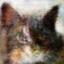} \includegraphics[scale = 0.25]{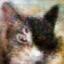} \includegraphics[scale = 0.25]{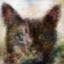} \includegraphics[scale = 0.25]{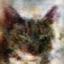} \includegraphics[scale = 0.25]{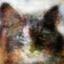} \includegraphics[scale = 0.25]{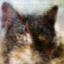} \includegraphics[scale = 0.25]{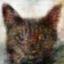} \includegraphics[scale = 0.25]{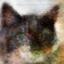} \includegraphics[scale = 0.25]{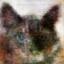}& 44\\ 
		32 & \includegraphics[scale = 0.25]{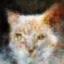} & \includegraphics[scale = 0.25]{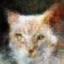} \includegraphics[scale = 0.25]{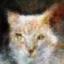} \includegraphics[scale = 0.25]{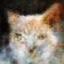} \includegraphics[scale = 0.25]{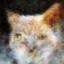} \includegraphics[scale = 0.25]{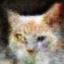} \includegraphics[scale = 0.25]{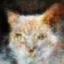} \includegraphics[scale = 0.25]{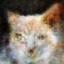} \includegraphics[scale = 0.25]{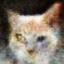} \includegraphics[scale = 0.25]{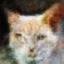} \includegraphics[scale = 0.25]{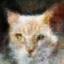} \includegraphics[scale = 0.25]{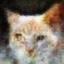} \includegraphics[scale = 0.25]{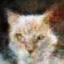} \includegraphics[scale = 0.25]{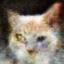} \includegraphics[scale = 0.25]{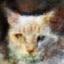} \includegraphics[scale = 0.25]{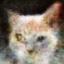}& 263\\ 
		\bottomrule
	\end{tabular} 
\end{center}
	\caption{Minima of Cat Faces in latent space for the DG depicted in Figure~\ref{fig:dg_cat}. Despite the shallow barriers found throughout the landscape, there are still metastable regions that capture consistent appearances.} 	\label{fig:dg_cat_table}
\end{figure}

For our final experiment, we map a Co-Op Network trained on aligned cat face images gathered from the internet. The results of our mapping are shown in Figure \ref{fig:dg_cat}, and Figure \ref{fig:dg_cat_table} gives a closer look at some of the basins. The DG has a single branch and the energy barriers are quite shallow. The main features of the local minima are the geometry and color of the cat faces, but these can be smoothly deformed during interpolation without encountering improbable images, in contrast to images such as digits, which must enter an improbable geometric configuration along an interpolation path. For this reason, the energy barriers throughout the cat landscape are very low. Nonetheless, the global basins found by ADELM coherently identify major groups of cat faces. AD can effectively identify landscape structure even when the majority of basin members have energy that is higher than the barrier at which the basin merges. This is further evidence that macroscopic basins influence the energy landscape in regions well above the lowest barrier between basins, and that metastability is a more suitable criterion than barrier height for identifying landscape structure.

\section{Conclusion}\label{sec:conclusion}
This work introduces a new MCMC tool called Attraction-Diffusion, which uses local sampling in an altered landscape to gain information about the relative stability of local minima in the original energy landscape. A unique feature of AD is the exploitation of the high autocorrelation that occurs when MCMC samples are trapped in local modes. In most MCMC research, this phenomenon is considered a major obstacle, but our work uses this aspect of MCMC sampling to measure landscape features. AD learns from both the success and failure of a local Markov sample as the chain is encouraged to escape from local barriers by an induced magnetization. The principles of AD can be traced back to magnetized energy functions from statistical physics, and AD can be interpreted as a way of measuring the metastable regions in the phase space $(T,\alpha)$.

We also introduce a new Energy Landscape Mapping algorithm called ADELM, which uses AD to sort local minima into separate metastable regions. By tuning the AD parameters to permit successful travel across the low-energy barriers \emph{within} metastable basins while respecting the large barriers \emph{between} metastable basins, it is possible to efficiently map the macroscopic structure of complex landscapes with noisy local structure. ADELM convergence is usually quite fast -- in the experiments presented, the main energy basins were identified within the first 100 iterations, and the mappings require only a few thousand iterations, whereas previous ELM methods require millions or billions of iterations. AD can also find energy barriers between minima that are lower than the barriers obtained from widely-used MEP estimation methods such as DNEB. The ADELM algorithm can be applied to a wide variety of continuous and discrete energy functions.

Using the ADELM Algorithm, we present a novel ELM application -- mapping the local minima structure of ConvNet functions which are trained to model real image data. Our experiments on Gibbs distributions defined by ConvNet functions show that it is possible to computationally identify image memories of a learned density, and that the structure of memories varies according to the images in the training set. The metastable basins identified by ADELM contain coherent groups of images, and the landscape structure of different image patterns reflects aspects of human visual intuition. Our mappings support the conjecture that the metastability of local minima is related to the perceptibility of differences between minima. The memory landscape forms many separate and stable basins when it is able to distinguish variation between low-entropy images, while large basins with little substructure are formed for memories of high-entropy images such as textures.

In future work, we plan to continue mapping the local minima structure of a wide variety of image densities. Although we encountered difficulties when directly mapping energy functions of realistically-sized image spaces, and overcame this by introducing a generator network with a low-dimensional latent space, we hope to eventually perform mapping using only an energy function over the image space. Energy functions trained using a CD-style algorithm develop serious degeneracies in regions of the image space that are distant from the pattern manifold, creating vast accidental low-energy basins that make mapping impossible. We hope to overcome this problem by using an ensemble of energy functions or energy functions at multiple scales to eliminate the accidental low-energy regions found in a single energy function. In the long term, we want to extend our method to identify hierarchical relations between image memories at different scales, and hope to define compositional ``dictionaries" that describe how image patches of smaller scales combine to form image patches of larger scales. ADELM shows great potential for future application to many other non-convex energy functions, including statistical loss functions and potential functions of physical systems.\\

\subsubsection*{Acknowledgments}

\noindent This work is supported by DARPA \#W911NF-16- 1-0579.

\bibliography{references}
\bibliographystyle{amsplain}

\end{document}